%% file: kammerer-eurocast2019.tex
\documentclass[runningheads]{llncs}
\usepackage{placeins}
\usepackage{graphicx}
\usepackage{pgfplots}
\usepackage{blindtext}
\usepackage[caption=false]{subfig}
\usepackage{tikz}
\usetikzlibrary{shapes,arrows,positioning,decorations.pathreplacing}
 
\begin{document}
\title{Data Aggregation for Reducing Training Data in Symbolic Regression}\footnotetext[1]{The final publication is available at \url{https://link.springer.com/chapter/10.1007\%2F978-3-030-45093-9\_46}}
\titlerunning{Data Aggregation in Symbolic Regression} 
%
\author{Lukas Kammerer\orcidID{0000-0001-8236-4294} \and
Gabriel Kronberger\orcidID{0000-0002-3012-3189} \and
Michael Kommenda}
\authorrunning{Kammerer et al.}
 
\institute{
Josef Ressel Center for Symbolic Regression \\
Heuristic and Evolutionary Algorithms Laboratory  \\
University of Applied Sciences Upper Austria, Hagenberg, Austria
\email{lukas.kammerer@fh-hagenberg.at}
}
\maketitle              
\begin{abstract}
The growing volume of data makes the use of computationally intense machine learning techniques such as symbolic regression with genetic programming more and more impractical. This work discusses methods to reduce the training data and thereby also the runtime of genetic programming. The data is aggregated in a preprocessing step before running the actual machine learning algorithm. K-means clustering and data binning is used for data aggregation and compared with random sampling as the simplest data reduction method. We analyze the achieved speed-up in training and the effects on the trained models' test accuracy for every method on four real-world data sets. The performance of genetic programming is compared with random forests and linear regression. It is shown, that k-means and random sampling lead to very small loss in test accuracy when the data is reduced down to only 30\% of the original data, while the speed-up is proportional to the size of the data set. Binning on the contrary, leads to models with very high test error.

\end{abstract}
\section{Introduction}

One of the first tasks in data-based modeling of systems is collection and selection of data with which a meaningful model can be learned. One challenge is to provide the right amount of data -- w.r.t.~both instances and features. There should be enough data to compensate noise and train a sufficiently complex model, but not too much to unnecessarily slow down the training. With the growing volume of data, especially the latter becomes more and more an issue when working with computationally intensive algorithms like genetic programming (GP).

\begin{figure}[t]
    \centering 
    \input{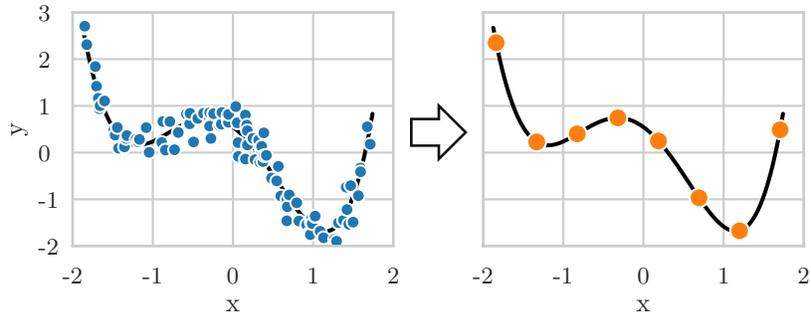}
    \caption{Schematic outline of extracting a data set of representative instances (right) out of original noisy data (left).}
    \label{fig:representatives}
\end{figure}

An intuitive idea to keep the training data small is outlined in Figure \ref{fig:representatives}. Instead of using all training data, only a few representative instances are extracted first and then used for the training. Ideally, these few instances retain all information that are necessary to train a well-generalizing model. This idea has been already applied for support vector machines for classification \cite{rychetsky1999accelerated} and regression tasks \cite{guo2007reducing}. In this previous work, the authors heuristically selected those instances, which are likely to determine the support vectors and therefore the SVM model. Another proposed approach for speeding up GP's evaluation is to use only a small random sample in every evaluation \cite{kommenda2010symbolic}. Kugler et al.~\cite{kugler2015softsensoren} suggested to aggregate similar instances together before training a neural network. Grouping together instances should cancel out noise and shrink the data set, which is similar to the initially mentioned idea.

This paper builds on the idea of Kugler et al.~\cite{kugler2015softsensoren} and uses clustering algorithms for aggregating and reducing training instances. The applied methods are random sampling, data binning and k-means clustering. The new, aggregated data are then used for training with different machine learning algorithms, with a focus on symbolic regression with GP. The trade-off between speed-up and loss in prediction accuracy due to potential removal of information is analyzed. We test how much we can reduce data so that we still can train accurate and complex models.

\section{Aggregation Methods}

All three aggregation methods require a predefined number of instances, which should be generated out of the original data. The first one, random sampling without replacement, serves as lower baseline. It will show, how much data can be removed without losing relevant information for modeling.

In data binning, the training data are aggregated based on the target variable. Similar to a histogram, instances with target values in fixed ranges are grouped together into bins. The range is determined by the minimum/maximum of the original target values and the predefined number of bins. To reduce the instances in a bin to one single instance, we use the median for all of the features. Binning should provide an equal target value distribution and reduce noise and variance \cite{kugler2015softsensoren}. However, binning also implies a ''many-to-one mapping'' between features and target variable, which incurs a loss of information in cases where interactions of features are relevant.

K-means clustering searches $k$ cluster centroids that minimize the sum of the Euclidean distance between each point and its closest centroid. In this work, the calculated centroids are used as new training data, while in usual applications, k-means is used as unsupervised learning algorithm to separate the data into $k$ groups. We deliberately consider both features and target values when aggregating the data because it is assumed that information about variable interactions is retained in contrast to data binning. The most common algorithm for determining the centroids is the heuristic Lloyd's algorithm \cite{lloyd1982least}, is infeasible for larger data sets with high $k$, since its runtime complexity is linear to the number of instances and $k$ -- especially when the goal is actually to speed up the overall modeling process. For our experiments, we use the mini-bach k-means algorithm \cite{sculley2010web}, which reduces the runtime of the preprocessing from days to a few minutes with comparable results.

\section{Experimental Setup}

This work focuses on the the effect of the described aggregation methods on symbolic regression with GP. We use the algorithm framework described by Winkler \cite{winkler:2009:thesis}, in which the prediction error of mathematical formulas in syntax tree representation as individuals is minimized. The algorithm implementation applies strict offspring selection with gender-specific selection \cite{affenzeller:2009}, a separate numerical optimization of constants in the formula \cite{kommenda2010symbolic} and explicit linear scaling \cite{keijzer2004scaled}. Since we focus on the effect of preprocessing on symbolic regression, we use a standard parameter setting which has shown in our experience to provide good results. GP's maximum selection pressure was set to 100, the mutation rate to 20\% and the population size to 300. The trees are build with a grammar of arithmetic and trigonometric symbols, as well as exponential and logarithm functions. The tree size is limited at most 50 symbols and a maximum depth of 30. The crossover operator is subtree swapping and mutation operators are point mutation, tree shaking, changing single symbols and replacing/removing branches \cite{affenzeller:2009}. We use for all experiments the \textit{HeuristicLab} framework\footnote{\url{https://dev.heuristiclab.com}} \cite{wagner2005heuristiclab}.

Random forests (RF) \cite{breiman2001random} and linear regression (LR) are run on the same data sets in order to provide comparability of the achieved results. Linear regression serves as a lower bound, as the resulting linear models have low complexity and all relations in the data sets are nonlinear. Random forests are used as a rough indicator, which accuracy values are achievable on different data sets, as this algorithm has shown to be both fast and often very accurate. The settings for RF were set train 50 trees and sample from 30\% of instances and 50\% of features for each tree.

This work uses four real-world data sets: The \textit{Chemical-I} data set (711 training instances, 57 features) \cite{White2013}, the \textit{Tower} data set \cite{White2013} (3136 training instances, 25 features), the \textit{SARCOS} data set \cite{pagie1997evolutionary} (44 500 training instances, 21 features)and the \textit{puma8NH} data set (6144 training instances, 8 features) from the Delve repository\footnote{\url{https://www.dcc.fc.up.pt/~ltorgo/Regression/puma.html}}. The data sets are split into training and test data set according to the cited papers. The data sets are normalized to a mean of zero and a standard deviation of 1 in the training data, which is especially important for the k-means algorithm, which uses the Euclidean distance.

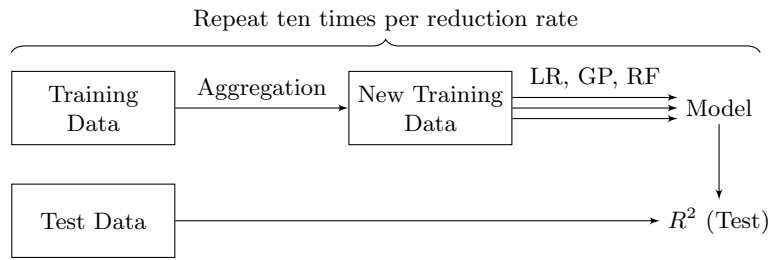
\begin{figure}[ht]
    \centering
    \definecolor{linecolor}{rgb}{0.8,0.8,0.8}

    \tikzstyle{decision} = [text centered, inner sep=.3em]
    \tikzstyle{block} = [rectangle, draw,
        text width=6em, text centered, minimum height=3em]
    \tikzstyle{line} = [draw, -latex']
        
    \begin{tikzpicture}[node distance = 1.5cm, auto]
        \node [block] (trainingData) {Training Data};
        \node [block, right=2.3cm of trainingData] (newTrainingData) {New Training Data};
        \node [decision, right=2.2cm of newTrainingData] (model) {Model};
        \node [decision, below of=model] (r2) {$R^2$ (Test)};

        \node [block, below of=trainingData] (testData) {Test Data};

        \path [line] (trainingData) -- node{Aggregation} (newTrainingData);
        \path [line, transform canvas={yshift=4pt}] (newTrainingData) -- node[text width=3.5cm, text centered]{LR, GP, RF} (model); 
        \path [line] (newTrainingData) -- (model);
        \path [line, transform canvas={yshift=-4pt}] (newTrainingData) -- (model);
        \path [line] (model) -- (r2);
        \path [line] (testData) -- (r2);

        \draw[decoration={brace,raise=5pt, amplitude=5pt},decorate, transform canvas={yshift=16pt}]
  (trainingData.west) -- node[above=10pt] {Repeat ten times per reduction rate} (model.east);
    \end{tikzpicture}

    \caption{Experiment workflow of reducing data for each data set and reduction rate.}
    \label{fig:workflow}
\end{figure}

To analyze the tradeoff between accuracy and speed-up, the training data are reduced stepwise, starting from a reduction to one percent up to 50 percent of the original training data size. For each of these steps, ten reduced training data sets are generated. The three machine learning algorithms are then run on each of these reduced data sets. Figure \ref{fig:workflow} outline the estimation of the generalization error, where we evaluate the trained models on the not-aggregated test data. The generalization error is measured with Pearson's $R^2$ coefficient, which describes the correlation between actual target values predictions in the interval $[0,1]$.

\section{Results}

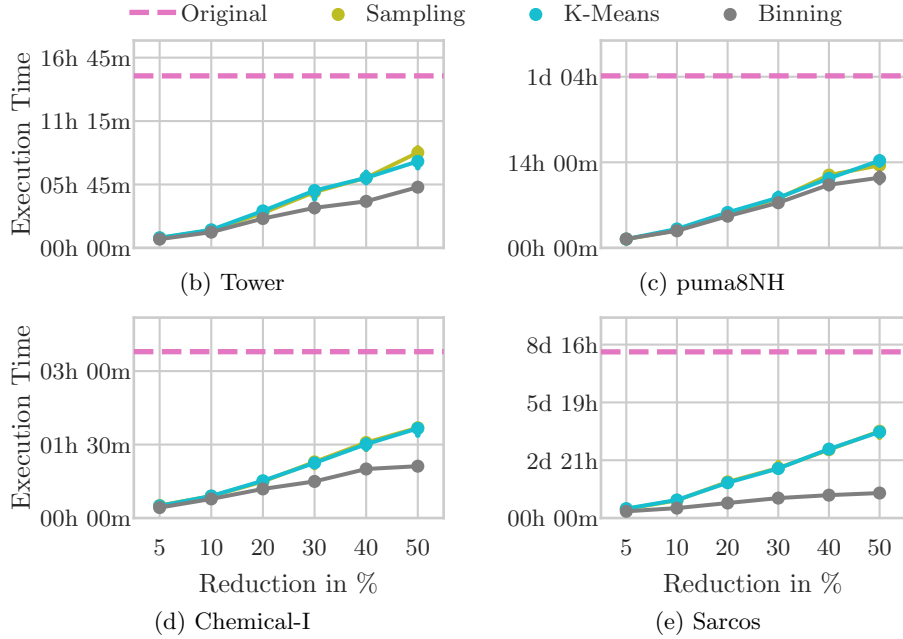
\begin{figure}[t]
    \raggedright
    \subfloat{%
        \input{images/plots/legend_runtime.pgf} 
    }\\[-10pt]

    \subfloat[Tower]{%
        \input{images/plots/exec_time_Tower.pgf}
    }\hfill
    \subfloat[puma8NH]{%
        \input{images/plots/exec_time_puma8NH.pgf} 
    }\\[-6pt]
    
    \subfloat[Chemical-I]{%
        \input{images/plots/exec_time_Chemical-I.pgf}
    }\hfill
    \subfloat[Sarcos]{
        \input{images/plots/exec_time_Sarcos.pgf} 
        \label{subfig:alg_accuracy_sarcos}
    }
    \caption{Median execution of GP for different rates of reduction of the original data set.}
    \label{fig:speedup}
\end{figure}

Figure \ref{fig:speedup} shows, that the speed-up of k-means clustering and sampling is proportional to the reduction rate -- how much the data has been reduced relative to the original size. This is expected, as GP spends most of its time evaluating individuals. The computational effort of evaluations increases proportionally with larger numbers of training instances, because every instance's prediction has to be calculated in every evaluation. However, the proportional speed-up also indicates, that algorithm dynamics such as the earlier convergence due to preprocessing can be ruled out. The runtime for the preprocessing step itself is neglected, because it made up at most three minutes (for the large Sarcos data set) per run, which is only a very small fraction of the runtime of GP. Figure \ref{fig:speedup} also excludes the runtime for RF and LR because both methods took only seconds to finish in all experiments.

While data binning and k-means led to similar execution times, GP runs with preceding data binning were slightly faster. This is most likely due to a loss of information about the relations in the data. Fewer relations in the training data make the search for a model, which fits the training data well, easier and therefore faster although it degrades its test errors. Table \ref{tab:osga_results} shows that data binning yielded throughout worse models regarding test accuracy.

\begin{table}[t]
    \centering
    \caption{Test $R^2$ median and interquartile range for symbolic regression with GP.}
    \input{images/tables/OSGA_results.tex}
    \label{tab:osga_results} 
\end{table}
 
K-means clustering and random sampling produced very similar results, as listed in Table \ref{tab:osga_results}. In both cases, the loss in prediction accuracy is small when the data is reduced to 30\%, 40\% or 50\% of the original data in comparison to runs with the original data. The only difference is the more stable behavior of k-means preprocessing when the training data is reduced to a size of 20\% or less of the original data set size. The small difference between preprocessing with k-means and data binning can be explained with the small number of instances, with which each centroid is computed -- e.g.~if the training data is reduced to 50\%, each centroid is the center of only two similar instances on average. This leaves little space for improvements over random sampling. However, all methods failed to yield meaningful models when the training data are reduced to only 1\% of their original size. 

\begin{figure}[t]
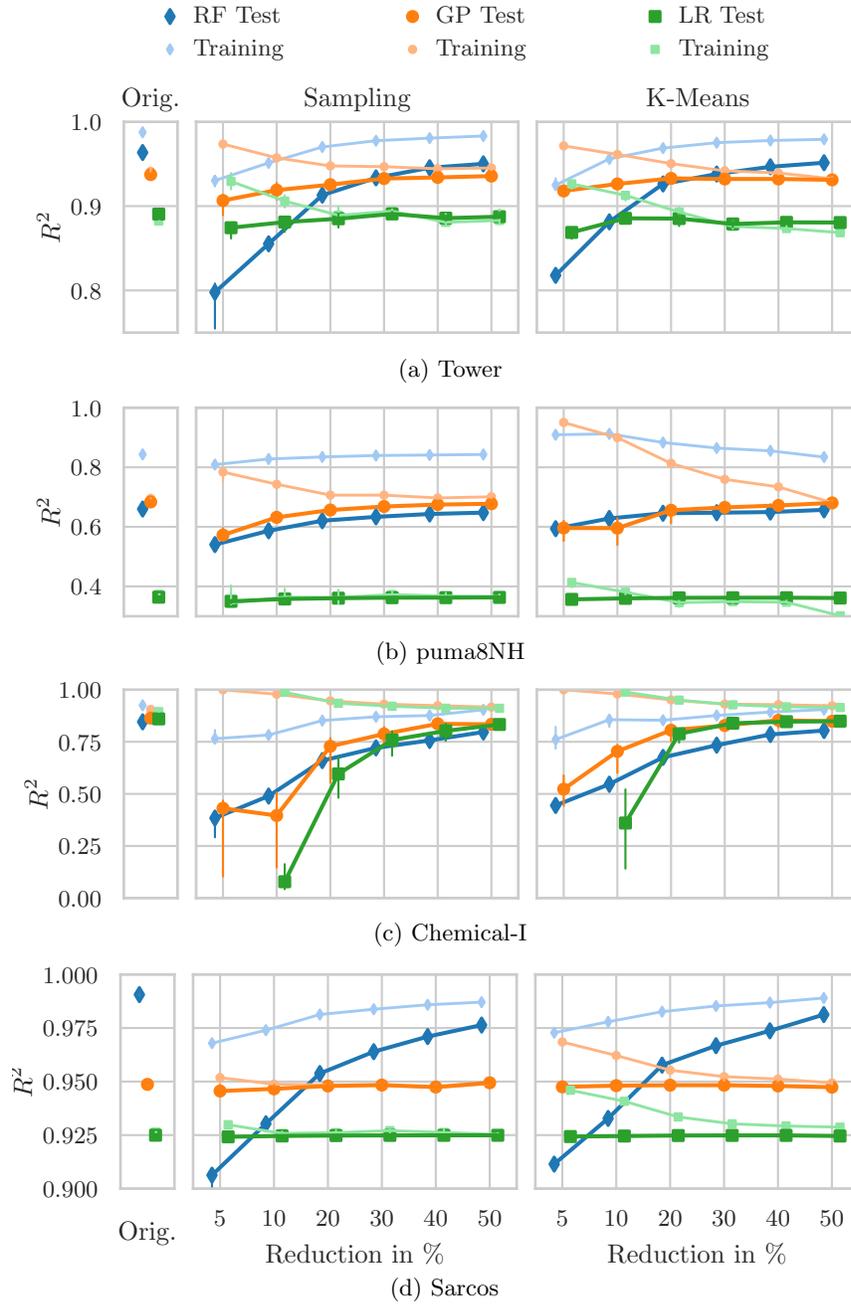

    \raggedright
    \hfill
    \subfloat{%
        \input{images/plots/legend_datasets_Sarcos.pgf} 
    }\vspace{-6pt}\\\addtocounter{subfigure}{-1}
    \hfill
    \subfloat[Tower]{%
        \input{images/plots/quality_Tower.pgf}
    }\vspace{-6pt}\\
    \hfill
    \subfloat[puma8NH]{%
        \input{images/plots/quality_puma8NH.pgf} 
    }\vspace{-6pt}\\
    \hfill
    \subfloat[Chemical-I]{%
    \input{images/plots/quality_Chemical-I.pgf}
    }\vspace{-6pt}\\
    \hfill
    \subfloat[Sarcos]{
        \input{images/plots/quality_Sarcos.pgf} 
    }
    \caption{Median $R^2$ values of the generated models.}
    \label{fig:alg_accuracy}
\end{figure}

When compared to RF and LR, GP achieves in most experiments similar accuracy as random forests. However, when the data is strongly reduced to only 30\% or less of the original data, GP tends to have less loss in test accuracy than RF.  Figure \ref{fig:alg_accuracy} describes the median $R^2$ results of all three algorithms with differently reduced data for each data set. Modeling results on the original data set are shown on the left. The results for data, that was reduced to 1\%, as well as the binning results are not shown in Figure \ref{fig:alg_accuracy} since they would degrade the axis scale.

\section{Conclusion}

The experimental results show, that the original training data can be reduced to only 30\% of the original size while still achieving only slightly worse test accuracy. While data binning led to unusable models regarding test accuracy, there is no noticeable difference in the resulting modeling accuracy between k-means and random sampling. Depending on the actual application, whether a higher loss in accuracy is acceptable, k-means might only be useful if the data is reduced to 5-20\%, as less variance among test errors compared to random sampling was observed in this range. Otherwise, it is more convenient to use random sampling instead of the additional effort of implementing k-means clustering in the modeling process. The impact of data reduction was smaller for GP than for RF, which underlines the stability and the generalization capabilities of symbolic regression.

\FloatBarrier

The speed-up for GP is proportional to the data reduction ratio, how much the original data was reduced to. While the proposed reduction methods might not be suitable for a final model in most applications due to the (even slight) loss in accuracy, random sampling as preprocessing step might be a suitable tool for speeding up early experimental phases and meta parameter tuning.

\subsubsection*{Acknowledgements.}
The authors gratefully acknowledge support by the Austrian Research Promotion Agency (FFG) within project \#867202, as well as the Christian Doppler Research Association and the Federal Ministry of Digital and Economic Affairs within the \emph{Josef Ressel Centre for Symbolic Regression}

%
%
%
\bibliographystyle{splncs04}
\bibliography{bibliography}

%
%
%
%
\end{document}

%% file: images/plots/legend_runtime.pgf
\begingroup%
\makeatletter%
\begin{pgfpicture}%
\pgfpathrectangle{\pgfpointorigin}{\pgfqpoint{5.000000in}{0.300000in}}%
\pgfusepath{use as bounding box, clip}%
\begin{pgfscope}%
\pgfsetbuttcap%
\pgfsetmiterjoin%
\pgfsetlinewidth{0.000000pt}%
\definecolor{currentstroke}{rgb}{0.000000,0.000000,0.000000}%
\pgfsetstrokecolor{currentstroke}%
\pgfsetstrokeopacity{0.000000}%
\pgfsetdash{}{0pt}%
\pgfpathmoveto{\pgfqpoint{0.000000in}{0.000000in}}%
\pgfpathlineto{\pgfqpoint{5.000000in}{0.000000in}}%
\pgfpathlineto{\pgfqpoint{5.000000in}{0.300000in}}%
\pgfpathlineto{\pgfqpoint{0.000000in}{0.300000in}}%
\pgfpathclose%
\pgfusepath{}%
\end{pgfscope}%
\begin{pgfscope}%
\pgfsetbuttcap%
\pgfsetroundjoin%
\pgfsetlinewidth{2.007500pt}%
\definecolor{currentstroke}{rgb}{0.890196,0.466667,0.760784}%
\pgfsetstrokecolor{currentstroke}%
\pgfsetdash{{7.400000pt}{3.200000pt}}{0.000000pt}%
\pgfpathmoveto{\pgfqpoint{0.658391in}{0.109972in}}%
\pgfpathlineto{\pgfqpoint{0.902836in}{0.109972in}}%
\pgfusepath{stroke}%
\end{pgfscope}%
\begin{pgfscope}%
\definecolor{textcolor}{rgb}{0.150000,0.150000,0.150000}%
\pgfsetstrokecolor{textcolor}%
\pgfsetfillcolor{textcolor}%
\pgftext[x=0.927280in,y=0.067194in,left,base]{\color{textcolor}\fontsize{8.800000}{10.560000}\selectfont Original}%
\end{pgfscope}%
\begin{pgfscope}%
\pgfsetbuttcap%
\pgfsetroundjoin%
\definecolor{currentfill}{rgb}{0.737255,0.741176,0.133333}%
\pgfsetfillcolor{currentfill}%
\pgfsetlinewidth{1.138252pt}%
\definecolor{currentstroke}{rgb}{0.737255,0.741176,0.133333}%
\pgfsetstrokecolor{currentstroke}%
\pgfsetdash{}{0pt}%
\pgfpathmoveto{\pgfqpoint{1.751023in}{0.072958in}}%
\pgfpathcurveto{\pgfqpoint{1.758003in}{0.072958in}}{\pgfqpoint{1.764698in}{0.075731in}}{\pgfqpoint{1.769634in}{0.080667in}}%
\pgfpathcurveto{\pgfqpoint{1.774569in}{0.085603in}}{\pgfqpoint{1.777343in}{0.092298in}}{\pgfqpoint{1.777343in}{0.099278in}}%
\pgfpathcurveto{\pgfqpoint{1.777343in}{0.106258in}}{\pgfqpoint{1.774569in}{0.112953in}}{\pgfqpoint{1.769634in}{0.117889in}}%
\pgfpathcurveto{\pgfqpoint{1.764698in}{0.122824in}}{\pgfqpoint{1.758003in}{0.125597in}}{\pgfqpoint{1.751023in}{0.125597in}}%
\pgfpathcurveto{\pgfqpoint{1.744043in}{0.125597in}}{\pgfqpoint{1.737348in}{0.122824in}}{\pgfqpoint{1.732412in}{0.117889in}}%
\pgfpathcurveto{\pgfqpoint{1.727477in}{0.112953in}}{\pgfqpoint{1.724703in}{0.106258in}}{\pgfqpoint{1.724703in}{0.099278in}}%
\pgfpathcurveto{\pgfqpoint{1.724703in}{0.092298in}}{\pgfqpoint{1.727477in}{0.085603in}}{\pgfqpoint{1.732412in}{0.080667in}}%
\pgfpathcurveto{\pgfqpoint{1.737348in}{0.075731in}}{\pgfqpoint{1.744043in}{0.072958in}}{\pgfqpoint{1.751023in}{0.072958in}}%
\pgfpathclose%
\pgfusepath{stroke,fill}%
\end{pgfscope}%
\begin{pgfscope}%
\definecolor{textcolor}{rgb}{0.150000,0.150000,0.150000}%
\pgfsetstrokecolor{textcolor}%
\pgfsetfillcolor{textcolor}%
\pgftext[x=1.897690in,y=0.067194in,left,base]{\color{textcolor}\fontsize{8.800000}{10.560000}\selectfont Sampling}%
\end{pgfscope}%
\begin{pgfscope}%
\pgfsetbuttcap%
\pgfsetroundjoin%
\definecolor{currentfill}{rgb}{0.090196,0.745098,0.811765}%
\pgfsetfillcolor{currentfill}%
\pgfsetlinewidth{1.138252pt}%
\definecolor{currentstroke}{rgb}{0.090196,0.745098,0.811765}%
\pgfsetstrokecolor{currentstroke}%
\pgfsetdash{}{0pt}%
\pgfpathmoveto{\pgfqpoint{2.785379in}{0.072958in}}%
\pgfpathcurveto{\pgfqpoint{2.792359in}{0.072958in}}{\pgfqpoint{2.799054in}{0.075731in}}{\pgfqpoint{2.803990in}{0.080667in}}%
\pgfpathcurveto{\pgfqpoint{2.808926in}{0.085603in}}{\pgfqpoint{2.811699in}{0.092298in}}{\pgfqpoint{2.811699in}{0.099278in}}%
\pgfpathcurveto{\pgfqpoint{2.811699in}{0.106258in}}{\pgfqpoint{2.808926in}{0.112953in}}{\pgfqpoint{2.803990in}{0.117889in}}%
\pgfpathcurveto{\pgfqpoint{2.799054in}{0.122824in}}{\pgfqpoint{2.792359in}{0.125597in}}{\pgfqpoint{2.785379in}{0.125597in}}%
\pgfpathcurveto{\pgfqpoint{2.778399in}{0.125597in}}{\pgfqpoint{2.771704in}{0.122824in}}{\pgfqpoint{2.766768in}{0.117889in}}%
\pgfpathcurveto{\pgfqpoint{2.761833in}{0.112953in}}{\pgfqpoint{2.759060in}{0.106258in}}{\pgfqpoint{2.759060in}{0.099278in}}%
\pgfpathcurveto{\pgfqpoint{2.759060in}{0.092298in}}{\pgfqpoint{2.761833in}{0.085603in}}{\pgfqpoint{2.766768in}{0.080667in}}%
\pgfpathcurveto{\pgfqpoint{2.771704in}{0.075731in}}{\pgfqpoint{2.778399in}{0.072958in}}{\pgfqpoint{2.785379in}{0.072958in}}%
\pgfpathclose%
\pgfusepath{stroke,fill}%
\end{pgfscope}%
\begin{pgfscope}%
\definecolor{textcolor}{rgb}{0.150000,0.150000,0.150000}%
\pgfsetstrokecolor{textcolor}%
\pgfsetfillcolor{textcolor}%
\pgftext[x=2.932046in,y=0.067194in,left,base]{\color{textcolor}\fontsize{8.800000}{10.560000}\selectfont K-Means}%
\end{pgfscope}%
\begin{pgfscope}%
\pgfsetbuttcap%
\pgfsetroundjoin%
\definecolor{currentfill}{rgb}{0.498039,0.498039,0.498039}%
\pgfsetfillcolor{currentfill}%
\pgfsetlinewidth{1.138252pt}%
\definecolor{currentstroke}{rgb}{0.498039,0.498039,0.498039}%
\pgfsetstrokecolor{currentstroke}%
\pgfsetdash{}{0pt}%
\pgfpathmoveto{\pgfqpoint{3.802615in}{0.072958in}}%
\pgfpathcurveto{\pgfqpoint{3.809595in}{0.072958in}}{\pgfqpoint{3.816290in}{0.075731in}}{\pgfqpoint{3.821226in}{0.080667in}}%
\pgfpathcurveto{\pgfqpoint{3.826161in}{0.085603in}}{\pgfqpoint{3.828935in}{0.092298in}}{\pgfqpoint{3.828935in}{0.099278in}}%
\pgfpathcurveto{\pgfqpoint{3.828935in}{0.106258in}}{\pgfqpoint{3.826161in}{0.112953in}}{\pgfqpoint{3.821226in}{0.117889in}}%
\pgfpathcurveto{\pgfqpoint{3.816290in}{0.122824in}}{\pgfqpoint{3.809595in}{0.125597in}}{\pgfqpoint{3.802615in}{0.125597in}}%
\pgfpathcurveto{\pgfqpoint{3.795635in}{0.125597in}}{\pgfqpoint{3.788940in}{0.122824in}}{\pgfqpoint{3.784004in}{0.117889in}}%
\pgfpathcurveto{\pgfqpoint{3.779069in}{0.112953in}}{\pgfqpoint{3.776295in}{0.106258in}}{\pgfqpoint{3.776295in}{0.099278in}}%
\pgfpathcurveto{\pgfqpoint{3.776295in}{0.092298in}}{\pgfqpoint{3.779069in}{0.085603in}}{\pgfqpoint{3.784004in}{0.080667in}}%
\pgfpathcurveto{\pgfqpoint{3.788940in}{0.075731in}}{\pgfqpoint{3.795635in}{0.072958in}}{\pgfqpoint{3.802615in}{0.072958in}}%
\pgfpathclose%
\pgfusepath{stroke,fill}%
\end{pgfscope}%
\begin{pgfscope}%
\definecolor{textcolor}{rgb}{0.150000,0.150000,0.150000}%
\pgfsetstrokecolor{textcolor}%
\pgfsetfillcolor{textcolor}%
\pgftext[x=3.949282in,y=0.067194in,left,base]{\color{textcolor}\fontsize{8.800000}{10.560000}\selectfont Binning}%
\end{pgfscope}%
\end{pgfpicture}%
\makeatother%
\endgroup%

%% file: images/plots/exec_time_Tower.pgf
\begingroup%
\makeatletter%
\begin{pgfpicture}%
\pgfpathrectangle{\pgfpointorigin}{\pgfqpoint{2.350000in}{1.168000in}}%
\pgfusepath{use as bounding box, clip}%
\begin{pgfscope}%
\pgfsetbuttcap%
\pgfsetmiterjoin%
\pgfsetlinewidth{0.000000pt}%
\definecolor{currentstroke}{rgb}{0.000000,0.000000,0.000000}%
\pgfsetstrokecolor{currentstroke}%
\pgfsetstrokeopacity{0.000000}%
\pgfsetdash{}{0pt}%
\pgfpathmoveto{\pgfqpoint{0.000000in}{0.000000in}}%
\pgfpathlineto{\pgfqpoint{2.350000in}{0.000000in}}%
\pgfpathlineto{\pgfqpoint{2.350000in}{1.168000in}}%
\pgfpathlineto{\pgfqpoint{0.000000in}{1.168000in}}%
\pgfpathclose%
\pgfusepath{}%
\end{pgfscope}%
\begin{pgfscope}%
\pgfsetbuttcap%
\pgfsetmiterjoin%
\pgfsetlinewidth{0.000000pt}%
\definecolor{currentstroke}{rgb}{0.000000,0.000000,0.000000}%
\pgfsetstrokecolor{currentstroke}%
\pgfsetstrokeopacity{0.000000}%
\pgfsetdash{}{0pt}%
\pgfpathmoveto{\pgfqpoint{0.681500in}{0.058400in}}%
\pgfpathlineto{\pgfqpoint{2.303000in}{0.058400in}}%
\pgfpathlineto{\pgfqpoint{2.303000in}{1.144640in}}%
\pgfpathlineto{\pgfqpoint{0.681500in}{1.144640in}}%
\pgfpathclose%
\pgfusepath{}%
\end{pgfscope}%
\begin{pgfscope}%
\pgfpathrectangle{\pgfqpoint{0.681500in}{0.058400in}}{\pgfqpoint{1.621500in}{1.086240in}} %
\pgfusepath{clip}%
\pgfsetroundcap%
\pgfsetroundjoin%
\pgfsetlinewidth{0.803000pt}%
\definecolor{currentstroke}{rgb}{0.800000,0.800000,0.800000}%
\pgfsetstrokecolor{currentstroke}%
\pgfsetdash{}{0pt}%
\pgfpathmoveto{\pgfqpoint{0.816625in}{0.058400in}}%
\pgfpathlineto{\pgfqpoint{0.816625in}{1.144640in}}%
\pgfusepath{stroke}%
\end{pgfscope}%
\begin{pgfscope}%
\pgfpathrectangle{\pgfqpoint{0.681500in}{0.058400in}}{\pgfqpoint{1.621500in}{1.086240in}} %
\pgfusepath{clip}%
\pgfsetroundcap%
\pgfsetroundjoin%
\pgfsetlinewidth{0.803000pt}%
\definecolor{currentstroke}{rgb}{0.800000,0.800000,0.800000}%
\pgfsetstrokecolor{currentstroke}%
\pgfsetdash{}{0pt}%
\pgfpathmoveto{\pgfqpoint{1.086875in}{0.058400in}}%
\pgfpathlineto{\pgfqpoint{1.086875in}{1.144640in}}%
\pgfusepath{stroke}%
\end{pgfscope}%
\begin{pgfscope}%
\pgfpathrectangle{\pgfqpoint{0.681500in}{0.058400in}}{\pgfqpoint{1.621500in}{1.086240in}} %
\pgfusepath{clip}%
\pgfsetroundcap%
\pgfsetroundjoin%
\pgfsetlinewidth{0.803000pt}%
\definecolor{currentstroke}{rgb}{0.800000,0.800000,0.800000}%
\pgfsetstrokecolor{currentstroke}%
\pgfsetdash{}{0pt}%
\pgfpathmoveto{\pgfqpoint{1.357125in}{0.058400in}}%
\pgfpathlineto{\pgfqpoint{1.357125in}{1.144640in}}%
\pgfusepath{stroke}%
\end{pgfscope}%
\begin{pgfscope}%
\pgfpathrectangle{\pgfqpoint{0.681500in}{0.058400in}}{\pgfqpoint{1.621500in}{1.086240in}} %
\pgfusepath{clip}%
\pgfsetroundcap%
\pgfsetroundjoin%
\pgfsetlinewidth{0.803000pt}%
\definecolor{currentstroke}{rgb}{0.800000,0.800000,0.800000}%
\pgfsetstrokecolor{currentstroke}%
\pgfsetdash{}{0pt}%
\pgfpathmoveto{\pgfqpoint{1.627375in}{0.058400in}}%
\pgfpathlineto{\pgfqpoint{1.627375in}{1.144640in}}%
\pgfusepath{stroke}%
\end{pgfscope}%
\begin{pgfscope}%
\pgfpathrectangle{\pgfqpoint{0.681500in}{0.058400in}}{\pgfqpoint{1.621500in}{1.086240in}} %
\pgfusepath{clip}%
\pgfsetroundcap%
\pgfsetroundjoin%
\pgfsetlinewidth{0.803000pt}%
\definecolor{currentstroke}{rgb}{0.800000,0.800000,0.800000}%
\pgfsetstrokecolor{currentstroke}%
\pgfsetdash{}{0pt}%
\pgfpathmoveto{\pgfqpoint{1.897625in}{0.058400in}}%
\pgfpathlineto{\pgfqpoint{1.897625in}{1.144640in}}%
\pgfusepath{stroke}%
\end{pgfscope}%
\begin{pgfscope}%
\pgfpathrectangle{\pgfqpoint{0.681500in}{0.058400in}}{\pgfqpoint{1.621500in}{1.086240in}} %
\pgfusepath{clip}%
\pgfsetroundcap%
\pgfsetroundjoin%
\pgfsetlinewidth{0.803000pt}%
\definecolor{currentstroke}{rgb}{0.800000,0.800000,0.800000}%
\pgfsetstrokecolor{currentstroke}%
\pgfsetdash{}{0pt}%
\pgfpathmoveto{\pgfqpoint{2.167875in}{0.058400in}}%
\pgfpathlineto{\pgfqpoint{2.167875in}{1.144640in}}%
\pgfusepath{stroke}%
\end{pgfscope}%
\begin{pgfscope}%
\pgfpathrectangle{\pgfqpoint{0.681500in}{0.058400in}}{\pgfqpoint{1.621500in}{1.086240in}} %
\pgfusepath{clip}%
\pgfsetroundcap%
\pgfsetroundjoin%
\pgfsetlinewidth{0.803000pt}%
\definecolor{currentstroke}{rgb}{0.800000,0.800000,0.800000}%
\pgfsetstrokecolor{currentstroke}%
\pgfsetdash{}{0pt}%
\pgfpathmoveto{\pgfqpoint{0.681500in}{0.058400in}}%
\pgfpathlineto{\pgfqpoint{2.303000in}{0.058400in}}%
\pgfusepath{stroke}%
\end{pgfscope}%
\begin{pgfscope}%
\definecolor{textcolor}{rgb}{0.150000,0.150000,0.150000}%
\pgfsetstrokecolor{textcolor}%
\pgfsetfillcolor{textcolor}%
\pgftext[x=0.192190in,y=0.014997in,left,base]{\color{textcolor}\fontsize{8.800000}{10.560000}\selectfont 00h 00m}%
\end{pgfscope}%
\begin{pgfscope}%
\pgfpathrectangle{\pgfqpoint{0.681500in}{0.058400in}}{\pgfqpoint{1.621500in}{1.086240in}} %
\pgfusepath{clip}%
\pgfsetroundcap%
\pgfsetroundjoin%
\pgfsetlinewidth{0.803000pt}%
\definecolor{currentstroke}{rgb}{0.800000,0.800000,0.800000}%
\pgfsetstrokecolor{currentstroke}%
\pgfsetdash{}{0pt}%
\pgfpathmoveto{\pgfqpoint{0.681500in}{0.390542in}}%
\pgfpathlineto{\pgfqpoint{2.303000in}{0.390542in}}%
\pgfusepath{stroke}%
\end{pgfscope}%
\begin{pgfscope}%
\definecolor{textcolor}{rgb}{0.150000,0.150000,0.150000}%
\pgfsetstrokecolor{textcolor}%
\pgfsetfillcolor{textcolor}%
\pgftext[x=0.192190in,y=0.347139in,left,base]{\color{textcolor}\fontsize{8.800000}{10.560000}\selectfont 05h 45m}%
\end{pgfscope}%
\begin{pgfscope}%
\pgfpathrectangle{\pgfqpoint{0.681500in}{0.058400in}}{\pgfqpoint{1.621500in}{1.086240in}} %
\pgfusepath{clip}%
\pgfsetroundcap%
\pgfsetroundjoin%
\pgfsetlinewidth{0.803000pt}%
\definecolor{currentstroke}{rgb}{0.800000,0.800000,0.800000}%
\pgfsetstrokecolor{currentstroke}%
\pgfsetdash{}{0pt}%
\pgfpathmoveto{\pgfqpoint{0.681500in}{0.722684in}}%
\pgfpathlineto{\pgfqpoint{2.303000in}{0.722684in}}%
\pgfusepath{stroke}%
\end{pgfscope}%
\begin{pgfscope}%
\definecolor{textcolor}{rgb}{0.150000,0.150000,0.150000}%
\pgfsetstrokecolor{textcolor}%
\pgfsetfillcolor{textcolor}%
\pgftext[x=0.192190in,y=0.679281in,left,base]{\color{textcolor}\fontsize{8.800000}{10.560000}\selectfont 11h 15m}%
\end{pgfscope}%
\begin{pgfscope}%
\pgfpathrectangle{\pgfqpoint{0.681500in}{0.058400in}}{\pgfqpoint{1.621500in}{1.086240in}} %
\pgfusepath{clip}%
\pgfsetroundcap%
\pgfsetroundjoin%
\pgfsetlinewidth{0.803000pt}%
\definecolor{currentstroke}{rgb}{0.800000,0.800000,0.800000}%
\pgfsetstrokecolor{currentstroke}%
\pgfsetdash{}{0pt}%
\pgfpathmoveto{\pgfqpoint{0.681500in}{1.054825in}}%
\pgfpathlineto{\pgfqpoint{2.303000in}{1.054825in}}%
\pgfusepath{stroke}%
\end{pgfscope}%
\begin{pgfscope}%
\definecolor{textcolor}{rgb}{0.150000,0.150000,0.150000}%
\pgfsetstrokecolor{textcolor}%
\pgfsetfillcolor{textcolor}%
\pgftext[x=0.192190in,y=1.011422in,left,base]{\color{textcolor}\fontsize{8.800000}{10.560000}\selectfont 16h 45m}%
\end{pgfscope}%
\begin{pgfscope}%
\definecolor{textcolor}{rgb}{0.150000,0.150000,0.150000}%
\pgfsetstrokecolor{textcolor}%
\pgfsetfillcolor{textcolor}%
\pgftext[x=0.136634in,y=0.601520in,,bottom,rotate=90.000000]{\color{textcolor}\fontsize{9.600000}{11.520000}\selectfont Execution Time}%
\end{pgfscope}%
\begin{pgfscope}%
\pgfpathrectangle{\pgfqpoint{0.681500in}{0.058400in}}{\pgfqpoint{1.621500in}{1.086240in}} %
\pgfusepath{clip}%
\pgfsetbuttcap%
\pgfsetroundjoin%
\definecolor{currentfill}{rgb}{0.737255,0.741176,0.133333}%
\pgfsetfillcolor{currentfill}%
\pgfsetlinewidth{1.138252pt}%
\definecolor{currentstroke}{rgb}{0.737255,0.741176,0.133333}%
\pgfsetstrokecolor{currentstroke}%
\pgfsetdash{}{0pt}%
\pgfpathmoveto{\pgfqpoint{0.816625in}{0.083789in}}%
\pgfpathcurveto{\pgfqpoint{0.823605in}{0.083789in}}{\pgfqpoint{0.830300in}{0.086562in}}{\pgfqpoint{0.835236in}{0.091498in}}%
\pgfpathcurveto{\pgfqpoint{0.840171in}{0.096434in}}{\pgfqpoint{0.842945in}{0.103129in}}{\pgfqpoint{0.842945in}{0.110109in}}%
\pgfpathcurveto{\pgfqpoint{0.842945in}{0.117089in}}{\pgfqpoint{0.840171in}{0.123784in}}{\pgfqpoint{0.835236in}{0.128720in}}%
\pgfpathcurveto{\pgfqpoint{0.830300in}{0.133655in}}{\pgfqpoint{0.823605in}{0.136428in}}{\pgfqpoint{0.816625in}{0.136428in}}%
\pgfpathcurveto{\pgfqpoint{0.809645in}{0.136428in}}{\pgfqpoint{0.802950in}{0.133655in}}{\pgfqpoint{0.798014in}{0.128720in}}%
\pgfpathcurveto{\pgfqpoint{0.793079in}{0.123784in}}{\pgfqpoint{0.790305in}{0.117089in}}{\pgfqpoint{0.790305in}{0.110109in}}%
\pgfpathcurveto{\pgfqpoint{0.790305in}{0.103129in}}{\pgfqpoint{0.793079in}{0.096434in}}{\pgfqpoint{0.798014in}{0.091498in}}%
\pgfpathcurveto{\pgfqpoint{0.802950in}{0.086562in}}{\pgfqpoint{0.809645in}{0.083789in}}{\pgfqpoint{0.816625in}{0.083789in}}%
\pgfpathclose%
\pgfusepath{stroke,fill}%
\end{pgfscope}%
\begin{pgfscope}%
\pgfpathrectangle{\pgfqpoint{0.681500in}{0.058400in}}{\pgfqpoint{1.621500in}{1.086240in}} %
\pgfusepath{clip}%
\pgfsetbuttcap%
\pgfsetroundjoin%
\definecolor{currentfill}{rgb}{0.737255,0.741176,0.133333}%
\pgfsetfillcolor{currentfill}%
\pgfsetlinewidth{1.138252pt}%
\definecolor{currentstroke}{rgb}{0.737255,0.741176,0.133333}%
\pgfsetstrokecolor{currentstroke}%
\pgfsetdash{}{0pt}%
\pgfpathmoveto{\pgfqpoint{1.086875in}{0.124707in}}%
\pgfpathcurveto{\pgfqpoint{1.093855in}{0.124707in}}{\pgfqpoint{1.100550in}{0.127480in}}{\pgfqpoint{1.105486in}{0.132416in}}%
\pgfpathcurveto{\pgfqpoint{1.110421in}{0.137351in}}{\pgfqpoint{1.113195in}{0.144047in}}{\pgfqpoint{1.113195in}{0.151027in}}%
\pgfpathcurveto{\pgfqpoint{1.113195in}{0.158007in}}{\pgfqpoint{1.110421in}{0.164702in}}{\pgfqpoint{1.105486in}{0.169637in}}%
\pgfpathcurveto{\pgfqpoint{1.100550in}{0.174573in}}{\pgfqpoint{1.093855in}{0.177346in}}{\pgfqpoint{1.086875in}{0.177346in}}%
\pgfpathcurveto{\pgfqpoint{1.079895in}{0.177346in}}{\pgfqpoint{1.073200in}{0.174573in}}{\pgfqpoint{1.068264in}{0.169637in}}%
\pgfpathcurveto{\pgfqpoint{1.063329in}{0.164702in}}{\pgfqpoint{1.060555in}{0.158007in}}{\pgfqpoint{1.060555in}{0.151027in}}%
\pgfpathcurveto{\pgfqpoint{1.060555in}{0.144047in}}{\pgfqpoint{1.063329in}{0.137351in}}{\pgfqpoint{1.068264in}{0.132416in}}%
\pgfpathcurveto{\pgfqpoint{1.073200in}{0.127480in}}{\pgfqpoint{1.079895in}{0.124707in}}{\pgfqpoint{1.086875in}{0.124707in}}%
\pgfpathclose%
\pgfusepath{stroke,fill}%
\end{pgfscope}%
\begin{pgfscope}%
\pgfpathrectangle{\pgfqpoint{0.681500in}{0.058400in}}{\pgfqpoint{1.621500in}{1.086240in}} %
\pgfusepath{clip}%
\pgfsetbuttcap%
\pgfsetroundjoin%
\definecolor{currentfill}{rgb}{0.737255,0.741176,0.133333}%
\pgfsetfillcolor{currentfill}%
\pgfsetlinewidth{1.138252pt}%
\definecolor{currentstroke}{rgb}{0.737255,0.741176,0.133333}%
\pgfsetstrokecolor{currentstroke}%
\pgfsetdash{}{0pt}%
\pgfpathmoveto{\pgfqpoint{1.357125in}{0.214578in}}%
\pgfpathcurveto{\pgfqpoint{1.364105in}{0.214578in}}{\pgfqpoint{1.370800in}{0.217351in}}{\pgfqpoint{1.375736in}{0.222287in}}%
\pgfpathcurveto{\pgfqpoint{1.380671in}{0.227223in}}{\pgfqpoint{1.383445in}{0.233918in}}{\pgfqpoint{1.383445in}{0.240898in}}%
\pgfpathcurveto{\pgfqpoint{1.383445in}{0.247878in}}{\pgfqpoint{1.380671in}{0.254573in}}{\pgfqpoint{1.375736in}{0.259509in}}%
\pgfpathcurveto{\pgfqpoint{1.370800in}{0.264444in}}{\pgfqpoint{1.364105in}{0.267217in}}{\pgfqpoint{1.357125in}{0.267217in}}%
\pgfpathcurveto{\pgfqpoint{1.350145in}{0.267217in}}{\pgfqpoint{1.343450in}{0.264444in}}{\pgfqpoint{1.338514in}{0.259509in}}%
\pgfpathcurveto{\pgfqpoint{1.333579in}{0.254573in}}{\pgfqpoint{1.330805in}{0.247878in}}{\pgfqpoint{1.330805in}{0.240898in}}%
\pgfpathcurveto{\pgfqpoint{1.330805in}{0.233918in}}{\pgfqpoint{1.333579in}{0.227223in}}{\pgfqpoint{1.338514in}{0.222287in}}%
\pgfpathcurveto{\pgfqpoint{1.343450in}{0.217351in}}{\pgfqpoint{1.350145in}{0.214578in}}{\pgfqpoint{1.357125in}{0.214578in}}%
\pgfpathclose%
\pgfusepath{stroke,fill}%
\end{pgfscope}%
\begin{pgfscope}%
\pgfpathrectangle{\pgfqpoint{0.681500in}{0.058400in}}{\pgfqpoint{1.621500in}{1.086240in}} %
\pgfusepath{clip}%
\pgfsetbuttcap%
\pgfsetroundjoin%
\definecolor{currentfill}{rgb}{0.737255,0.741176,0.133333}%
\pgfsetfillcolor{currentfill}%
\pgfsetlinewidth{1.138252pt}%
\definecolor{currentstroke}{rgb}{0.737255,0.741176,0.133333}%
\pgfsetstrokecolor{currentstroke}%
\pgfsetdash{}{0pt}%
\pgfpathmoveto{\pgfqpoint{1.627375in}{0.324296in}}%
\pgfpathcurveto{\pgfqpoint{1.634355in}{0.324296in}}{\pgfqpoint{1.641050in}{0.327070in}}{\pgfqpoint{1.645986in}{0.332005in}}%
\pgfpathcurveto{\pgfqpoint{1.650921in}{0.336941in}}{\pgfqpoint{1.653695in}{0.343636in}}{\pgfqpoint{1.653695in}{0.350616in}}%
\pgfpathcurveto{\pgfqpoint{1.653695in}{0.357596in}}{\pgfqpoint{1.650921in}{0.364291in}}{\pgfqpoint{1.645986in}{0.369227in}}%
\pgfpathcurveto{\pgfqpoint{1.641050in}{0.374162in}}{\pgfqpoint{1.634355in}{0.376936in}}{\pgfqpoint{1.627375in}{0.376936in}}%
\pgfpathcurveto{\pgfqpoint{1.620395in}{0.376936in}}{\pgfqpoint{1.613700in}{0.374162in}}{\pgfqpoint{1.608764in}{0.369227in}}%
\pgfpathcurveto{\pgfqpoint{1.603829in}{0.364291in}}{\pgfqpoint{1.601055in}{0.357596in}}{\pgfqpoint{1.601055in}{0.350616in}}%
\pgfpathcurveto{\pgfqpoint{1.601055in}{0.343636in}}{\pgfqpoint{1.603829in}{0.336941in}}{\pgfqpoint{1.608764in}{0.332005in}}%
\pgfpathcurveto{\pgfqpoint{1.613700in}{0.327070in}}{\pgfqpoint{1.620395in}{0.324296in}}{\pgfqpoint{1.627375in}{0.324296in}}%
\pgfpathclose%
\pgfusepath{stroke,fill}%
\end{pgfscope}%
\begin{pgfscope}%
\pgfpathrectangle{\pgfqpoint{0.681500in}{0.058400in}}{\pgfqpoint{1.621500in}{1.086240in}} %
\pgfusepath{clip}%
\pgfsetbuttcap%
\pgfsetroundjoin%
\definecolor{currentfill}{rgb}{0.737255,0.741176,0.133333}%
\pgfsetfillcolor{currentfill}%
\pgfsetlinewidth{1.138252pt}%
\definecolor{currentstroke}{rgb}{0.737255,0.741176,0.133333}%
\pgfsetstrokecolor{currentstroke}%
\pgfsetdash{}{0pt}%
\pgfpathmoveto{\pgfqpoint{1.897625in}{0.399524in}}%
\pgfpathcurveto{\pgfqpoint{1.904605in}{0.399524in}}{\pgfqpoint{1.911300in}{0.402297in}}{\pgfqpoint{1.916236in}{0.407233in}}%
\pgfpathcurveto{\pgfqpoint{1.921171in}{0.412169in}}{\pgfqpoint{1.923945in}{0.418864in}}{\pgfqpoint{1.923945in}{0.425844in}}%
\pgfpathcurveto{\pgfqpoint{1.923945in}{0.432824in}}{\pgfqpoint{1.921171in}{0.439519in}}{\pgfqpoint{1.916236in}{0.444455in}}%
\pgfpathcurveto{\pgfqpoint{1.911300in}{0.449390in}}{\pgfqpoint{1.904605in}{0.452163in}}{\pgfqpoint{1.897625in}{0.452163in}}%
\pgfpathcurveto{\pgfqpoint{1.890645in}{0.452163in}}{\pgfqpoint{1.883950in}{0.449390in}}{\pgfqpoint{1.879014in}{0.444455in}}%
\pgfpathcurveto{\pgfqpoint{1.874079in}{0.439519in}}{\pgfqpoint{1.871305in}{0.432824in}}{\pgfqpoint{1.871305in}{0.425844in}}%
\pgfpathcurveto{\pgfqpoint{1.871305in}{0.418864in}}{\pgfqpoint{1.874079in}{0.412169in}}{\pgfqpoint{1.879014in}{0.407233in}}%
\pgfpathcurveto{\pgfqpoint{1.883950in}{0.402297in}}{\pgfqpoint{1.890645in}{0.399524in}}{\pgfqpoint{1.897625in}{0.399524in}}%
\pgfpathclose%
\pgfusepath{stroke,fill}%
\end{pgfscope}%
\begin{pgfscope}%
\pgfpathrectangle{\pgfqpoint{0.681500in}{0.058400in}}{\pgfqpoint{1.621500in}{1.086240in}} %
\pgfusepath{clip}%
\pgfsetbuttcap%
\pgfsetroundjoin%
\definecolor{currentfill}{rgb}{0.737255,0.741176,0.133333}%
\pgfsetfillcolor{currentfill}%
\pgfsetlinewidth{1.138252pt}%
\definecolor{currentstroke}{rgb}{0.737255,0.741176,0.133333}%
\pgfsetstrokecolor{currentstroke}%
\pgfsetdash{}{0pt}%
\pgfpathmoveto{\pgfqpoint{2.167875in}{0.531409in}}%
\pgfpathcurveto{\pgfqpoint{2.174855in}{0.531409in}}{\pgfqpoint{2.181550in}{0.534182in}}{\pgfqpoint{2.186486in}{0.539118in}}%
\pgfpathcurveto{\pgfqpoint{2.191421in}{0.544054in}}{\pgfqpoint{2.194195in}{0.550749in}}{\pgfqpoint{2.194195in}{0.557729in}}%
\pgfpathcurveto{\pgfqpoint{2.194195in}{0.564709in}}{\pgfqpoint{2.191421in}{0.571404in}}{\pgfqpoint{2.186486in}{0.576339in}}%
\pgfpathcurveto{\pgfqpoint{2.181550in}{0.581275in}}{\pgfqpoint{2.174855in}{0.584048in}}{\pgfqpoint{2.167875in}{0.584048in}}%
\pgfpathcurveto{\pgfqpoint{2.160895in}{0.584048in}}{\pgfqpoint{2.154200in}{0.581275in}}{\pgfqpoint{2.149264in}{0.576339in}}%
\pgfpathcurveto{\pgfqpoint{2.144329in}{0.571404in}}{\pgfqpoint{2.141555in}{0.564709in}}{\pgfqpoint{2.141555in}{0.557729in}}%
\pgfpathcurveto{\pgfqpoint{2.141555in}{0.550749in}}{\pgfqpoint{2.144329in}{0.544054in}}{\pgfqpoint{2.149264in}{0.539118in}}%
\pgfpathcurveto{\pgfqpoint{2.154200in}{0.534182in}}{\pgfqpoint{2.160895in}{0.531409in}}{\pgfqpoint{2.167875in}{0.531409in}}%
\pgfpathclose%
\pgfusepath{stroke,fill}%
\end{pgfscope}%
\begin{pgfscope}%
\pgfpathrectangle{\pgfqpoint{0.681500in}{0.058400in}}{\pgfqpoint{1.621500in}{1.086240in}} %
\pgfusepath{clip}%
\pgfsetroundcap%
\pgfsetroundjoin%
\pgfsetlinewidth{1.517670pt}%
\definecolor{currentstroke}{rgb}{0.737255,0.741176,0.133333}%
\pgfsetstrokecolor{currentstroke}%
\pgfsetdash{}{0pt}%
\pgfpathmoveto{\pgfqpoint{0.816625in}{0.110109in}}%
\pgfpathlineto{\pgfqpoint{1.086875in}{0.151027in}}%
\pgfpathlineto{\pgfqpoint{1.357125in}{0.240898in}}%
\pgfpathlineto{\pgfqpoint{1.627375in}{0.350616in}}%
\pgfpathlineto{\pgfqpoint{1.897625in}{0.425844in}}%
\pgfpathlineto{\pgfqpoint{2.167875in}{0.557729in}}%
\pgfusepath{stroke}%
\end{pgfscope}%
\begin{pgfscope}%
\pgfpathrectangle{\pgfqpoint{0.681500in}{0.058400in}}{\pgfqpoint{1.621500in}{1.086240in}} %
\pgfusepath{clip}%
\pgfsetroundcap%
\pgfsetroundjoin%
\pgfsetlinewidth{2.168100pt}%
\definecolor{currentstroke}{rgb}{0.737255,0.741176,0.133333}%
\pgfsetstrokecolor{currentstroke}%
\pgfsetdash{}{0pt}%
\pgfpathmoveto{\pgfqpoint{0.816625in}{0.106964in}}%
\pgfpathlineto{\pgfqpoint{0.816625in}{0.111558in}}%
\pgfusepath{stroke}%
\end{pgfscope}%
\begin{pgfscope}%
\pgfpathrectangle{\pgfqpoint{0.681500in}{0.058400in}}{\pgfqpoint{1.621500in}{1.086240in}} %
\pgfusepath{clip}%
\pgfsetroundcap%
\pgfsetroundjoin%
\pgfsetlinewidth{2.168100pt}%
\definecolor{currentstroke}{rgb}{0.737255,0.741176,0.133333}%
\pgfsetstrokecolor{currentstroke}%
\pgfsetdash{}{0pt}%
\pgfpathmoveto{\pgfqpoint{1.086875in}{0.147053in}}%
\pgfpathlineto{\pgfqpoint{1.086875in}{0.155204in}}%
\pgfusepath{stroke}%
\end{pgfscope}%
\begin{pgfscope}%
\pgfpathrectangle{\pgfqpoint{0.681500in}{0.058400in}}{\pgfqpoint{1.621500in}{1.086240in}} %
\pgfusepath{clip}%
\pgfsetroundcap%
\pgfsetroundjoin%
\pgfsetlinewidth{2.168100pt}%
\definecolor{currentstroke}{rgb}{0.737255,0.741176,0.133333}%
\pgfsetstrokecolor{currentstroke}%
\pgfsetdash{}{0pt}%
\pgfpathmoveto{\pgfqpoint{1.357125in}{0.237075in}}%
\pgfpathlineto{\pgfqpoint{1.357125in}{0.248260in}}%
\pgfusepath{stroke}%
\end{pgfscope}%
\begin{pgfscope}%
\pgfpathrectangle{\pgfqpoint{0.681500in}{0.058400in}}{\pgfqpoint{1.621500in}{1.086240in}} %
\pgfusepath{clip}%
\pgfsetroundcap%
\pgfsetroundjoin%
\pgfsetlinewidth{2.168100pt}%
\definecolor{currentstroke}{rgb}{0.737255,0.741176,0.133333}%
\pgfsetstrokecolor{currentstroke}%
\pgfsetdash{}{0pt}%
\pgfpathmoveto{\pgfqpoint{1.627375in}{0.330023in}}%
\pgfpathlineto{\pgfqpoint{1.627375in}{0.373760in}}%
\pgfusepath{stroke}%
\end{pgfscope}%
\begin{pgfscope}%
\pgfpathrectangle{\pgfqpoint{0.681500in}{0.058400in}}{\pgfqpoint{1.621500in}{1.086240in}} %
\pgfusepath{clip}%
\pgfsetroundcap%
\pgfsetroundjoin%
\pgfsetlinewidth{2.168100pt}%
\definecolor{currentstroke}{rgb}{0.737255,0.741176,0.133333}%
\pgfsetstrokecolor{currentstroke}%
\pgfsetdash{}{0pt}%
\pgfpathmoveto{\pgfqpoint{1.897625in}{0.417705in}}%
\pgfpathlineto{\pgfqpoint{1.897625in}{0.449386in}}%
\pgfusepath{stroke}%
\end{pgfscope}%
\begin{pgfscope}%
\pgfpathrectangle{\pgfqpoint{0.681500in}{0.058400in}}{\pgfqpoint{1.621500in}{1.086240in}} %
\pgfusepath{clip}%
\pgfsetroundcap%
\pgfsetroundjoin%
\pgfsetlinewidth{2.168100pt}%
\definecolor{currentstroke}{rgb}{0.737255,0.741176,0.133333}%
\pgfsetstrokecolor{currentstroke}%
\pgfsetdash{}{0pt}%
\pgfpathmoveto{\pgfqpoint{2.167875in}{0.525043in}}%
\pgfpathlineto{\pgfqpoint{2.167875in}{0.582647in}}%
\pgfusepath{stroke}%
\end{pgfscope}%
\begin{pgfscope}%
\pgfpathrectangle{\pgfqpoint{0.681500in}{0.058400in}}{\pgfqpoint{1.621500in}{1.086240in}} %
\pgfusepath{clip}%
\pgfsetbuttcap%
\pgfsetroundjoin%
\definecolor{currentfill}{rgb}{0.090196,0.745098,0.811765}%
\pgfsetfillcolor{currentfill}%
\pgfsetlinewidth{1.138252pt}%
\definecolor{currentstroke}{rgb}{0.090196,0.745098,0.811765}%
\pgfsetstrokecolor{currentstroke}%
\pgfsetdash{}{0pt}%
\pgfpathmoveto{\pgfqpoint{0.816625in}{0.085324in}}%
\pgfpathcurveto{\pgfqpoint{0.823605in}{0.085324in}}{\pgfqpoint{0.830300in}{0.088098in}}{\pgfqpoint{0.835236in}{0.093033in}}%
\pgfpathcurveto{\pgfqpoint{0.840171in}{0.097969in}}{\pgfqpoint{0.842945in}{0.104664in}}{\pgfqpoint{0.842945in}{0.111644in}}%
\pgfpathcurveto{\pgfqpoint{0.842945in}{0.118624in}}{\pgfqpoint{0.840171in}{0.125319in}}{\pgfqpoint{0.835236in}{0.130255in}}%
\pgfpathcurveto{\pgfqpoint{0.830300in}{0.135190in}}{\pgfqpoint{0.823605in}{0.137964in}}{\pgfqpoint{0.816625in}{0.137964in}}%
\pgfpathcurveto{\pgfqpoint{0.809645in}{0.137964in}}{\pgfqpoint{0.802950in}{0.135190in}}{\pgfqpoint{0.798014in}{0.130255in}}%
\pgfpathcurveto{\pgfqpoint{0.793079in}{0.125319in}}{\pgfqpoint{0.790305in}{0.118624in}}{\pgfqpoint{0.790305in}{0.111644in}}%
\pgfpathcurveto{\pgfqpoint{0.790305in}{0.104664in}}{\pgfqpoint{0.793079in}{0.097969in}}{\pgfqpoint{0.798014in}{0.093033in}}%
\pgfpathcurveto{\pgfqpoint{0.802950in}{0.088098in}}{\pgfqpoint{0.809645in}{0.085324in}}{\pgfqpoint{0.816625in}{0.085324in}}%
\pgfpathclose%
\pgfusepath{stroke,fill}%
\end{pgfscope}%
\begin{pgfscope}%
\pgfpathrectangle{\pgfqpoint{0.681500in}{0.058400in}}{\pgfqpoint{1.621500in}{1.086240in}} %
\pgfusepath{clip}%
\pgfsetbuttcap%
\pgfsetroundjoin%
\definecolor{currentfill}{rgb}{0.090196,0.745098,0.811765}%
\pgfsetfillcolor{currentfill}%
\pgfsetlinewidth{1.138252pt}%
\definecolor{currentstroke}{rgb}{0.090196,0.745098,0.811765}%
\pgfsetstrokecolor{currentstroke}%
\pgfsetdash{}{0pt}%
\pgfpathmoveto{\pgfqpoint{1.086875in}{0.126645in}}%
\pgfpathcurveto{\pgfqpoint{1.093855in}{0.126645in}}{\pgfqpoint{1.100550in}{0.129418in}}{\pgfqpoint{1.105486in}{0.134354in}}%
\pgfpathcurveto{\pgfqpoint{1.110421in}{0.139289in}}{\pgfqpoint{1.113195in}{0.145984in}}{\pgfqpoint{1.113195in}{0.152964in}}%
\pgfpathcurveto{\pgfqpoint{1.113195in}{0.159944in}}{\pgfqpoint{1.110421in}{0.166639in}}{\pgfqpoint{1.105486in}{0.171575in}}%
\pgfpathcurveto{\pgfqpoint{1.100550in}{0.176511in}}{\pgfqpoint{1.093855in}{0.179284in}}{\pgfqpoint{1.086875in}{0.179284in}}%
\pgfpathcurveto{\pgfqpoint{1.079895in}{0.179284in}}{\pgfqpoint{1.073200in}{0.176511in}}{\pgfqpoint{1.068264in}{0.171575in}}%
\pgfpathcurveto{\pgfqpoint{1.063329in}{0.166639in}}{\pgfqpoint{1.060555in}{0.159944in}}{\pgfqpoint{1.060555in}{0.152964in}}%
\pgfpathcurveto{\pgfqpoint{1.060555in}{0.145984in}}{\pgfqpoint{1.063329in}{0.139289in}}{\pgfqpoint{1.068264in}{0.134354in}}%
\pgfpathcurveto{\pgfqpoint{1.073200in}{0.129418in}}{\pgfqpoint{1.079895in}{0.126645in}}{\pgfqpoint{1.086875in}{0.126645in}}%
\pgfpathclose%
\pgfusepath{stroke,fill}%
\end{pgfscope}%
\begin{pgfscope}%
\pgfpathrectangle{\pgfqpoint{0.681500in}{0.058400in}}{\pgfqpoint{1.621500in}{1.086240in}} %
\pgfusepath{clip}%
\pgfsetbuttcap%
\pgfsetroundjoin%
\definecolor{currentfill}{rgb}{0.090196,0.745098,0.811765}%
\pgfsetfillcolor{currentfill}%
\pgfsetlinewidth{1.138252pt}%
\definecolor{currentstroke}{rgb}{0.090196,0.745098,0.811765}%
\pgfsetstrokecolor{currentstroke}%
\pgfsetdash{}{0pt}%
\pgfpathmoveto{\pgfqpoint{1.357125in}{0.225844in}}%
\pgfpathcurveto{\pgfqpoint{1.364105in}{0.225844in}}{\pgfqpoint{1.370800in}{0.228617in}}{\pgfqpoint{1.375736in}{0.233553in}}%
\pgfpathcurveto{\pgfqpoint{1.380671in}{0.238488in}}{\pgfqpoint{1.383445in}{0.245183in}}{\pgfqpoint{1.383445in}{0.252163in}}%
\pgfpathcurveto{\pgfqpoint{1.383445in}{0.259144in}}{\pgfqpoint{1.380671in}{0.265839in}}{\pgfqpoint{1.375736in}{0.270774in}}%
\pgfpathcurveto{\pgfqpoint{1.370800in}{0.275710in}}{\pgfqpoint{1.364105in}{0.278483in}}{\pgfqpoint{1.357125in}{0.278483in}}%
\pgfpathcurveto{\pgfqpoint{1.350145in}{0.278483in}}{\pgfqpoint{1.343450in}{0.275710in}}{\pgfqpoint{1.338514in}{0.270774in}}%
\pgfpathcurveto{\pgfqpoint{1.333579in}{0.265839in}}{\pgfqpoint{1.330805in}{0.259144in}}{\pgfqpoint{1.330805in}{0.252163in}}%
\pgfpathcurveto{\pgfqpoint{1.330805in}{0.245183in}}{\pgfqpoint{1.333579in}{0.238488in}}{\pgfqpoint{1.338514in}{0.233553in}}%
\pgfpathcurveto{\pgfqpoint{1.343450in}{0.228617in}}{\pgfqpoint{1.350145in}{0.225844in}}{\pgfqpoint{1.357125in}{0.225844in}}%
\pgfpathclose%
\pgfusepath{stroke,fill}%
\end{pgfscope}%
\begin{pgfscope}%
\pgfpathrectangle{\pgfqpoint{0.681500in}{0.058400in}}{\pgfqpoint{1.621500in}{1.086240in}} %
\pgfusepath{clip}%
\pgfsetbuttcap%
\pgfsetroundjoin%
\definecolor{currentfill}{rgb}{0.090196,0.745098,0.811765}%
\pgfsetfillcolor{currentfill}%
\pgfsetlinewidth{1.138252pt}%
\definecolor{currentstroke}{rgb}{0.090196,0.745098,0.811765}%
\pgfsetstrokecolor{currentstroke}%
\pgfsetdash{}{0pt}%
\pgfpathmoveto{\pgfqpoint{1.627375in}{0.332464in}}%
\pgfpathcurveto{\pgfqpoint{1.634355in}{0.332464in}}{\pgfqpoint{1.641050in}{0.335237in}}{\pgfqpoint{1.645986in}{0.340173in}}%
\pgfpathcurveto{\pgfqpoint{1.650921in}{0.345109in}}{\pgfqpoint{1.653695in}{0.351804in}}{\pgfqpoint{1.653695in}{0.358784in}}%
\pgfpathcurveto{\pgfqpoint{1.653695in}{0.365764in}}{\pgfqpoint{1.650921in}{0.372459in}}{\pgfqpoint{1.645986in}{0.377395in}}%
\pgfpathcurveto{\pgfqpoint{1.641050in}{0.382330in}}{\pgfqpoint{1.634355in}{0.385103in}}{\pgfqpoint{1.627375in}{0.385103in}}%
\pgfpathcurveto{\pgfqpoint{1.620395in}{0.385103in}}{\pgfqpoint{1.613700in}{0.382330in}}{\pgfqpoint{1.608764in}{0.377395in}}%
\pgfpathcurveto{\pgfqpoint{1.603829in}{0.372459in}}{\pgfqpoint{1.601055in}{0.365764in}}{\pgfqpoint{1.601055in}{0.358784in}}%
\pgfpathcurveto{\pgfqpoint{1.601055in}{0.351804in}}{\pgfqpoint{1.603829in}{0.345109in}}{\pgfqpoint{1.608764in}{0.340173in}}%
\pgfpathcurveto{\pgfqpoint{1.613700in}{0.335237in}}{\pgfqpoint{1.620395in}{0.332464in}}{\pgfqpoint{1.627375in}{0.332464in}}%
\pgfpathclose%
\pgfusepath{stroke,fill}%
\end{pgfscope}%
\begin{pgfscope}%
\pgfpathrectangle{\pgfqpoint{0.681500in}{0.058400in}}{\pgfqpoint{1.621500in}{1.086240in}} %
\pgfusepath{clip}%
\pgfsetbuttcap%
\pgfsetroundjoin%
\definecolor{currentfill}{rgb}{0.090196,0.745098,0.811765}%
\pgfsetfillcolor{currentfill}%
\pgfsetlinewidth{1.138252pt}%
\definecolor{currentstroke}{rgb}{0.090196,0.745098,0.811765}%
\pgfsetstrokecolor{currentstroke}%
\pgfsetdash{}{0pt}%
\pgfpathmoveto{\pgfqpoint{1.897625in}{0.398966in}}%
\pgfpathcurveto{\pgfqpoint{1.904605in}{0.398966in}}{\pgfqpoint{1.911300in}{0.401740in}}{\pgfqpoint{1.916236in}{0.406675in}}%
\pgfpathcurveto{\pgfqpoint{1.921171in}{0.411611in}}{\pgfqpoint{1.923945in}{0.418306in}}{\pgfqpoint{1.923945in}{0.425286in}}%
\pgfpathcurveto{\pgfqpoint{1.923945in}{0.432266in}}{\pgfqpoint{1.921171in}{0.438961in}}{\pgfqpoint{1.916236in}{0.443897in}}%
\pgfpathcurveto{\pgfqpoint{1.911300in}{0.448832in}}{\pgfqpoint{1.904605in}{0.451605in}}{\pgfqpoint{1.897625in}{0.451605in}}%
\pgfpathcurveto{\pgfqpoint{1.890645in}{0.451605in}}{\pgfqpoint{1.883950in}{0.448832in}}{\pgfqpoint{1.879014in}{0.443897in}}%
\pgfpathcurveto{\pgfqpoint{1.874079in}{0.438961in}}{\pgfqpoint{1.871305in}{0.432266in}}{\pgfqpoint{1.871305in}{0.425286in}}%
\pgfpathcurveto{\pgfqpoint{1.871305in}{0.418306in}}{\pgfqpoint{1.874079in}{0.411611in}}{\pgfqpoint{1.879014in}{0.406675in}}%
\pgfpathcurveto{\pgfqpoint{1.883950in}{0.401740in}}{\pgfqpoint{1.890645in}{0.398966in}}{\pgfqpoint{1.897625in}{0.398966in}}%
\pgfpathclose%
\pgfusepath{stroke,fill}%
\end{pgfscope}%
\begin{pgfscope}%
\pgfpathrectangle{\pgfqpoint{0.681500in}{0.058400in}}{\pgfqpoint{1.621500in}{1.086240in}} %
\pgfusepath{clip}%
\pgfsetbuttcap%
\pgfsetroundjoin%
\definecolor{currentfill}{rgb}{0.090196,0.745098,0.811765}%
\pgfsetfillcolor{currentfill}%
\pgfsetlinewidth{1.138252pt}%
\definecolor{currentstroke}{rgb}{0.090196,0.745098,0.811765}%
\pgfsetstrokecolor{currentstroke}%
\pgfsetdash{}{0pt}%
\pgfpathmoveto{\pgfqpoint{2.167875in}{0.485186in}}%
\pgfpathcurveto{\pgfqpoint{2.174855in}{0.485186in}}{\pgfqpoint{2.181550in}{0.487959in}}{\pgfqpoint{2.186486in}{0.492895in}}%
\pgfpathcurveto{\pgfqpoint{2.191421in}{0.497830in}}{\pgfqpoint{2.194195in}{0.504525in}}{\pgfqpoint{2.194195in}{0.511505in}}%
\pgfpathcurveto{\pgfqpoint{2.194195in}{0.518485in}}{\pgfqpoint{2.191421in}{0.525181in}}{\pgfqpoint{2.186486in}{0.530116in}}%
\pgfpathcurveto{\pgfqpoint{2.181550in}{0.535052in}}{\pgfqpoint{2.174855in}{0.537825in}}{\pgfqpoint{2.167875in}{0.537825in}}%
\pgfpathcurveto{\pgfqpoint{2.160895in}{0.537825in}}{\pgfqpoint{2.154200in}{0.535052in}}{\pgfqpoint{2.149264in}{0.530116in}}%
\pgfpathcurveto{\pgfqpoint{2.144329in}{0.525181in}}{\pgfqpoint{2.141555in}{0.518485in}}{\pgfqpoint{2.141555in}{0.511505in}}%
\pgfpathcurveto{\pgfqpoint{2.141555in}{0.504525in}}{\pgfqpoint{2.144329in}{0.497830in}}{\pgfqpoint{2.149264in}{0.492895in}}%
\pgfpathcurveto{\pgfqpoint{2.154200in}{0.487959in}}{\pgfqpoint{2.160895in}{0.485186in}}{\pgfqpoint{2.167875in}{0.485186in}}%
\pgfpathclose%
\pgfusepath{stroke,fill}%
\end{pgfscope}%
\begin{pgfscope}%
\pgfpathrectangle{\pgfqpoint{0.681500in}{0.058400in}}{\pgfqpoint{1.621500in}{1.086240in}} %
\pgfusepath{clip}%
\pgfsetroundcap%
\pgfsetroundjoin%
\pgfsetlinewidth{1.517670pt}%
\definecolor{currentstroke}{rgb}{0.090196,0.745098,0.811765}%
\pgfsetstrokecolor{currentstroke}%
\pgfsetdash{}{0pt}%
\pgfpathmoveto{\pgfqpoint{0.816625in}{0.111644in}}%
\pgfpathlineto{\pgfqpoint{1.086875in}{0.152964in}}%
\pgfpathlineto{\pgfqpoint{1.357125in}{0.252163in}}%
\pgfpathlineto{\pgfqpoint{1.627375in}{0.358784in}}%
\pgfpathlineto{\pgfqpoint{1.897625in}{0.425286in}}%
\pgfpathlineto{\pgfqpoint{2.167875in}{0.511505in}}%
\pgfusepath{stroke}%
\end{pgfscope}%
\begin{pgfscope}%
\pgfpathrectangle{\pgfqpoint{0.681500in}{0.058400in}}{\pgfqpoint{1.621500in}{1.086240in}} %
\pgfusepath{clip}%
\pgfsetroundcap%
\pgfsetroundjoin%
\pgfsetlinewidth{2.168100pt}%
\definecolor{currentstroke}{rgb}{0.090196,0.745098,0.811765}%
\pgfsetstrokecolor{currentstroke}%
\pgfsetdash{}{0pt}%
\pgfpathmoveto{\pgfqpoint{0.816625in}{0.108902in}}%
\pgfpathlineto{\pgfqpoint{0.816625in}{0.112764in}}%
\pgfusepath{stroke}%
\end{pgfscope}%
\begin{pgfscope}%
\pgfpathrectangle{\pgfqpoint{0.681500in}{0.058400in}}{\pgfqpoint{1.621500in}{1.086240in}} %
\pgfusepath{clip}%
\pgfsetroundcap%
\pgfsetroundjoin%
\pgfsetlinewidth{2.168100pt}%
\definecolor{currentstroke}{rgb}{0.090196,0.745098,0.811765}%
\pgfsetstrokecolor{currentstroke}%
\pgfsetdash{}{0pt}%
\pgfpathmoveto{\pgfqpoint{1.086875in}{0.148231in}}%
\pgfpathlineto{\pgfqpoint{1.086875in}{0.157036in}}%
\pgfusepath{stroke}%
\end{pgfscope}%
\begin{pgfscope}%
\pgfpathrectangle{\pgfqpoint{0.681500in}{0.058400in}}{\pgfqpoint{1.621500in}{1.086240in}} %
\pgfusepath{clip}%
\pgfsetroundcap%
\pgfsetroundjoin%
\pgfsetlinewidth{2.168100pt}%
\definecolor{currentstroke}{rgb}{0.090196,0.745098,0.811765}%
\pgfsetstrokecolor{currentstroke}%
\pgfsetdash{}{0pt}%
\pgfpathmoveto{\pgfqpoint{1.357125in}{0.242843in}}%
\pgfpathlineto{\pgfqpoint{1.357125in}{0.259768in}}%
\pgfusepath{stroke}%
\end{pgfscope}%
\begin{pgfscope}%
\pgfpathrectangle{\pgfqpoint{0.681500in}{0.058400in}}{\pgfqpoint{1.621500in}{1.086240in}} %
\pgfusepath{clip}%
\pgfsetroundcap%
\pgfsetroundjoin%
\pgfsetlinewidth{2.168100pt}%
\definecolor{currentstroke}{rgb}{0.090196,0.745098,0.811765}%
\pgfsetstrokecolor{currentstroke}%
\pgfsetdash{}{0pt}%
\pgfpathmoveto{\pgfqpoint{1.627375in}{0.315792in}}%
\pgfpathlineto{\pgfqpoint{1.627375in}{0.363204in}}%
\pgfusepath{stroke}%
\end{pgfscope}%
\begin{pgfscope}%
\pgfpathrectangle{\pgfqpoint{0.681500in}{0.058400in}}{\pgfqpoint{1.621500in}{1.086240in}} %
\pgfusepath{clip}%
\pgfsetroundcap%
\pgfsetroundjoin%
\pgfsetlinewidth{2.168100pt}%
\definecolor{currentstroke}{rgb}{0.090196,0.745098,0.811765}%
\pgfsetstrokecolor{currentstroke}%
\pgfsetdash{}{0pt}%
\pgfpathmoveto{\pgfqpoint{1.897625in}{0.399973in}}%
\pgfpathlineto{\pgfqpoint{1.897625in}{0.453817in}}%
\pgfusepath{stroke}%
\end{pgfscope}%
\begin{pgfscope}%
\pgfpathrectangle{\pgfqpoint{0.681500in}{0.058400in}}{\pgfqpoint{1.621500in}{1.086240in}} %
\pgfusepath{clip}%
\pgfsetroundcap%
\pgfsetroundjoin%
\pgfsetlinewidth{2.168100pt}%
\definecolor{currentstroke}{rgb}{0.090196,0.745098,0.811765}%
\pgfsetstrokecolor{currentstroke}%
\pgfsetdash{}{0pt}%
\pgfpathmoveto{\pgfqpoint{2.167875in}{0.478522in}}%
\pgfpathlineto{\pgfqpoint{2.167875in}{0.521547in}}%
\pgfusepath{stroke}%
\end{pgfscope}%
\begin{pgfscope}%
\pgfsetrectcap%
\pgfsetmiterjoin%
\pgfsetlinewidth{1.003750pt}%
\definecolor{currentstroke}{rgb}{0.800000,0.800000,0.800000}%
\pgfsetstrokecolor{currentstroke}%
\pgfsetdash{}{0pt}%
\pgfpathmoveto{\pgfqpoint{0.681500in}{0.058400in}}%
\pgfpathlineto{\pgfqpoint{0.681500in}{1.144640in}}%
\pgfusepath{stroke}%
\end{pgfscope}%
\begin{pgfscope}%
\pgfsetrectcap%
\pgfsetmiterjoin%
\pgfsetlinewidth{1.003750pt}%
\definecolor{currentstroke}{rgb}{0.800000,0.800000,0.800000}%
\pgfsetstrokecolor{currentstroke}%
\pgfsetdash{}{0pt}%
\pgfpathmoveto{\pgfqpoint{2.303000in}{0.058400in}}%
\pgfpathlineto{\pgfqpoint{2.303000in}{1.144640in}}%
\pgfusepath{stroke}%
\end{pgfscope}%
\begin{pgfscope}%
\pgfsetrectcap%
\pgfsetmiterjoin%
\pgfsetlinewidth{1.003750pt}%
\definecolor{currentstroke}{rgb}{0.800000,0.800000,0.800000}%
\pgfsetstrokecolor{currentstroke}%
\pgfsetdash{}{0pt}%
\pgfpathmoveto{\pgfqpoint{0.681500in}{0.058400in}}%
\pgfpathlineto{\pgfqpoint{2.303000in}{0.058400in}}%
\pgfusepath{stroke}%
\end{pgfscope}%
\begin{pgfscope}%
\pgfsetrectcap%
\pgfsetmiterjoin%
\pgfsetlinewidth{1.003750pt}%
\definecolor{currentstroke}{rgb}{0.800000,0.800000,0.800000}%
\pgfsetstrokecolor{currentstroke}%
\pgfsetdash{}{0pt}%
\pgfpathmoveto{\pgfqpoint{0.681500in}{1.144640in}}%
\pgfpathlineto{\pgfqpoint{2.303000in}{1.144640in}}%
\pgfusepath{stroke}%
\end{pgfscope}%
\begin{pgfscope}%
\pgfpathrectangle{\pgfqpoint{0.681500in}{0.058400in}}{\pgfqpoint{1.621500in}{1.086240in}} %
\pgfusepath{clip}%
\pgfsetbuttcap%
\pgfsetroundjoin%
\definecolor{currentfill}{rgb}{0.498039,0.498039,0.498039}%
\pgfsetfillcolor{currentfill}%
\pgfsetlinewidth{1.138252pt}%
\definecolor{currentstroke}{rgb}{0.498039,0.498039,0.498039}%
\pgfsetstrokecolor{currentstroke}%
\pgfsetdash{}{0pt}%
\pgfpathmoveto{\pgfqpoint{0.816625in}{0.077922in}}%
\pgfpathcurveto{\pgfqpoint{0.823605in}{0.077922in}}{\pgfqpoint{0.830300in}{0.080695in}}{\pgfqpoint{0.835236in}{0.085631in}}%
\pgfpathcurveto{\pgfqpoint{0.840171in}{0.090566in}}{\pgfqpoint{0.842945in}{0.097261in}}{\pgfqpoint{0.842945in}{0.104241in}}%
\pgfpathcurveto{\pgfqpoint{0.842945in}{0.111222in}}{\pgfqpoint{0.840171in}{0.117917in}}{\pgfqpoint{0.835236in}{0.122852in}}%
\pgfpathcurveto{\pgfqpoint{0.830300in}{0.127788in}}{\pgfqpoint{0.823605in}{0.130561in}}{\pgfqpoint{0.816625in}{0.130561in}}%
\pgfpathcurveto{\pgfqpoint{0.809645in}{0.130561in}}{\pgfqpoint{0.802950in}{0.127788in}}{\pgfqpoint{0.798014in}{0.122852in}}%
\pgfpathcurveto{\pgfqpoint{0.793079in}{0.117917in}}{\pgfqpoint{0.790305in}{0.111222in}}{\pgfqpoint{0.790305in}{0.104241in}}%
\pgfpathcurveto{\pgfqpoint{0.790305in}{0.097261in}}{\pgfqpoint{0.793079in}{0.090566in}}{\pgfqpoint{0.798014in}{0.085631in}}%
\pgfpathcurveto{\pgfqpoint{0.802950in}{0.080695in}}{\pgfqpoint{0.809645in}{0.077922in}}{\pgfqpoint{0.816625in}{0.077922in}}%
\pgfpathclose%
\pgfusepath{stroke,fill}%
\end{pgfscope}%
\begin{pgfscope}%
\pgfpathrectangle{\pgfqpoint{0.681500in}{0.058400in}}{\pgfqpoint{1.621500in}{1.086240in}} %
\pgfusepath{clip}%
\pgfsetbuttcap%
\pgfsetroundjoin%
\definecolor{currentfill}{rgb}{0.498039,0.498039,0.498039}%
\pgfsetfillcolor{currentfill}%
\pgfsetlinewidth{1.138252pt}%
\definecolor{currentstroke}{rgb}{0.498039,0.498039,0.498039}%
\pgfsetstrokecolor{currentstroke}%
\pgfsetdash{}{0pt}%
\pgfpathmoveto{\pgfqpoint{1.086875in}{0.114473in}}%
\pgfpathcurveto{\pgfqpoint{1.093855in}{0.114473in}}{\pgfqpoint{1.100550in}{0.117246in}}{\pgfqpoint{1.105486in}{0.122182in}}%
\pgfpathcurveto{\pgfqpoint{1.110421in}{0.127117in}}{\pgfqpoint{1.113195in}{0.133812in}}{\pgfqpoint{1.113195in}{0.140792in}}%
\pgfpathcurveto{\pgfqpoint{1.113195in}{0.147772in}}{\pgfqpoint{1.110421in}{0.154467in}}{\pgfqpoint{1.105486in}{0.159403in}}%
\pgfpathcurveto{\pgfqpoint{1.100550in}{0.164339in}}{\pgfqpoint{1.093855in}{0.167112in}}{\pgfqpoint{1.086875in}{0.167112in}}%
\pgfpathcurveto{\pgfqpoint{1.079895in}{0.167112in}}{\pgfqpoint{1.073200in}{0.164339in}}{\pgfqpoint{1.068264in}{0.159403in}}%
\pgfpathcurveto{\pgfqpoint{1.063329in}{0.154467in}}{\pgfqpoint{1.060555in}{0.147772in}}{\pgfqpoint{1.060555in}{0.140792in}}%
\pgfpathcurveto{\pgfqpoint{1.060555in}{0.133812in}}{\pgfqpoint{1.063329in}{0.127117in}}{\pgfqpoint{1.068264in}{0.122182in}}%
\pgfpathcurveto{\pgfqpoint{1.073200in}{0.117246in}}{\pgfqpoint{1.079895in}{0.114473in}}{\pgfqpoint{1.086875in}{0.114473in}}%
\pgfpathclose%
\pgfusepath{stroke,fill}%
\end{pgfscope}%
\begin{pgfscope}%
\pgfpathrectangle{\pgfqpoint{0.681500in}{0.058400in}}{\pgfqpoint{1.621500in}{1.086240in}} %
\pgfusepath{clip}%
\pgfsetbuttcap%
\pgfsetroundjoin%
\definecolor{currentfill}{rgb}{0.498039,0.498039,0.498039}%
\pgfsetfillcolor{currentfill}%
\pgfsetlinewidth{1.138252pt}%
\definecolor{currentstroke}{rgb}{0.498039,0.498039,0.498039}%
\pgfsetstrokecolor{currentstroke}%
\pgfsetdash{}{0pt}%
\pgfpathmoveto{\pgfqpoint{1.357125in}{0.185741in}}%
\pgfpathcurveto{\pgfqpoint{1.364105in}{0.185741in}}{\pgfqpoint{1.370800in}{0.188514in}}{\pgfqpoint{1.375736in}{0.193450in}}%
\pgfpathcurveto{\pgfqpoint{1.380671in}{0.198386in}}{\pgfqpoint{1.383445in}{0.205081in}}{\pgfqpoint{1.383445in}{0.212061in}}%
\pgfpathcurveto{\pgfqpoint{1.383445in}{0.219041in}}{\pgfqpoint{1.380671in}{0.225736in}}{\pgfqpoint{1.375736in}{0.230672in}}%
\pgfpathcurveto{\pgfqpoint{1.370800in}{0.235607in}}{\pgfqpoint{1.364105in}{0.238380in}}{\pgfqpoint{1.357125in}{0.238380in}}%
\pgfpathcurveto{\pgfqpoint{1.350145in}{0.238380in}}{\pgfqpoint{1.343450in}{0.235607in}}{\pgfqpoint{1.338514in}{0.230672in}}%
\pgfpathcurveto{\pgfqpoint{1.333579in}{0.225736in}}{\pgfqpoint{1.330805in}{0.219041in}}{\pgfqpoint{1.330805in}{0.212061in}}%
\pgfpathcurveto{\pgfqpoint{1.330805in}{0.205081in}}{\pgfqpoint{1.333579in}{0.198386in}}{\pgfqpoint{1.338514in}{0.193450in}}%
\pgfpathcurveto{\pgfqpoint{1.343450in}{0.188514in}}{\pgfqpoint{1.350145in}{0.185741in}}{\pgfqpoint{1.357125in}{0.185741in}}%
\pgfpathclose%
\pgfusepath{stroke,fill}%
\end{pgfscope}%
\begin{pgfscope}%
\pgfpathrectangle{\pgfqpoint{0.681500in}{0.058400in}}{\pgfqpoint{1.621500in}{1.086240in}} %
\pgfusepath{clip}%
\pgfsetbuttcap%
\pgfsetroundjoin%
\definecolor{currentfill}{rgb}{0.498039,0.498039,0.498039}%
\pgfsetfillcolor{currentfill}%
\pgfsetlinewidth{1.138252pt}%
\definecolor{currentstroke}{rgb}{0.498039,0.498039,0.498039}%
\pgfsetstrokecolor{currentstroke}%
\pgfsetdash{}{0pt}%
\pgfpathmoveto{\pgfqpoint{1.627375in}{0.241808in}}%
\pgfpathcurveto{\pgfqpoint{1.634355in}{0.241808in}}{\pgfqpoint{1.641050in}{0.244581in}}{\pgfqpoint{1.645986in}{0.249517in}}%
\pgfpathcurveto{\pgfqpoint{1.650921in}{0.254453in}}{\pgfqpoint{1.653695in}{0.261148in}}{\pgfqpoint{1.653695in}{0.268128in}}%
\pgfpathcurveto{\pgfqpoint{1.653695in}{0.275108in}}{\pgfqpoint{1.650921in}{0.281803in}}{\pgfqpoint{1.645986in}{0.286739in}}%
\pgfpathcurveto{\pgfqpoint{1.641050in}{0.291674in}}{\pgfqpoint{1.634355in}{0.294447in}}{\pgfqpoint{1.627375in}{0.294447in}}%
\pgfpathcurveto{\pgfqpoint{1.620395in}{0.294447in}}{\pgfqpoint{1.613700in}{0.291674in}}{\pgfqpoint{1.608764in}{0.286739in}}%
\pgfpathcurveto{\pgfqpoint{1.603829in}{0.281803in}}{\pgfqpoint{1.601055in}{0.275108in}}{\pgfqpoint{1.601055in}{0.268128in}}%
\pgfpathcurveto{\pgfqpoint{1.601055in}{0.261148in}}{\pgfqpoint{1.603829in}{0.254453in}}{\pgfqpoint{1.608764in}{0.249517in}}%
\pgfpathcurveto{\pgfqpoint{1.613700in}{0.244581in}}{\pgfqpoint{1.620395in}{0.241808in}}{\pgfqpoint{1.627375in}{0.241808in}}%
\pgfpathclose%
\pgfusepath{stroke,fill}%
\end{pgfscope}%
\begin{pgfscope}%
\pgfpathrectangle{\pgfqpoint{0.681500in}{0.058400in}}{\pgfqpoint{1.621500in}{1.086240in}} %
\pgfusepath{clip}%
\pgfsetbuttcap%
\pgfsetroundjoin%
\definecolor{currentfill}{rgb}{0.498039,0.498039,0.498039}%
\pgfsetfillcolor{currentfill}%
\pgfsetlinewidth{1.138252pt}%
\definecolor{currentstroke}{rgb}{0.498039,0.498039,0.498039}%
\pgfsetstrokecolor{currentstroke}%
\pgfsetdash{}{0pt}%
\pgfpathmoveto{\pgfqpoint{1.897625in}{0.275870in}}%
\pgfpathcurveto{\pgfqpoint{1.904605in}{0.275870in}}{\pgfqpoint{1.911300in}{0.278643in}}{\pgfqpoint{1.916236in}{0.283579in}}%
\pgfpathcurveto{\pgfqpoint{1.921171in}{0.288514in}}{\pgfqpoint{1.923945in}{0.295209in}}{\pgfqpoint{1.923945in}{0.302189in}}%
\pgfpathcurveto{\pgfqpoint{1.923945in}{0.309169in}}{\pgfqpoint{1.921171in}{0.315864in}}{\pgfqpoint{1.916236in}{0.320800in}}%
\pgfpathcurveto{\pgfqpoint{1.911300in}{0.325736in}}{\pgfqpoint{1.904605in}{0.328509in}}{\pgfqpoint{1.897625in}{0.328509in}}%
\pgfpathcurveto{\pgfqpoint{1.890645in}{0.328509in}}{\pgfqpoint{1.883950in}{0.325736in}}{\pgfqpoint{1.879014in}{0.320800in}}%
\pgfpathcurveto{\pgfqpoint{1.874079in}{0.315864in}}{\pgfqpoint{1.871305in}{0.309169in}}{\pgfqpoint{1.871305in}{0.302189in}}%
\pgfpathcurveto{\pgfqpoint{1.871305in}{0.295209in}}{\pgfqpoint{1.874079in}{0.288514in}}{\pgfqpoint{1.879014in}{0.283579in}}%
\pgfpathcurveto{\pgfqpoint{1.883950in}{0.278643in}}{\pgfqpoint{1.890645in}{0.275870in}}{\pgfqpoint{1.897625in}{0.275870in}}%
\pgfpathclose%
\pgfusepath{stroke,fill}%
\end{pgfscope}%
\begin{pgfscope}%
\pgfpathrectangle{\pgfqpoint{0.681500in}{0.058400in}}{\pgfqpoint{1.621500in}{1.086240in}} %
\pgfusepath{clip}%
\pgfsetbuttcap%
\pgfsetroundjoin%
\definecolor{currentfill}{rgb}{0.498039,0.498039,0.498039}%
\pgfsetfillcolor{currentfill}%
\pgfsetlinewidth{1.138252pt}%
\definecolor{currentstroke}{rgb}{0.498039,0.498039,0.498039}%
\pgfsetstrokecolor{currentstroke}%
\pgfsetdash{}{0pt}%
\pgfpathmoveto{\pgfqpoint{2.167875in}{0.350100in}}%
\pgfpathcurveto{\pgfqpoint{2.174855in}{0.350100in}}{\pgfqpoint{2.181550in}{0.352873in}}{\pgfqpoint{2.186486in}{0.357808in}}%
\pgfpathcurveto{\pgfqpoint{2.191421in}{0.362744in}}{\pgfqpoint{2.194195in}{0.369439in}}{\pgfqpoint{2.194195in}{0.376419in}}%
\pgfpathcurveto{\pgfqpoint{2.194195in}{0.383399in}}{\pgfqpoint{2.191421in}{0.390094in}}{\pgfqpoint{2.186486in}{0.395030in}}%
\pgfpathcurveto{\pgfqpoint{2.181550in}{0.399966in}}{\pgfqpoint{2.174855in}{0.402739in}}{\pgfqpoint{2.167875in}{0.402739in}}%
\pgfpathcurveto{\pgfqpoint{2.160895in}{0.402739in}}{\pgfqpoint{2.154200in}{0.399966in}}{\pgfqpoint{2.149264in}{0.395030in}}%
\pgfpathcurveto{\pgfqpoint{2.144329in}{0.390094in}}{\pgfqpoint{2.141555in}{0.383399in}}{\pgfqpoint{2.141555in}{0.376419in}}%
\pgfpathcurveto{\pgfqpoint{2.141555in}{0.369439in}}{\pgfqpoint{2.144329in}{0.362744in}}{\pgfqpoint{2.149264in}{0.357808in}}%
\pgfpathcurveto{\pgfqpoint{2.154200in}{0.352873in}}{\pgfqpoint{2.160895in}{0.350100in}}{\pgfqpoint{2.167875in}{0.350100in}}%
\pgfpathclose%
\pgfusepath{stroke,fill}%
\end{pgfscope}%
\begin{pgfscope}%
\pgfpathrectangle{\pgfqpoint{0.681500in}{0.058400in}}{\pgfqpoint{1.621500in}{1.086240in}} %
\pgfusepath{clip}%
\pgfsetroundcap%
\pgfsetroundjoin%
\pgfsetlinewidth{1.517670pt}%
\definecolor{currentstroke}{rgb}{0.498039,0.498039,0.498039}%
\pgfsetstrokecolor{currentstroke}%
\pgfsetdash{}{0pt}%
\pgfpathmoveto{\pgfqpoint{0.816625in}{0.104241in}}%
\pgfpathlineto{\pgfqpoint{1.086875in}{0.140792in}}%
\pgfpathlineto{\pgfqpoint{1.357125in}{0.212061in}}%
\pgfpathlineto{\pgfqpoint{1.627375in}{0.268128in}}%
\pgfpathlineto{\pgfqpoint{1.897625in}{0.302189in}}%
\pgfpathlineto{\pgfqpoint{2.167875in}{0.376419in}}%
\pgfusepath{stroke}%
\end{pgfscope}%
\begin{pgfscope}%
\pgfpathrectangle{\pgfqpoint{0.681500in}{0.058400in}}{\pgfqpoint{1.621500in}{1.086240in}} %
\pgfusepath{clip}%
\pgfsetroundcap%
\pgfsetroundjoin%
\pgfsetlinewidth{2.168100pt}%
\definecolor{currentstroke}{rgb}{0.498039,0.498039,0.498039}%
\pgfsetstrokecolor{currentstroke}%
\pgfsetdash{}{0pt}%
\pgfpathmoveto{\pgfqpoint{0.816625in}{0.103263in}}%
\pgfpathlineto{\pgfqpoint{0.816625in}{0.106414in}}%
\pgfusepath{stroke}%
\end{pgfscope}%
\begin{pgfscope}%
\pgfpathrectangle{\pgfqpoint{0.681500in}{0.058400in}}{\pgfqpoint{1.621500in}{1.086240in}} %
\pgfusepath{clip}%
\pgfsetroundcap%
\pgfsetroundjoin%
\pgfsetlinewidth{2.168100pt}%
\definecolor{currentstroke}{rgb}{0.498039,0.498039,0.498039}%
\pgfsetstrokecolor{currentstroke}%
\pgfsetdash{}{0pt}%
\pgfpathmoveto{\pgfqpoint{1.086875in}{0.138265in}}%
\pgfpathlineto{\pgfqpoint{1.086875in}{0.142053in}}%
\pgfusepath{stroke}%
\end{pgfscope}%
\begin{pgfscope}%
\pgfpathrectangle{\pgfqpoint{0.681500in}{0.058400in}}{\pgfqpoint{1.621500in}{1.086240in}} %
\pgfusepath{clip}%
\pgfsetroundcap%
\pgfsetroundjoin%
\pgfsetlinewidth{2.168100pt}%
\definecolor{currentstroke}{rgb}{0.498039,0.498039,0.498039}%
\pgfsetstrokecolor{currentstroke}%
\pgfsetdash{}{0pt}%
\pgfpathmoveto{\pgfqpoint{1.357125in}{0.204249in}}%
\pgfpathlineto{\pgfqpoint{1.357125in}{0.220603in}}%
\pgfusepath{stroke}%
\end{pgfscope}%
\begin{pgfscope}%
\pgfpathrectangle{\pgfqpoint{0.681500in}{0.058400in}}{\pgfqpoint{1.621500in}{1.086240in}} %
\pgfusepath{clip}%
\pgfsetroundcap%
\pgfsetroundjoin%
\pgfsetlinewidth{2.168100pt}%
\definecolor{currentstroke}{rgb}{0.498039,0.498039,0.498039}%
\pgfsetstrokecolor{currentstroke}%
\pgfsetdash{}{0pt}%
\pgfpathmoveto{\pgfqpoint{1.627375in}{0.248722in}}%
\pgfpathlineto{\pgfqpoint{1.627375in}{0.277650in}}%
\pgfusepath{stroke}%
\end{pgfscope}%
\begin{pgfscope}%
\pgfpathrectangle{\pgfqpoint{0.681500in}{0.058400in}}{\pgfqpoint{1.621500in}{1.086240in}} %
\pgfusepath{clip}%
\pgfsetroundcap%
\pgfsetroundjoin%
\pgfsetlinewidth{2.168100pt}%
\definecolor{currentstroke}{rgb}{0.498039,0.498039,0.498039}%
\pgfsetstrokecolor{currentstroke}%
\pgfsetdash{}{0pt}%
\pgfpathmoveto{\pgfqpoint{1.897625in}{0.287745in}}%
\pgfpathlineto{\pgfqpoint{1.897625in}{0.317809in}}%
\pgfusepath{stroke}%
\end{pgfscope}%
\begin{pgfscope}%
\pgfpathrectangle{\pgfqpoint{0.681500in}{0.058400in}}{\pgfqpoint{1.621500in}{1.086240in}} %
\pgfusepath{clip}%
\pgfsetroundcap%
\pgfsetroundjoin%
\pgfsetlinewidth{2.168100pt}%
\definecolor{currentstroke}{rgb}{0.498039,0.498039,0.498039}%
\pgfsetstrokecolor{currentstroke}%
\pgfsetdash{}{0pt}%
\pgfpathmoveto{\pgfqpoint{2.167875in}{0.370136in}}%
\pgfpathlineto{\pgfqpoint{2.167875in}{0.385265in}}%
\pgfusepath{stroke}%
\end{pgfscope}%
\begin{pgfscope}%
\pgfpathrectangle{\pgfqpoint{0.681500in}{0.058400in}}{\pgfqpoint{1.621500in}{1.086240in}} %
\pgfusepath{clip}%
\pgfsetbuttcap%
\pgfsetroundjoin%
\pgfsetlinewidth{2.007500pt}%
\definecolor{currentstroke}{rgb}{0.890196,0.466667,0.760784}%
\pgfsetstrokecolor{currentstroke}%
\pgfsetdash{{7.400000pt}{3.200000pt}}{0.000000pt}%
\pgfpathmoveto{\pgfqpoint{0.681500in}{0.959417in}}%
\pgfpathlineto{\pgfqpoint{2.303000in}{0.959417in}}%
\pgfusepath{stroke}%
\end{pgfscope}%
\end{pgfpicture}%
\makeatother%
\endgroup%

%% file: images/plots/exec_time_puma8NH.pgf
\begingroup%
\makeatletter%
\begin{pgfpicture}%
\pgfpathrectangle{\pgfpointorigin}{\pgfqpoint{2.150000in}{1.168000in}}%
\pgfusepath{use as bounding box, clip}%
\begin{pgfscope}%
\pgfsetbuttcap%
\pgfsetmiterjoin%
\pgfsetlinewidth{0.000000pt}%
\definecolor{currentstroke}{rgb}{0.000000,0.000000,0.000000}%
\pgfsetstrokecolor{currentstroke}%
\pgfsetstrokeopacity{0.000000}%
\pgfsetdash{}{0pt}%
\pgfpathmoveto{\pgfqpoint{0.000000in}{0.000000in}}%
\pgfpathlineto{\pgfqpoint{2.150000in}{0.000000in}}%
\pgfpathlineto{\pgfqpoint{2.150000in}{1.168000in}}%
\pgfpathlineto{\pgfqpoint{0.000000in}{1.168000in}}%
\pgfpathclose%
\pgfusepath{}%
\end{pgfscope}%
\begin{pgfscope}%
\pgfsetbuttcap%
\pgfsetmiterjoin%
\pgfsetlinewidth{0.000000pt}%
\definecolor{currentstroke}{rgb}{0.000000,0.000000,0.000000}%
\pgfsetstrokecolor{currentstroke}%
\pgfsetstrokeopacity{0.000000}%
\pgfsetdash{}{0pt}%
\pgfpathmoveto{\pgfqpoint{0.516000in}{0.058400in}}%
\pgfpathlineto{\pgfqpoint{2.107000in}{0.058400in}}%
\pgfpathlineto{\pgfqpoint{2.107000in}{1.144640in}}%
\pgfpathlineto{\pgfqpoint{0.516000in}{1.144640in}}%
\pgfpathclose%
\pgfusepath{}%
\end{pgfscope}%
\begin{pgfscope}%
\pgfpathrectangle{\pgfqpoint{0.516000in}{0.058400in}}{\pgfqpoint{1.591000in}{1.086240in}} %
\pgfusepath{clip}%
\pgfsetroundcap%
\pgfsetroundjoin%
\pgfsetlinewidth{0.803000pt}%
\definecolor{currentstroke}{rgb}{0.800000,0.800000,0.800000}%
\pgfsetstrokecolor{currentstroke}%
\pgfsetdash{}{0pt}%
\pgfpathmoveto{\pgfqpoint{0.648583in}{0.058400in}}%
\pgfpathlineto{\pgfqpoint{0.648583in}{1.144640in}}%
\pgfusepath{stroke}%
\end{pgfscope}%
\begin{pgfscope}%
\pgfpathrectangle{\pgfqpoint{0.516000in}{0.058400in}}{\pgfqpoint{1.591000in}{1.086240in}} %
\pgfusepath{clip}%
\pgfsetroundcap%
\pgfsetroundjoin%
\pgfsetlinewidth{0.803000pt}%
\definecolor{currentstroke}{rgb}{0.800000,0.800000,0.800000}%
\pgfsetstrokecolor{currentstroke}%
\pgfsetdash{}{0pt}%
\pgfpathmoveto{\pgfqpoint{0.913750in}{0.058400in}}%
\pgfpathlineto{\pgfqpoint{0.913750in}{1.144640in}}%
\pgfusepath{stroke}%
\end{pgfscope}%
\begin{pgfscope}%
\pgfpathrectangle{\pgfqpoint{0.516000in}{0.058400in}}{\pgfqpoint{1.591000in}{1.086240in}} %
\pgfusepath{clip}%
\pgfsetroundcap%
\pgfsetroundjoin%
\pgfsetlinewidth{0.803000pt}%
\definecolor{currentstroke}{rgb}{0.800000,0.800000,0.800000}%
\pgfsetstrokecolor{currentstroke}%
\pgfsetdash{}{0pt}%
\pgfpathmoveto{\pgfqpoint{1.178917in}{0.058400in}}%
\pgfpathlineto{\pgfqpoint{1.178917in}{1.144640in}}%
\pgfusepath{stroke}%
\end{pgfscope}%
\begin{pgfscope}%
\pgfpathrectangle{\pgfqpoint{0.516000in}{0.058400in}}{\pgfqpoint{1.591000in}{1.086240in}} %
\pgfusepath{clip}%
\pgfsetroundcap%
\pgfsetroundjoin%
\pgfsetlinewidth{0.803000pt}%
\definecolor{currentstroke}{rgb}{0.800000,0.800000,0.800000}%
\pgfsetstrokecolor{currentstroke}%
\pgfsetdash{}{0pt}%
\pgfpathmoveto{\pgfqpoint{1.444083in}{0.058400in}}%
\pgfpathlineto{\pgfqpoint{1.444083in}{1.144640in}}%
\pgfusepath{stroke}%
\end{pgfscope}%
\begin{pgfscope}%
\pgfpathrectangle{\pgfqpoint{0.516000in}{0.058400in}}{\pgfqpoint{1.591000in}{1.086240in}} %
\pgfusepath{clip}%
\pgfsetroundcap%
\pgfsetroundjoin%
\pgfsetlinewidth{0.803000pt}%
\definecolor{currentstroke}{rgb}{0.800000,0.800000,0.800000}%
\pgfsetstrokecolor{currentstroke}%
\pgfsetdash{}{0pt}%
\pgfpathmoveto{\pgfqpoint{1.709250in}{0.058400in}}%
\pgfpathlineto{\pgfqpoint{1.709250in}{1.144640in}}%
\pgfusepath{stroke}%
\end{pgfscope}%
\begin{pgfscope}%
\pgfpathrectangle{\pgfqpoint{0.516000in}{0.058400in}}{\pgfqpoint{1.591000in}{1.086240in}} %
\pgfusepath{clip}%
\pgfsetroundcap%
\pgfsetroundjoin%
\pgfsetlinewidth{0.803000pt}%
\definecolor{currentstroke}{rgb}{0.800000,0.800000,0.800000}%
\pgfsetstrokecolor{currentstroke}%
\pgfsetdash{}{0pt}%
\pgfpathmoveto{\pgfqpoint{1.974417in}{0.058400in}}%
\pgfpathlineto{\pgfqpoint{1.974417in}{1.144640in}}%
\pgfusepath{stroke}%
\end{pgfscope}%
\begin{pgfscope}%
\pgfpathrectangle{\pgfqpoint{0.516000in}{0.058400in}}{\pgfqpoint{1.591000in}{1.086240in}} %
\pgfusepath{clip}%
\pgfsetroundcap%
\pgfsetroundjoin%
\pgfsetlinewidth{0.803000pt}%
\definecolor{currentstroke}{rgb}{0.800000,0.800000,0.800000}%
\pgfsetstrokecolor{currentstroke}%
\pgfsetdash{}{0pt}%
\pgfpathmoveto{\pgfqpoint{0.516000in}{0.058400in}}%
\pgfpathlineto{\pgfqpoint{2.107000in}{0.058400in}}%
\pgfusepath{stroke}%
\end{pgfscope}%
\begin{pgfscope}%
\definecolor{textcolor}{rgb}{0.150000,0.150000,0.150000}%
\pgfsetstrokecolor{textcolor}%
\pgfsetfillcolor{textcolor}%
\pgftext[x=0.026690in,y=0.014997in,left,base]{\color{textcolor}\fontsize{8.800000}{10.560000}\selectfont 00h 00m}%
\end{pgfscope}%
\begin{pgfscope}%
\pgfpathrectangle{\pgfqpoint{0.516000in}{0.058400in}}{\pgfqpoint{1.591000in}{1.086240in}} %
\pgfusepath{clip}%
\pgfsetroundcap%
\pgfsetroundjoin%
\pgfsetlinewidth{0.803000pt}%
\definecolor{currentstroke}{rgb}{0.800000,0.800000,0.800000}%
\pgfsetstrokecolor{currentstroke}%
\pgfsetdash{}{0pt}%
\pgfpathmoveto{\pgfqpoint{0.516000in}{0.506706in}}%
\pgfpathlineto{\pgfqpoint{2.107000in}{0.506706in}}%
\pgfusepath{stroke}%
\end{pgfscope}%
\begin{pgfscope}%
\definecolor{textcolor}{rgb}{0.150000,0.150000,0.150000}%
\pgfsetstrokecolor{textcolor}%
\pgfsetfillcolor{textcolor}%
\pgftext[x=0.026690in,y=0.463303in,left,base]{\color{textcolor}\fontsize{8.800000}{10.560000}\selectfont 14h 00m}%
\end{pgfscope}%
\begin{pgfscope}%
\pgfpathrectangle{\pgfqpoint{0.516000in}{0.058400in}}{\pgfqpoint{1.591000in}{1.086240in}} %
\pgfusepath{clip}%
\pgfsetroundcap%
\pgfsetroundjoin%
\pgfsetlinewidth{0.803000pt}%
\definecolor{currentstroke}{rgb}{0.800000,0.800000,0.800000}%
\pgfsetstrokecolor{currentstroke}%
\pgfsetdash{}{0pt}%
\pgfpathmoveto{\pgfqpoint{0.516000in}{0.955012in}}%
\pgfpathlineto{\pgfqpoint{2.107000in}{0.955012in}}%
\pgfusepath{stroke}%
\end{pgfscope}%
\begin{pgfscope}%
\definecolor{textcolor}{rgb}{0.150000,0.150000,0.150000}%
\pgfsetstrokecolor{textcolor}%
\pgfsetfillcolor{textcolor}%
\pgftext[x=0.126612in,y=0.911609in,left,base]{\color{textcolor}\fontsize{8.800000}{10.560000}\selectfont 1d 04h}%
\end{pgfscope}%
\begin{pgfscope}%
\pgfpathrectangle{\pgfqpoint{0.516000in}{0.058400in}}{\pgfqpoint{1.591000in}{1.086240in}} %
\pgfusepath{clip}%
\pgfsetbuttcap%
\pgfsetroundjoin%
\definecolor{currentfill}{rgb}{0.737255,0.741176,0.133333}%
\pgfsetfillcolor{currentfill}%
\pgfsetlinewidth{1.138252pt}%
\definecolor{currentstroke}{rgb}{0.737255,0.741176,0.133333}%
\pgfsetstrokecolor{currentstroke}%
\pgfsetdash{}{0pt}%
\pgfpathmoveto{\pgfqpoint{0.648583in}{0.078019in}}%
\pgfpathcurveto{\pgfqpoint{0.655563in}{0.078019in}}{\pgfqpoint{0.662258in}{0.080793in}}{\pgfqpoint{0.667194in}{0.085728in}}%
\pgfpathcurveto{\pgfqpoint{0.672130in}{0.090664in}}{\pgfqpoint{0.674903in}{0.097359in}}{\pgfqpoint{0.674903in}{0.104339in}}%
\pgfpathcurveto{\pgfqpoint{0.674903in}{0.111319in}}{\pgfqpoint{0.672130in}{0.118014in}}{\pgfqpoint{0.667194in}{0.122950in}}%
\pgfpathcurveto{\pgfqpoint{0.662258in}{0.127885in}}{\pgfqpoint{0.655563in}{0.130659in}}{\pgfqpoint{0.648583in}{0.130659in}}%
\pgfpathcurveto{\pgfqpoint{0.641603in}{0.130659in}}{\pgfqpoint{0.634908in}{0.127885in}}{\pgfqpoint{0.629973in}{0.122950in}}%
\pgfpathcurveto{\pgfqpoint{0.625037in}{0.118014in}}{\pgfqpoint{0.622264in}{0.111319in}}{\pgfqpoint{0.622264in}{0.104339in}}%
\pgfpathcurveto{\pgfqpoint{0.622264in}{0.097359in}}{\pgfqpoint{0.625037in}{0.090664in}}{\pgfqpoint{0.629973in}{0.085728in}}%
\pgfpathcurveto{\pgfqpoint{0.634908in}{0.080793in}}{\pgfqpoint{0.641603in}{0.078019in}}{\pgfqpoint{0.648583in}{0.078019in}}%
\pgfpathclose%
\pgfusepath{stroke,fill}%
\end{pgfscope}%
\begin{pgfscope}%
\pgfpathrectangle{\pgfqpoint{0.516000in}{0.058400in}}{\pgfqpoint{1.591000in}{1.086240in}} %
\pgfusepath{clip}%
\pgfsetbuttcap%
\pgfsetroundjoin%
\definecolor{currentfill}{rgb}{0.737255,0.741176,0.133333}%
\pgfsetfillcolor{currentfill}%
\pgfsetlinewidth{1.138252pt}%
\definecolor{currentstroke}{rgb}{0.737255,0.741176,0.133333}%
\pgfsetstrokecolor{currentstroke}%
\pgfsetdash{}{0pt}%
\pgfpathmoveto{\pgfqpoint{0.913750in}{0.124300in}}%
\pgfpathcurveto{\pgfqpoint{0.920730in}{0.124300in}}{\pgfqpoint{0.927425in}{0.127073in}}{\pgfqpoint{0.932361in}{0.132009in}}%
\pgfpathcurveto{\pgfqpoint{0.937296in}{0.136944in}}{\pgfqpoint{0.940070in}{0.143639in}}{\pgfqpoint{0.940070in}{0.150619in}}%
\pgfpathcurveto{\pgfqpoint{0.940070in}{0.157599in}}{\pgfqpoint{0.937296in}{0.164295in}}{\pgfqpoint{0.932361in}{0.169230in}}%
\pgfpathcurveto{\pgfqpoint{0.927425in}{0.174166in}}{\pgfqpoint{0.920730in}{0.176939in}}{\pgfqpoint{0.913750in}{0.176939in}}%
\pgfpathcurveto{\pgfqpoint{0.906770in}{0.176939in}}{\pgfqpoint{0.900075in}{0.174166in}}{\pgfqpoint{0.895139in}{0.169230in}}%
\pgfpathcurveto{\pgfqpoint{0.890204in}{0.164295in}}{\pgfqpoint{0.887430in}{0.157599in}}{\pgfqpoint{0.887430in}{0.150619in}}%
\pgfpathcurveto{\pgfqpoint{0.887430in}{0.143639in}}{\pgfqpoint{0.890204in}{0.136944in}}{\pgfqpoint{0.895139in}{0.132009in}}%
\pgfpathcurveto{\pgfqpoint{0.900075in}{0.127073in}}{\pgfqpoint{0.906770in}{0.124300in}}{\pgfqpoint{0.913750in}{0.124300in}}%
\pgfpathclose%
\pgfusepath{stroke,fill}%
\end{pgfscope}%
\begin{pgfscope}%
\pgfpathrectangle{\pgfqpoint{0.516000in}{0.058400in}}{\pgfqpoint{1.591000in}{1.086240in}} %
\pgfusepath{clip}%
\pgfsetbuttcap%
\pgfsetroundjoin%
\definecolor{currentfill}{rgb}{0.737255,0.741176,0.133333}%
\pgfsetfillcolor{currentfill}%
\pgfsetlinewidth{1.138252pt}%
\definecolor{currentstroke}{rgb}{0.737255,0.741176,0.133333}%
\pgfsetstrokecolor{currentstroke}%
\pgfsetdash{}{0pt}%
\pgfpathmoveto{\pgfqpoint{1.178917in}{0.213434in}}%
\pgfpathcurveto{\pgfqpoint{1.185897in}{0.213434in}}{\pgfqpoint{1.192592in}{0.216207in}}{\pgfqpoint{1.197527in}{0.221143in}}%
\pgfpathcurveto{\pgfqpoint{1.202463in}{0.226079in}}{\pgfqpoint{1.205236in}{0.232774in}}{\pgfqpoint{1.205236in}{0.239754in}}%
\pgfpathcurveto{\pgfqpoint{1.205236in}{0.246734in}}{\pgfqpoint{1.202463in}{0.253429in}}{\pgfqpoint{1.197527in}{0.258364in}}%
\pgfpathcurveto{\pgfqpoint{1.192592in}{0.263300in}}{\pgfqpoint{1.185897in}{0.266073in}}{\pgfqpoint{1.178917in}{0.266073in}}%
\pgfpathcurveto{\pgfqpoint{1.171937in}{0.266073in}}{\pgfqpoint{1.165242in}{0.263300in}}{\pgfqpoint{1.160306in}{0.258364in}}%
\pgfpathcurveto{\pgfqpoint{1.155370in}{0.253429in}}{\pgfqpoint{1.152597in}{0.246734in}}{\pgfqpoint{1.152597in}{0.239754in}}%
\pgfpathcurveto{\pgfqpoint{1.152597in}{0.232774in}}{\pgfqpoint{1.155370in}{0.226079in}}{\pgfqpoint{1.160306in}{0.221143in}}%
\pgfpathcurveto{\pgfqpoint{1.165242in}{0.216207in}}{\pgfqpoint{1.171937in}{0.213434in}}{\pgfqpoint{1.178917in}{0.213434in}}%
\pgfpathclose%
\pgfusepath{stroke,fill}%
\end{pgfscope}%
\begin{pgfscope}%
\pgfpathrectangle{\pgfqpoint{0.516000in}{0.058400in}}{\pgfqpoint{1.591000in}{1.086240in}} %
\pgfusepath{clip}%
\pgfsetbuttcap%
\pgfsetroundjoin%
\definecolor{currentfill}{rgb}{0.737255,0.741176,0.133333}%
\pgfsetfillcolor{currentfill}%
\pgfsetlinewidth{1.138252pt}%
\definecolor{currentstroke}{rgb}{0.737255,0.741176,0.133333}%
\pgfsetstrokecolor{currentstroke}%
\pgfsetdash{}{0pt}%
\pgfpathmoveto{\pgfqpoint{1.444083in}{0.290895in}}%
\pgfpathcurveto{\pgfqpoint{1.451063in}{0.290895in}}{\pgfqpoint{1.457758in}{0.293669in}}{\pgfqpoint{1.462694in}{0.298604in}}%
\pgfpathcurveto{\pgfqpoint{1.467630in}{0.303540in}}{\pgfqpoint{1.470403in}{0.310235in}}{\pgfqpoint{1.470403in}{0.317215in}}%
\pgfpathcurveto{\pgfqpoint{1.470403in}{0.324195in}}{\pgfqpoint{1.467630in}{0.330890in}}{\pgfqpoint{1.462694in}{0.335826in}}%
\pgfpathcurveto{\pgfqpoint{1.457758in}{0.340761in}}{\pgfqpoint{1.451063in}{0.343535in}}{\pgfqpoint{1.444083in}{0.343535in}}%
\pgfpathcurveto{\pgfqpoint{1.437103in}{0.343535in}}{\pgfqpoint{1.430408in}{0.340761in}}{\pgfqpoint{1.425473in}{0.335826in}}%
\pgfpathcurveto{\pgfqpoint{1.420537in}{0.330890in}}{\pgfqpoint{1.417764in}{0.324195in}}{\pgfqpoint{1.417764in}{0.317215in}}%
\pgfpathcurveto{\pgfqpoint{1.417764in}{0.310235in}}{\pgfqpoint{1.420537in}{0.303540in}}{\pgfqpoint{1.425473in}{0.298604in}}%
\pgfpathcurveto{\pgfqpoint{1.430408in}{0.293669in}}{\pgfqpoint{1.437103in}{0.290895in}}{\pgfqpoint{1.444083in}{0.290895in}}%
\pgfpathclose%
\pgfusepath{stroke,fill}%
\end{pgfscope}%
\begin{pgfscope}%
\pgfpathrectangle{\pgfqpoint{0.516000in}{0.058400in}}{\pgfqpoint{1.591000in}{1.086240in}} %
\pgfusepath{clip}%
\pgfsetbuttcap%
\pgfsetroundjoin%
\definecolor{currentfill}{rgb}{0.737255,0.741176,0.133333}%
\pgfsetfillcolor{currentfill}%
\pgfsetlinewidth{1.138252pt}%
\definecolor{currentstroke}{rgb}{0.737255,0.741176,0.133333}%
\pgfsetstrokecolor{currentstroke}%
\pgfsetdash{}{0pt}%
\pgfpathmoveto{\pgfqpoint{1.709250in}{0.413322in}}%
\pgfpathcurveto{\pgfqpoint{1.716230in}{0.413322in}}{\pgfqpoint{1.722925in}{0.416095in}}{\pgfqpoint{1.727861in}{0.421031in}}%
\pgfpathcurveto{\pgfqpoint{1.732796in}{0.425966in}}{\pgfqpoint{1.735570in}{0.432661in}}{\pgfqpoint{1.735570in}{0.439641in}}%
\pgfpathcurveto{\pgfqpoint{1.735570in}{0.446621in}}{\pgfqpoint{1.732796in}{0.453316in}}{\pgfqpoint{1.727861in}{0.458252in}}%
\pgfpathcurveto{\pgfqpoint{1.722925in}{0.463188in}}{\pgfqpoint{1.716230in}{0.465961in}}{\pgfqpoint{1.709250in}{0.465961in}}%
\pgfpathcurveto{\pgfqpoint{1.702270in}{0.465961in}}{\pgfqpoint{1.695575in}{0.463188in}}{\pgfqpoint{1.690639in}{0.458252in}}%
\pgfpathcurveto{\pgfqpoint{1.685704in}{0.453316in}}{\pgfqpoint{1.682930in}{0.446621in}}{\pgfqpoint{1.682930in}{0.439641in}}%
\pgfpathcurveto{\pgfqpoint{1.682930in}{0.432661in}}{\pgfqpoint{1.685704in}{0.425966in}}{\pgfqpoint{1.690639in}{0.421031in}}%
\pgfpathcurveto{\pgfqpoint{1.695575in}{0.416095in}}{\pgfqpoint{1.702270in}{0.413322in}}{\pgfqpoint{1.709250in}{0.413322in}}%
\pgfpathclose%
\pgfusepath{stroke,fill}%
\end{pgfscope}%
\begin{pgfscope}%
\pgfpathrectangle{\pgfqpoint{0.516000in}{0.058400in}}{\pgfqpoint{1.591000in}{1.086240in}} %
\pgfusepath{clip}%
\pgfsetbuttcap%
\pgfsetroundjoin%
\definecolor{currentfill}{rgb}{0.737255,0.741176,0.133333}%
\pgfsetfillcolor{currentfill}%
\pgfsetlinewidth{1.138252pt}%
\definecolor{currentstroke}{rgb}{0.737255,0.741176,0.133333}%
\pgfsetstrokecolor{currentstroke}%
\pgfsetdash{}{0pt}%
\pgfpathmoveto{\pgfqpoint{1.974417in}{0.464679in}}%
\pgfpathcurveto{\pgfqpoint{1.981397in}{0.464679in}}{\pgfqpoint{1.988092in}{0.467452in}}{\pgfqpoint{1.993027in}{0.472387in}}%
\pgfpathcurveto{\pgfqpoint{1.997963in}{0.477323in}}{\pgfqpoint{2.000736in}{0.484018in}}{\pgfqpoint{2.000736in}{0.490998in}}%
\pgfpathcurveto{\pgfqpoint{2.000736in}{0.497978in}}{\pgfqpoint{1.997963in}{0.504673in}}{\pgfqpoint{1.993027in}{0.509609in}}%
\pgfpathcurveto{\pgfqpoint{1.988092in}{0.514545in}}{\pgfqpoint{1.981397in}{0.517318in}}{\pgfqpoint{1.974417in}{0.517318in}}%
\pgfpathcurveto{\pgfqpoint{1.967437in}{0.517318in}}{\pgfqpoint{1.960742in}{0.514545in}}{\pgfqpoint{1.955806in}{0.509609in}}%
\pgfpathcurveto{\pgfqpoint{1.950870in}{0.504673in}}{\pgfqpoint{1.948097in}{0.497978in}}{\pgfqpoint{1.948097in}{0.490998in}}%
\pgfpathcurveto{\pgfqpoint{1.948097in}{0.484018in}}{\pgfqpoint{1.950870in}{0.477323in}}{\pgfqpoint{1.955806in}{0.472387in}}%
\pgfpathcurveto{\pgfqpoint{1.960742in}{0.467452in}}{\pgfqpoint{1.967437in}{0.464679in}}{\pgfqpoint{1.974417in}{0.464679in}}%
\pgfpathclose%
\pgfusepath{stroke,fill}%
\end{pgfscope}%
\begin{pgfscope}%
\pgfpathrectangle{\pgfqpoint{0.516000in}{0.058400in}}{\pgfqpoint{1.591000in}{1.086240in}} %
\pgfusepath{clip}%
\pgfsetroundcap%
\pgfsetroundjoin%
\pgfsetlinewidth{1.517670pt}%
\definecolor{currentstroke}{rgb}{0.737255,0.741176,0.133333}%
\pgfsetstrokecolor{currentstroke}%
\pgfsetdash{}{0pt}%
\pgfpathmoveto{\pgfqpoint{0.648583in}{0.104339in}}%
\pgfpathlineto{\pgfqpoint{0.913750in}{0.150619in}}%
\pgfpathlineto{\pgfqpoint{1.178917in}{0.239754in}}%
\pgfpathlineto{\pgfqpoint{1.444083in}{0.317215in}}%
\pgfpathlineto{\pgfqpoint{1.709250in}{0.439641in}}%
\pgfpathlineto{\pgfqpoint{1.974417in}{0.490998in}}%
\pgfusepath{stroke}%
\end{pgfscope}%
\begin{pgfscope}%
\pgfpathrectangle{\pgfqpoint{0.516000in}{0.058400in}}{\pgfqpoint{1.591000in}{1.086240in}} %
\pgfusepath{clip}%
\pgfsetroundcap%
\pgfsetroundjoin%
\pgfsetlinewidth{2.168100pt}%
\definecolor{currentstroke}{rgb}{0.737255,0.741176,0.133333}%
\pgfsetstrokecolor{currentstroke}%
\pgfsetdash{}{0pt}%
\pgfpathmoveto{\pgfqpoint{0.648583in}{0.101850in}}%
\pgfpathlineto{\pgfqpoint{0.648583in}{0.105250in}}%
\pgfusepath{stroke}%
\end{pgfscope}%
\begin{pgfscope}%
\pgfpathrectangle{\pgfqpoint{0.516000in}{0.058400in}}{\pgfqpoint{1.591000in}{1.086240in}} %
\pgfusepath{clip}%
\pgfsetroundcap%
\pgfsetroundjoin%
\pgfsetlinewidth{2.168100pt}%
\definecolor{currentstroke}{rgb}{0.737255,0.741176,0.133333}%
\pgfsetstrokecolor{currentstroke}%
\pgfsetdash{}{0pt}%
\pgfpathmoveto{\pgfqpoint{0.913750in}{0.147450in}}%
\pgfpathlineto{\pgfqpoint{0.913750in}{0.158793in}}%
\pgfusepath{stroke}%
\end{pgfscope}%
\begin{pgfscope}%
\pgfpathrectangle{\pgfqpoint{0.516000in}{0.058400in}}{\pgfqpoint{1.591000in}{1.086240in}} %
\pgfusepath{clip}%
\pgfsetroundcap%
\pgfsetroundjoin%
\pgfsetlinewidth{2.168100pt}%
\definecolor{currentstroke}{rgb}{0.737255,0.741176,0.133333}%
\pgfsetstrokecolor{currentstroke}%
\pgfsetdash{}{0pt}%
\pgfpathmoveto{\pgfqpoint{1.178917in}{0.233407in}}%
\pgfpathlineto{\pgfqpoint{1.178917in}{0.245256in}}%
\pgfusepath{stroke}%
\end{pgfscope}%
\begin{pgfscope}%
\pgfpathrectangle{\pgfqpoint{0.516000in}{0.058400in}}{\pgfqpoint{1.591000in}{1.086240in}} %
\pgfusepath{clip}%
\pgfsetroundcap%
\pgfsetroundjoin%
\pgfsetlinewidth{2.168100pt}%
\definecolor{currentstroke}{rgb}{0.737255,0.741176,0.133333}%
\pgfsetstrokecolor{currentstroke}%
\pgfsetdash{}{0pt}%
\pgfpathmoveto{\pgfqpoint{1.444083in}{0.314607in}}%
\pgfpathlineto{\pgfqpoint{1.444083in}{0.329830in}}%
\pgfusepath{stroke}%
\end{pgfscope}%
\begin{pgfscope}%
\pgfpathrectangle{\pgfqpoint{0.516000in}{0.058400in}}{\pgfqpoint{1.591000in}{1.086240in}} %
\pgfusepath{clip}%
\pgfsetroundcap%
\pgfsetroundjoin%
\pgfsetlinewidth{2.168100pt}%
\definecolor{currentstroke}{rgb}{0.737255,0.741176,0.133333}%
\pgfsetstrokecolor{currentstroke}%
\pgfsetdash{}{0pt}%
\pgfpathmoveto{\pgfqpoint{1.709250in}{0.426517in}}%
\pgfpathlineto{\pgfqpoint{1.709250in}{0.456619in}}%
\pgfusepath{stroke}%
\end{pgfscope}%
\begin{pgfscope}%
\pgfpathrectangle{\pgfqpoint{0.516000in}{0.058400in}}{\pgfqpoint{1.591000in}{1.086240in}} %
\pgfusepath{clip}%
\pgfsetroundcap%
\pgfsetroundjoin%
\pgfsetlinewidth{2.168100pt}%
\definecolor{currentstroke}{rgb}{0.737255,0.741176,0.133333}%
\pgfsetstrokecolor{currentstroke}%
\pgfsetdash{}{0pt}%
\pgfpathmoveto{\pgfqpoint{1.974417in}{0.471282in}}%
\pgfpathlineto{\pgfqpoint{1.974417in}{0.519097in}}%
\pgfusepath{stroke}%
\end{pgfscope}%
\begin{pgfscope}%
\pgfpathrectangle{\pgfqpoint{0.516000in}{0.058400in}}{\pgfqpoint{1.591000in}{1.086240in}} %
\pgfusepath{clip}%
\pgfsetbuttcap%
\pgfsetroundjoin%
\definecolor{currentfill}{rgb}{0.090196,0.745098,0.811765}%
\pgfsetfillcolor{currentfill}%
\pgfsetlinewidth{1.138252pt}%
\definecolor{currentstroke}{rgb}{0.090196,0.745098,0.811765}%
\pgfsetstrokecolor{currentstroke}%
\pgfsetdash{}{0pt}%
\pgfpathmoveto{\pgfqpoint{0.648583in}{0.078191in}}%
\pgfpathcurveto{\pgfqpoint{0.655563in}{0.078191in}}{\pgfqpoint{0.662258in}{0.080965in}}{\pgfqpoint{0.667194in}{0.085900in}}%
\pgfpathcurveto{\pgfqpoint{0.672130in}{0.090836in}}{\pgfqpoint{0.674903in}{0.097531in}}{\pgfqpoint{0.674903in}{0.104511in}}%
\pgfpathcurveto{\pgfqpoint{0.674903in}{0.111491in}}{\pgfqpoint{0.672130in}{0.118186in}}{\pgfqpoint{0.667194in}{0.123122in}}%
\pgfpathcurveto{\pgfqpoint{0.662258in}{0.128057in}}{\pgfqpoint{0.655563in}{0.130831in}}{\pgfqpoint{0.648583in}{0.130831in}}%
\pgfpathcurveto{\pgfqpoint{0.641603in}{0.130831in}}{\pgfqpoint{0.634908in}{0.128057in}}{\pgfqpoint{0.629973in}{0.123122in}}%
\pgfpathcurveto{\pgfqpoint{0.625037in}{0.118186in}}{\pgfqpoint{0.622264in}{0.111491in}}{\pgfqpoint{0.622264in}{0.104511in}}%
\pgfpathcurveto{\pgfqpoint{0.622264in}{0.097531in}}{\pgfqpoint{0.625037in}{0.090836in}}{\pgfqpoint{0.629973in}{0.085900in}}%
\pgfpathcurveto{\pgfqpoint{0.634908in}{0.080965in}}{\pgfqpoint{0.641603in}{0.078191in}}{\pgfqpoint{0.648583in}{0.078191in}}%
\pgfpathclose%
\pgfusepath{stroke,fill}%
\end{pgfscope}%
\begin{pgfscope}%
\pgfpathrectangle{\pgfqpoint{0.516000in}{0.058400in}}{\pgfqpoint{1.591000in}{1.086240in}} %
\pgfusepath{clip}%
\pgfsetbuttcap%
\pgfsetroundjoin%
\definecolor{currentfill}{rgb}{0.090196,0.745098,0.811765}%
\pgfsetfillcolor{currentfill}%
\pgfsetlinewidth{1.138252pt}%
\definecolor{currentstroke}{rgb}{0.090196,0.745098,0.811765}%
\pgfsetstrokecolor{currentstroke}%
\pgfsetdash{}{0pt}%
\pgfpathmoveto{\pgfqpoint{0.913750in}{0.131264in}}%
\pgfpathcurveto{\pgfqpoint{0.920730in}{0.131264in}}{\pgfqpoint{0.927425in}{0.134038in}}{\pgfqpoint{0.932361in}{0.138973in}}%
\pgfpathcurveto{\pgfqpoint{0.937296in}{0.143909in}}{\pgfqpoint{0.940070in}{0.150604in}}{\pgfqpoint{0.940070in}{0.157584in}}%
\pgfpathcurveto{\pgfqpoint{0.940070in}{0.164564in}}{\pgfqpoint{0.937296in}{0.171259in}}{\pgfqpoint{0.932361in}{0.176195in}}%
\pgfpathcurveto{\pgfqpoint{0.927425in}{0.181130in}}{\pgfqpoint{0.920730in}{0.183904in}}{\pgfqpoint{0.913750in}{0.183904in}}%
\pgfpathcurveto{\pgfqpoint{0.906770in}{0.183904in}}{\pgfqpoint{0.900075in}{0.181130in}}{\pgfqpoint{0.895139in}{0.176195in}}%
\pgfpathcurveto{\pgfqpoint{0.890204in}{0.171259in}}{\pgfqpoint{0.887430in}{0.164564in}}{\pgfqpoint{0.887430in}{0.157584in}}%
\pgfpathcurveto{\pgfqpoint{0.887430in}{0.150604in}}{\pgfqpoint{0.890204in}{0.143909in}}{\pgfqpoint{0.895139in}{0.138973in}}%
\pgfpathcurveto{\pgfqpoint{0.900075in}{0.134038in}}{\pgfqpoint{0.906770in}{0.131264in}}{\pgfqpoint{0.913750in}{0.131264in}}%
\pgfpathclose%
\pgfusepath{stroke,fill}%
\end{pgfscope}%
\begin{pgfscope}%
\pgfpathrectangle{\pgfqpoint{0.516000in}{0.058400in}}{\pgfqpoint{1.591000in}{1.086240in}} %
\pgfusepath{clip}%
\pgfsetbuttcap%
\pgfsetroundjoin%
\definecolor{currentfill}{rgb}{0.090196,0.745098,0.811765}%
\pgfsetfillcolor{currentfill}%
\pgfsetlinewidth{1.138252pt}%
\definecolor{currentstroke}{rgb}{0.090196,0.745098,0.811765}%
\pgfsetstrokecolor{currentstroke}%
\pgfsetdash{}{0pt}%
\pgfpathmoveto{\pgfqpoint{1.178917in}{0.217685in}}%
\pgfpathcurveto{\pgfqpoint{1.185897in}{0.217685in}}{\pgfqpoint{1.192592in}{0.220459in}}{\pgfqpoint{1.197527in}{0.225394in}}%
\pgfpathcurveto{\pgfqpoint{1.202463in}{0.230330in}}{\pgfqpoint{1.205236in}{0.237025in}}{\pgfqpoint{1.205236in}{0.244005in}}%
\pgfpathcurveto{\pgfqpoint{1.205236in}{0.250985in}}{\pgfqpoint{1.202463in}{0.257680in}}{\pgfqpoint{1.197527in}{0.262616in}}%
\pgfpathcurveto{\pgfqpoint{1.192592in}{0.267551in}}{\pgfqpoint{1.185897in}{0.270325in}}{\pgfqpoint{1.178917in}{0.270325in}}%
\pgfpathcurveto{\pgfqpoint{1.171937in}{0.270325in}}{\pgfqpoint{1.165242in}{0.267551in}}{\pgfqpoint{1.160306in}{0.262616in}}%
\pgfpathcurveto{\pgfqpoint{1.155370in}{0.257680in}}{\pgfqpoint{1.152597in}{0.250985in}}{\pgfqpoint{1.152597in}{0.244005in}}%
\pgfpathcurveto{\pgfqpoint{1.152597in}{0.237025in}}{\pgfqpoint{1.155370in}{0.230330in}}{\pgfqpoint{1.160306in}{0.225394in}}%
\pgfpathcurveto{\pgfqpoint{1.165242in}{0.220459in}}{\pgfqpoint{1.171937in}{0.217685in}}{\pgfqpoint{1.178917in}{0.217685in}}%
\pgfpathclose%
\pgfusepath{stroke,fill}%
\end{pgfscope}%
\begin{pgfscope}%
\pgfpathrectangle{\pgfqpoint{0.516000in}{0.058400in}}{\pgfqpoint{1.591000in}{1.086240in}} %
\pgfusepath{clip}%
\pgfsetbuttcap%
\pgfsetroundjoin%
\definecolor{currentfill}{rgb}{0.090196,0.745098,0.811765}%
\pgfsetfillcolor{currentfill}%
\pgfsetlinewidth{1.138252pt}%
\definecolor{currentstroke}{rgb}{0.090196,0.745098,0.811765}%
\pgfsetstrokecolor{currentstroke}%
\pgfsetdash{}{0pt}%
\pgfpathmoveto{\pgfqpoint{1.444083in}{0.295148in}}%
\pgfpathcurveto{\pgfqpoint{1.451063in}{0.295148in}}{\pgfqpoint{1.457758in}{0.297921in}}{\pgfqpoint{1.462694in}{0.302857in}}%
\pgfpathcurveto{\pgfqpoint{1.467630in}{0.307793in}}{\pgfqpoint{1.470403in}{0.314488in}}{\pgfqpoint{1.470403in}{0.321468in}}%
\pgfpathcurveto{\pgfqpoint{1.470403in}{0.328448in}}{\pgfqpoint{1.467630in}{0.335143in}}{\pgfqpoint{1.462694in}{0.340078in}}%
\pgfpathcurveto{\pgfqpoint{1.457758in}{0.345014in}}{\pgfqpoint{1.451063in}{0.347787in}}{\pgfqpoint{1.444083in}{0.347787in}}%
\pgfpathcurveto{\pgfqpoint{1.437103in}{0.347787in}}{\pgfqpoint{1.430408in}{0.345014in}}{\pgfqpoint{1.425473in}{0.340078in}}%
\pgfpathcurveto{\pgfqpoint{1.420537in}{0.335143in}}{\pgfqpoint{1.417764in}{0.328448in}}{\pgfqpoint{1.417764in}{0.321468in}}%
\pgfpathcurveto{\pgfqpoint{1.417764in}{0.314488in}}{\pgfqpoint{1.420537in}{0.307793in}}{\pgfqpoint{1.425473in}{0.302857in}}%
\pgfpathcurveto{\pgfqpoint{1.430408in}{0.297921in}}{\pgfqpoint{1.437103in}{0.295148in}}{\pgfqpoint{1.444083in}{0.295148in}}%
\pgfpathclose%
\pgfusepath{stroke,fill}%
\end{pgfscope}%
\begin{pgfscope}%
\pgfpathrectangle{\pgfqpoint{0.516000in}{0.058400in}}{\pgfqpoint{1.591000in}{1.086240in}} %
\pgfusepath{clip}%
\pgfsetbuttcap%
\pgfsetroundjoin%
\definecolor{currentfill}{rgb}{0.090196,0.745098,0.811765}%
\pgfsetfillcolor{currentfill}%
\pgfsetlinewidth{1.138252pt}%
\definecolor{currentstroke}{rgb}{0.090196,0.745098,0.811765}%
\pgfsetstrokecolor{currentstroke}%
\pgfsetdash{}{0pt}%
\pgfpathmoveto{\pgfqpoint{1.709250in}{0.396161in}}%
\pgfpathcurveto{\pgfqpoint{1.716230in}{0.396161in}}{\pgfqpoint{1.722925in}{0.398934in}}{\pgfqpoint{1.727861in}{0.403869in}}%
\pgfpathcurveto{\pgfqpoint{1.732796in}{0.408805in}}{\pgfqpoint{1.735570in}{0.415500in}}{\pgfqpoint{1.735570in}{0.422480in}}%
\pgfpathcurveto{\pgfqpoint{1.735570in}{0.429460in}}{\pgfqpoint{1.732796in}{0.436155in}}{\pgfqpoint{1.727861in}{0.441091in}}%
\pgfpathcurveto{\pgfqpoint{1.722925in}{0.446027in}}{\pgfqpoint{1.716230in}{0.448800in}}{\pgfqpoint{1.709250in}{0.448800in}}%
\pgfpathcurveto{\pgfqpoint{1.702270in}{0.448800in}}{\pgfqpoint{1.695575in}{0.446027in}}{\pgfqpoint{1.690639in}{0.441091in}}%
\pgfpathcurveto{\pgfqpoint{1.685704in}{0.436155in}}{\pgfqpoint{1.682930in}{0.429460in}}{\pgfqpoint{1.682930in}{0.422480in}}%
\pgfpathcurveto{\pgfqpoint{1.682930in}{0.415500in}}{\pgfqpoint{1.685704in}{0.408805in}}{\pgfqpoint{1.690639in}{0.403869in}}%
\pgfpathcurveto{\pgfqpoint{1.695575in}{0.398934in}}{\pgfqpoint{1.702270in}{0.396161in}}{\pgfqpoint{1.709250in}{0.396161in}}%
\pgfpathclose%
\pgfusepath{stroke,fill}%
\end{pgfscope}%
\begin{pgfscope}%
\pgfpathrectangle{\pgfqpoint{0.516000in}{0.058400in}}{\pgfqpoint{1.591000in}{1.086240in}} %
\pgfusepath{clip}%
\pgfsetbuttcap%
\pgfsetroundjoin%
\definecolor{currentfill}{rgb}{0.090196,0.745098,0.811765}%
\pgfsetfillcolor{currentfill}%
\pgfsetlinewidth{1.138252pt}%
\definecolor{currentstroke}{rgb}{0.090196,0.745098,0.811765}%
\pgfsetstrokecolor{currentstroke}%
\pgfsetdash{}{0pt}%
\pgfpathmoveto{\pgfqpoint{1.974417in}{0.488870in}}%
\pgfpathcurveto{\pgfqpoint{1.981397in}{0.488870in}}{\pgfqpoint{1.988092in}{0.491644in}}{\pgfqpoint{1.993027in}{0.496579in}}%
\pgfpathcurveto{\pgfqpoint{1.997963in}{0.501515in}}{\pgfqpoint{2.000736in}{0.508210in}}{\pgfqpoint{2.000736in}{0.515190in}}%
\pgfpathcurveto{\pgfqpoint{2.000736in}{0.522170in}}{\pgfqpoint{1.997963in}{0.528865in}}{\pgfqpoint{1.993027in}{0.533801in}}%
\pgfpathcurveto{\pgfqpoint{1.988092in}{0.538736in}}{\pgfqpoint{1.981397in}{0.541510in}}{\pgfqpoint{1.974417in}{0.541510in}}%
\pgfpathcurveto{\pgfqpoint{1.967437in}{0.541510in}}{\pgfqpoint{1.960742in}{0.538736in}}{\pgfqpoint{1.955806in}{0.533801in}}%
\pgfpathcurveto{\pgfqpoint{1.950870in}{0.528865in}}{\pgfqpoint{1.948097in}{0.522170in}}{\pgfqpoint{1.948097in}{0.515190in}}%
\pgfpathcurveto{\pgfqpoint{1.948097in}{0.508210in}}{\pgfqpoint{1.950870in}{0.501515in}}{\pgfqpoint{1.955806in}{0.496579in}}%
\pgfpathcurveto{\pgfqpoint{1.960742in}{0.491644in}}{\pgfqpoint{1.967437in}{0.488870in}}{\pgfqpoint{1.974417in}{0.488870in}}%
\pgfpathclose%
\pgfusepath{stroke,fill}%
\end{pgfscope}%
\begin{pgfscope}%
\pgfpathrectangle{\pgfqpoint{0.516000in}{0.058400in}}{\pgfqpoint{1.591000in}{1.086240in}} %
\pgfusepath{clip}%
\pgfsetroundcap%
\pgfsetroundjoin%
\pgfsetlinewidth{1.517670pt}%
\definecolor{currentstroke}{rgb}{0.090196,0.745098,0.811765}%
\pgfsetstrokecolor{currentstroke}%
\pgfsetdash{}{0pt}%
\pgfpathmoveto{\pgfqpoint{0.648583in}{0.104511in}}%
\pgfpathlineto{\pgfqpoint{0.913750in}{0.157584in}}%
\pgfpathlineto{\pgfqpoint{1.178917in}{0.244005in}}%
\pgfpathlineto{\pgfqpoint{1.444083in}{0.321468in}}%
\pgfpathlineto{\pgfqpoint{1.709250in}{0.422480in}}%
\pgfpathlineto{\pgfqpoint{1.974417in}{0.515190in}}%
\pgfusepath{stroke}%
\end{pgfscope}%
\begin{pgfscope}%
\pgfpathrectangle{\pgfqpoint{0.516000in}{0.058400in}}{\pgfqpoint{1.591000in}{1.086240in}} %
\pgfusepath{clip}%
\pgfsetroundcap%
\pgfsetroundjoin%
\pgfsetlinewidth{2.168100pt}%
\definecolor{currentstroke}{rgb}{0.090196,0.745098,0.811765}%
\pgfsetstrokecolor{currentstroke}%
\pgfsetdash{}{0pt}%
\pgfpathmoveto{\pgfqpoint{0.648583in}{0.102919in}}%
\pgfpathlineto{\pgfqpoint{0.648583in}{0.107305in}}%
\pgfusepath{stroke}%
\end{pgfscope}%
\begin{pgfscope}%
\pgfpathrectangle{\pgfqpoint{0.516000in}{0.058400in}}{\pgfqpoint{1.591000in}{1.086240in}} %
\pgfusepath{clip}%
\pgfsetroundcap%
\pgfsetroundjoin%
\pgfsetlinewidth{2.168100pt}%
\definecolor{currentstroke}{rgb}{0.090196,0.745098,0.811765}%
\pgfsetstrokecolor{currentstroke}%
\pgfsetdash{}{0pt}%
\pgfpathmoveto{\pgfqpoint{0.913750in}{0.151476in}}%
\pgfpathlineto{\pgfqpoint{0.913750in}{0.162509in}}%
\pgfusepath{stroke}%
\end{pgfscope}%
\begin{pgfscope}%
\pgfpathrectangle{\pgfqpoint{0.516000in}{0.058400in}}{\pgfqpoint{1.591000in}{1.086240in}} %
\pgfusepath{clip}%
\pgfsetroundcap%
\pgfsetroundjoin%
\pgfsetlinewidth{2.168100pt}%
\definecolor{currentstroke}{rgb}{0.090196,0.745098,0.811765}%
\pgfsetstrokecolor{currentstroke}%
\pgfsetdash{}{0pt}%
\pgfpathmoveto{\pgfqpoint{1.178917in}{0.238758in}}%
\pgfpathlineto{\pgfqpoint{1.178917in}{0.248227in}}%
\pgfusepath{stroke}%
\end{pgfscope}%
\begin{pgfscope}%
\pgfpathrectangle{\pgfqpoint{0.516000in}{0.058400in}}{\pgfqpoint{1.591000in}{1.086240in}} %
\pgfusepath{clip}%
\pgfsetroundcap%
\pgfsetroundjoin%
\pgfsetlinewidth{2.168100pt}%
\definecolor{currentstroke}{rgb}{0.090196,0.745098,0.811765}%
\pgfsetstrokecolor{currentstroke}%
\pgfsetdash{}{0pt}%
\pgfpathmoveto{\pgfqpoint{1.444083in}{0.313300in}}%
\pgfpathlineto{\pgfqpoint{1.444083in}{0.344455in}}%
\pgfusepath{stroke}%
\end{pgfscope}%
\begin{pgfscope}%
\pgfpathrectangle{\pgfqpoint{0.516000in}{0.058400in}}{\pgfqpoint{1.591000in}{1.086240in}} %
\pgfusepath{clip}%
\pgfsetroundcap%
\pgfsetroundjoin%
\pgfsetlinewidth{2.168100pt}%
\definecolor{currentstroke}{rgb}{0.090196,0.745098,0.811765}%
\pgfsetstrokecolor{currentstroke}%
\pgfsetdash{}{0pt}%
\pgfpathmoveto{\pgfqpoint{1.709250in}{0.400230in}}%
\pgfpathlineto{\pgfqpoint{1.709250in}{0.438682in}}%
\pgfusepath{stroke}%
\end{pgfscope}%
\begin{pgfscope}%
\pgfpathrectangle{\pgfqpoint{0.516000in}{0.058400in}}{\pgfqpoint{1.591000in}{1.086240in}} %
\pgfusepath{clip}%
\pgfsetroundcap%
\pgfsetroundjoin%
\pgfsetlinewidth{2.168100pt}%
\definecolor{currentstroke}{rgb}{0.090196,0.745098,0.811765}%
\pgfsetstrokecolor{currentstroke}%
\pgfsetdash{}{0pt}%
\pgfpathmoveto{\pgfqpoint{1.974417in}{0.492070in}}%
\pgfpathlineto{\pgfqpoint{1.974417in}{0.535557in}}%
\pgfusepath{stroke}%
\end{pgfscope}%
\begin{pgfscope}%
\pgfsetrectcap%
\pgfsetmiterjoin%
\pgfsetlinewidth{1.003750pt}%
\definecolor{currentstroke}{rgb}{0.800000,0.800000,0.800000}%
\pgfsetstrokecolor{currentstroke}%
\pgfsetdash{}{0pt}%
\pgfpathmoveto{\pgfqpoint{0.516000in}{0.058400in}}%
\pgfpathlineto{\pgfqpoint{0.516000in}{1.144640in}}%
\pgfusepath{stroke}%
\end{pgfscope}%
\begin{pgfscope}%
\pgfsetrectcap%
\pgfsetmiterjoin%
\pgfsetlinewidth{1.003750pt}%
\definecolor{currentstroke}{rgb}{0.800000,0.800000,0.800000}%
\pgfsetstrokecolor{currentstroke}%
\pgfsetdash{}{0pt}%
\pgfpathmoveto{\pgfqpoint{2.107000in}{0.058400in}}%
\pgfpathlineto{\pgfqpoint{2.107000in}{1.144640in}}%
\pgfusepath{stroke}%
\end{pgfscope}%
\begin{pgfscope}%
\pgfsetrectcap%
\pgfsetmiterjoin%
\pgfsetlinewidth{1.003750pt}%
\definecolor{currentstroke}{rgb}{0.800000,0.800000,0.800000}%
\pgfsetstrokecolor{currentstroke}%
\pgfsetdash{}{0pt}%
\pgfpathmoveto{\pgfqpoint{0.516000in}{0.058400in}}%
\pgfpathlineto{\pgfqpoint{2.107000in}{0.058400in}}%
\pgfusepath{stroke}%
\end{pgfscope}%
\begin{pgfscope}%
\pgfsetrectcap%
\pgfsetmiterjoin%
\pgfsetlinewidth{1.003750pt}%
\definecolor{currentstroke}{rgb}{0.800000,0.800000,0.800000}%
\pgfsetstrokecolor{currentstroke}%
\pgfsetdash{}{0pt}%
\pgfpathmoveto{\pgfqpoint{0.516000in}{1.144640in}}%
\pgfpathlineto{\pgfqpoint{2.107000in}{1.144640in}}%
\pgfusepath{stroke}%
\end{pgfscope}%
\begin{pgfscope}%
\pgfpathrectangle{\pgfqpoint{0.516000in}{0.058400in}}{\pgfqpoint{1.591000in}{1.086240in}} %
\pgfusepath{clip}%
\pgfsetbuttcap%
\pgfsetroundjoin%
\definecolor{currentfill}{rgb}{0.498039,0.498039,0.498039}%
\pgfsetfillcolor{currentfill}%
\pgfsetlinewidth{1.138252pt}%
\definecolor{currentstroke}{rgb}{0.498039,0.498039,0.498039}%
\pgfsetstrokecolor{currentstroke}%
\pgfsetdash{}{0pt}%
\pgfpathmoveto{\pgfqpoint{0.648583in}{0.078766in}}%
\pgfpathcurveto{\pgfqpoint{0.655563in}{0.078766in}}{\pgfqpoint{0.662258in}{0.081540in}}{\pgfqpoint{0.667194in}{0.086475in}}%
\pgfpathcurveto{\pgfqpoint{0.672130in}{0.091411in}}{\pgfqpoint{0.674903in}{0.098106in}}{\pgfqpoint{0.674903in}{0.105086in}}%
\pgfpathcurveto{\pgfqpoint{0.674903in}{0.112066in}}{\pgfqpoint{0.672130in}{0.118761in}}{\pgfqpoint{0.667194in}{0.123697in}}%
\pgfpathcurveto{\pgfqpoint{0.662258in}{0.128632in}}{\pgfqpoint{0.655563in}{0.131406in}}{\pgfqpoint{0.648583in}{0.131406in}}%
\pgfpathcurveto{\pgfqpoint{0.641603in}{0.131406in}}{\pgfqpoint{0.634908in}{0.128632in}}{\pgfqpoint{0.629973in}{0.123697in}}%
\pgfpathcurveto{\pgfqpoint{0.625037in}{0.118761in}}{\pgfqpoint{0.622264in}{0.112066in}}{\pgfqpoint{0.622264in}{0.105086in}}%
\pgfpathcurveto{\pgfqpoint{0.622264in}{0.098106in}}{\pgfqpoint{0.625037in}{0.091411in}}{\pgfqpoint{0.629973in}{0.086475in}}%
\pgfpathcurveto{\pgfqpoint{0.634908in}{0.081540in}}{\pgfqpoint{0.641603in}{0.078766in}}{\pgfqpoint{0.648583in}{0.078766in}}%
\pgfpathclose%
\pgfusepath{stroke,fill}%
\end{pgfscope}%
\begin{pgfscope}%
\pgfpathrectangle{\pgfqpoint{0.516000in}{0.058400in}}{\pgfqpoint{1.591000in}{1.086240in}} %
\pgfusepath{clip}%
\pgfsetbuttcap%
\pgfsetroundjoin%
\definecolor{currentfill}{rgb}{0.498039,0.498039,0.498039}%
\pgfsetfillcolor{currentfill}%
\pgfsetlinewidth{1.138252pt}%
\definecolor{currentstroke}{rgb}{0.498039,0.498039,0.498039}%
\pgfsetstrokecolor{currentstroke}%
\pgfsetdash{}{0pt}%
\pgfpathmoveto{\pgfqpoint{0.913750in}{0.121597in}}%
\pgfpathcurveto{\pgfqpoint{0.920730in}{0.121597in}}{\pgfqpoint{0.927425in}{0.124370in}}{\pgfqpoint{0.932361in}{0.129306in}}%
\pgfpathcurveto{\pgfqpoint{0.937296in}{0.134241in}}{\pgfqpoint{0.940070in}{0.140936in}}{\pgfqpoint{0.940070in}{0.147916in}}%
\pgfpathcurveto{\pgfqpoint{0.940070in}{0.154896in}}{\pgfqpoint{0.937296in}{0.161591in}}{\pgfqpoint{0.932361in}{0.166527in}}%
\pgfpathcurveto{\pgfqpoint{0.927425in}{0.171463in}}{\pgfqpoint{0.920730in}{0.174236in}}{\pgfqpoint{0.913750in}{0.174236in}}%
\pgfpathcurveto{\pgfqpoint{0.906770in}{0.174236in}}{\pgfqpoint{0.900075in}{0.171463in}}{\pgfqpoint{0.895139in}{0.166527in}}%
\pgfpathcurveto{\pgfqpoint{0.890204in}{0.161591in}}{\pgfqpoint{0.887430in}{0.154896in}}{\pgfqpoint{0.887430in}{0.147916in}}%
\pgfpathcurveto{\pgfqpoint{0.887430in}{0.140936in}}{\pgfqpoint{0.890204in}{0.134241in}}{\pgfqpoint{0.895139in}{0.129306in}}%
\pgfpathcurveto{\pgfqpoint{0.900075in}{0.124370in}}{\pgfqpoint{0.906770in}{0.121597in}}{\pgfqpoint{0.913750in}{0.121597in}}%
\pgfpathclose%
\pgfusepath{stroke,fill}%
\end{pgfscope}%
\begin{pgfscope}%
\pgfpathrectangle{\pgfqpoint{0.516000in}{0.058400in}}{\pgfqpoint{1.591000in}{1.086240in}} %
\pgfusepath{clip}%
\pgfsetbuttcap%
\pgfsetroundjoin%
\definecolor{currentfill}{rgb}{0.498039,0.498039,0.498039}%
\pgfsetfillcolor{currentfill}%
\pgfsetlinewidth{1.138252pt}%
\definecolor{currentstroke}{rgb}{0.498039,0.498039,0.498039}%
\pgfsetstrokecolor{currentstroke}%
\pgfsetdash{}{0pt}%
\pgfpathmoveto{\pgfqpoint{1.178917in}{0.198134in}}%
\pgfpathcurveto{\pgfqpoint{1.185897in}{0.198134in}}{\pgfqpoint{1.192592in}{0.200907in}}{\pgfqpoint{1.197527in}{0.205843in}}%
\pgfpathcurveto{\pgfqpoint{1.202463in}{0.210779in}}{\pgfqpoint{1.205236in}{0.217474in}}{\pgfqpoint{1.205236in}{0.224454in}}%
\pgfpathcurveto{\pgfqpoint{1.205236in}{0.231434in}}{\pgfqpoint{1.202463in}{0.238129in}}{\pgfqpoint{1.197527in}{0.243064in}}%
\pgfpathcurveto{\pgfqpoint{1.192592in}{0.248000in}}{\pgfqpoint{1.185897in}{0.250773in}}{\pgfqpoint{1.178917in}{0.250773in}}%
\pgfpathcurveto{\pgfqpoint{1.171937in}{0.250773in}}{\pgfqpoint{1.165242in}{0.248000in}}{\pgfqpoint{1.160306in}{0.243064in}}%
\pgfpathcurveto{\pgfqpoint{1.155370in}{0.238129in}}{\pgfqpoint{1.152597in}{0.231434in}}{\pgfqpoint{1.152597in}{0.224454in}}%
\pgfpathcurveto{\pgfqpoint{1.152597in}{0.217474in}}{\pgfqpoint{1.155370in}{0.210779in}}{\pgfqpoint{1.160306in}{0.205843in}}%
\pgfpathcurveto{\pgfqpoint{1.165242in}{0.200907in}}{\pgfqpoint{1.171937in}{0.198134in}}{\pgfqpoint{1.178917in}{0.198134in}}%
\pgfpathclose%
\pgfusepath{stroke,fill}%
\end{pgfscope}%
\begin{pgfscope}%
\pgfpathrectangle{\pgfqpoint{0.516000in}{0.058400in}}{\pgfqpoint{1.591000in}{1.086240in}} %
\pgfusepath{clip}%
\pgfsetbuttcap%
\pgfsetroundjoin%
\definecolor{currentfill}{rgb}{0.498039,0.498039,0.498039}%
\pgfsetfillcolor{currentfill}%
\pgfsetlinewidth{1.138252pt}%
\definecolor{currentstroke}{rgb}{0.498039,0.498039,0.498039}%
\pgfsetstrokecolor{currentstroke}%
\pgfsetdash{}{0pt}%
\pgfpathmoveto{\pgfqpoint{1.444083in}{0.268387in}}%
\pgfpathcurveto{\pgfqpoint{1.451063in}{0.268387in}}{\pgfqpoint{1.457758in}{0.271160in}}{\pgfqpoint{1.462694in}{0.276096in}}%
\pgfpathcurveto{\pgfqpoint{1.467630in}{0.281031in}}{\pgfqpoint{1.470403in}{0.287727in}}{\pgfqpoint{1.470403in}{0.294707in}}%
\pgfpathcurveto{\pgfqpoint{1.470403in}{0.301687in}}{\pgfqpoint{1.467630in}{0.308382in}}{\pgfqpoint{1.462694in}{0.313317in}}%
\pgfpathcurveto{\pgfqpoint{1.457758in}{0.318253in}}{\pgfqpoint{1.451063in}{0.321026in}}{\pgfqpoint{1.444083in}{0.321026in}}%
\pgfpathcurveto{\pgfqpoint{1.437103in}{0.321026in}}{\pgfqpoint{1.430408in}{0.318253in}}{\pgfqpoint{1.425473in}{0.313317in}}%
\pgfpathcurveto{\pgfqpoint{1.420537in}{0.308382in}}{\pgfqpoint{1.417764in}{0.301687in}}{\pgfqpoint{1.417764in}{0.294707in}}%
\pgfpathcurveto{\pgfqpoint{1.417764in}{0.287727in}}{\pgfqpoint{1.420537in}{0.281031in}}{\pgfqpoint{1.425473in}{0.276096in}}%
\pgfpathcurveto{\pgfqpoint{1.430408in}{0.271160in}}{\pgfqpoint{1.437103in}{0.268387in}}{\pgfqpoint{1.444083in}{0.268387in}}%
\pgfpathclose%
\pgfusepath{stroke,fill}%
\end{pgfscope}%
\begin{pgfscope}%
\pgfpathrectangle{\pgfqpoint{0.516000in}{0.058400in}}{\pgfqpoint{1.591000in}{1.086240in}} %
\pgfusepath{clip}%
\pgfsetbuttcap%
\pgfsetroundjoin%
\definecolor{currentfill}{rgb}{0.498039,0.498039,0.498039}%
\pgfsetfillcolor{currentfill}%
\pgfsetlinewidth{1.138252pt}%
\definecolor{currentstroke}{rgb}{0.498039,0.498039,0.498039}%
\pgfsetstrokecolor{currentstroke}%
\pgfsetdash{}{0pt}%
\pgfpathmoveto{\pgfqpoint{1.709250in}{0.362646in}}%
\pgfpathcurveto{\pgfqpoint{1.716230in}{0.362646in}}{\pgfqpoint{1.722925in}{0.365419in}}{\pgfqpoint{1.727861in}{0.370355in}}%
\pgfpathcurveto{\pgfqpoint{1.732796in}{0.375290in}}{\pgfqpoint{1.735570in}{0.381985in}}{\pgfqpoint{1.735570in}{0.388965in}}%
\pgfpathcurveto{\pgfqpoint{1.735570in}{0.395945in}}{\pgfqpoint{1.732796in}{0.402640in}}{\pgfqpoint{1.727861in}{0.407576in}}%
\pgfpathcurveto{\pgfqpoint{1.722925in}{0.412512in}}{\pgfqpoint{1.716230in}{0.415285in}}{\pgfqpoint{1.709250in}{0.415285in}}%
\pgfpathcurveto{\pgfqpoint{1.702270in}{0.415285in}}{\pgfqpoint{1.695575in}{0.412512in}}{\pgfqpoint{1.690639in}{0.407576in}}%
\pgfpathcurveto{\pgfqpoint{1.685704in}{0.402640in}}{\pgfqpoint{1.682930in}{0.395945in}}{\pgfqpoint{1.682930in}{0.388965in}}%
\pgfpathcurveto{\pgfqpoint{1.682930in}{0.381985in}}{\pgfqpoint{1.685704in}{0.375290in}}{\pgfqpoint{1.690639in}{0.370355in}}%
\pgfpathcurveto{\pgfqpoint{1.695575in}{0.365419in}}{\pgfqpoint{1.702270in}{0.362646in}}{\pgfqpoint{1.709250in}{0.362646in}}%
\pgfpathclose%
\pgfusepath{stroke,fill}%
\end{pgfscope}%
\begin{pgfscope}%
\pgfpathrectangle{\pgfqpoint{0.516000in}{0.058400in}}{\pgfqpoint{1.591000in}{1.086240in}} %
\pgfusepath{clip}%
\pgfsetbuttcap%
\pgfsetroundjoin%
\definecolor{currentfill}{rgb}{0.498039,0.498039,0.498039}%
\pgfsetfillcolor{currentfill}%
\pgfsetlinewidth{1.138252pt}%
\definecolor{currentstroke}{rgb}{0.498039,0.498039,0.498039}%
\pgfsetstrokecolor{currentstroke}%
\pgfsetdash{}{0pt}%
\pgfpathmoveto{\pgfqpoint{1.974417in}{0.399702in}}%
\pgfpathcurveto{\pgfqpoint{1.981397in}{0.399702in}}{\pgfqpoint{1.988092in}{0.402475in}}{\pgfqpoint{1.993027in}{0.407411in}}%
\pgfpathcurveto{\pgfqpoint{1.997963in}{0.412347in}}{\pgfqpoint{2.000736in}{0.419042in}}{\pgfqpoint{2.000736in}{0.426022in}}%
\pgfpathcurveto{\pgfqpoint{2.000736in}{0.433002in}}{\pgfqpoint{1.997963in}{0.439697in}}{\pgfqpoint{1.993027in}{0.444633in}}%
\pgfpathcurveto{\pgfqpoint{1.988092in}{0.449568in}}{\pgfqpoint{1.981397in}{0.452341in}}{\pgfqpoint{1.974417in}{0.452341in}}%
\pgfpathcurveto{\pgfqpoint{1.967437in}{0.452341in}}{\pgfqpoint{1.960742in}{0.449568in}}{\pgfqpoint{1.955806in}{0.444633in}}%
\pgfpathcurveto{\pgfqpoint{1.950870in}{0.439697in}}{\pgfqpoint{1.948097in}{0.433002in}}{\pgfqpoint{1.948097in}{0.426022in}}%
\pgfpathcurveto{\pgfqpoint{1.948097in}{0.419042in}}{\pgfqpoint{1.950870in}{0.412347in}}{\pgfqpoint{1.955806in}{0.407411in}}%
\pgfpathcurveto{\pgfqpoint{1.960742in}{0.402475in}}{\pgfqpoint{1.967437in}{0.399702in}}{\pgfqpoint{1.974417in}{0.399702in}}%
\pgfpathclose%
\pgfusepath{stroke,fill}%
\end{pgfscope}%
\begin{pgfscope}%
\pgfpathrectangle{\pgfqpoint{0.516000in}{0.058400in}}{\pgfqpoint{1.591000in}{1.086240in}} %
\pgfusepath{clip}%
\pgfsetroundcap%
\pgfsetroundjoin%
\pgfsetlinewidth{1.517670pt}%
\definecolor{currentstroke}{rgb}{0.498039,0.498039,0.498039}%
\pgfsetstrokecolor{currentstroke}%
\pgfsetdash{}{0pt}%
\pgfpathmoveto{\pgfqpoint{0.648583in}{0.105086in}}%
\pgfpathlineto{\pgfqpoint{0.913750in}{0.147916in}}%
\pgfpathlineto{\pgfqpoint{1.178917in}{0.224454in}}%
\pgfpathlineto{\pgfqpoint{1.444083in}{0.294707in}}%
\pgfpathlineto{\pgfqpoint{1.709250in}{0.388965in}}%
\pgfpathlineto{\pgfqpoint{1.974417in}{0.426022in}}%
\pgfusepath{stroke}%
\end{pgfscope}%
\begin{pgfscope}%
\pgfpathrectangle{\pgfqpoint{0.516000in}{0.058400in}}{\pgfqpoint{1.591000in}{1.086240in}} %
\pgfusepath{clip}%
\pgfsetroundcap%
\pgfsetroundjoin%
\pgfsetlinewidth{2.168100pt}%
\definecolor{currentstroke}{rgb}{0.498039,0.498039,0.498039}%
\pgfsetstrokecolor{currentstroke}%
\pgfsetdash{}{0pt}%
\pgfpathmoveto{\pgfqpoint{0.648583in}{0.102301in}}%
\pgfpathlineto{\pgfqpoint{0.648583in}{0.109179in}}%
\pgfusepath{stroke}%
\end{pgfscope}%
\begin{pgfscope}%
\pgfpathrectangle{\pgfqpoint{0.516000in}{0.058400in}}{\pgfqpoint{1.591000in}{1.086240in}} %
\pgfusepath{clip}%
\pgfsetroundcap%
\pgfsetroundjoin%
\pgfsetlinewidth{2.168100pt}%
\definecolor{currentstroke}{rgb}{0.498039,0.498039,0.498039}%
\pgfsetstrokecolor{currentstroke}%
\pgfsetdash{}{0pt}%
\pgfpathmoveto{\pgfqpoint{0.913750in}{0.145276in}}%
\pgfpathlineto{\pgfqpoint{0.913750in}{0.151720in}}%
\pgfusepath{stroke}%
\end{pgfscope}%
\begin{pgfscope}%
\pgfpathrectangle{\pgfqpoint{0.516000in}{0.058400in}}{\pgfqpoint{1.591000in}{1.086240in}} %
\pgfusepath{clip}%
\pgfsetroundcap%
\pgfsetroundjoin%
\pgfsetlinewidth{2.168100pt}%
\definecolor{currentstroke}{rgb}{0.498039,0.498039,0.498039}%
\pgfsetstrokecolor{currentstroke}%
\pgfsetdash{}{0pt}%
\pgfpathmoveto{\pgfqpoint{1.178917in}{0.214443in}}%
\pgfpathlineto{\pgfqpoint{1.178917in}{0.243883in}}%
\pgfusepath{stroke}%
\end{pgfscope}%
\begin{pgfscope}%
\pgfpathrectangle{\pgfqpoint{0.516000in}{0.058400in}}{\pgfqpoint{1.591000in}{1.086240in}} %
\pgfusepath{clip}%
\pgfsetroundcap%
\pgfsetroundjoin%
\pgfsetlinewidth{2.168100pt}%
\definecolor{currentstroke}{rgb}{0.498039,0.498039,0.498039}%
\pgfsetstrokecolor{currentstroke}%
\pgfsetdash{}{0pt}%
\pgfpathmoveto{\pgfqpoint{1.444083in}{0.282187in}}%
\pgfpathlineto{\pgfqpoint{1.444083in}{0.314380in}}%
\pgfusepath{stroke}%
\end{pgfscope}%
\begin{pgfscope}%
\pgfpathrectangle{\pgfqpoint{0.516000in}{0.058400in}}{\pgfqpoint{1.591000in}{1.086240in}} %
\pgfusepath{clip}%
\pgfsetroundcap%
\pgfsetroundjoin%
\pgfsetlinewidth{2.168100pt}%
\definecolor{currentstroke}{rgb}{0.498039,0.498039,0.498039}%
\pgfsetstrokecolor{currentstroke}%
\pgfsetdash{}{0pt}%
\pgfpathmoveto{\pgfqpoint{1.709250in}{0.368211in}}%
\pgfpathlineto{\pgfqpoint{1.709250in}{0.404020in}}%
\pgfusepath{stroke}%
\end{pgfscope}%
\begin{pgfscope}%
\pgfpathrectangle{\pgfqpoint{0.516000in}{0.058400in}}{\pgfqpoint{1.591000in}{1.086240in}} %
\pgfusepath{clip}%
\pgfsetroundcap%
\pgfsetroundjoin%
\pgfsetlinewidth{2.168100pt}%
\definecolor{currentstroke}{rgb}{0.498039,0.498039,0.498039}%
\pgfsetstrokecolor{currentstroke}%
\pgfsetdash{}{0pt}%
\pgfpathmoveto{\pgfqpoint{1.974417in}{0.400283in}}%
\pgfpathlineto{\pgfqpoint{1.974417in}{0.451076in}}%
\pgfusepath{stroke}%
\end{pgfscope}%
\begin{pgfscope}%
\pgfpathrectangle{\pgfqpoint{0.516000in}{0.058400in}}{\pgfqpoint{1.591000in}{1.086240in}} %
\pgfusepath{clip}%
\pgfsetbuttcap%
\pgfsetroundjoin%
\pgfsetlinewidth{2.007500pt}%
\definecolor{currentstroke}{rgb}{0.890196,0.466667,0.760784}%
\pgfsetstrokecolor{currentstroke}%
\pgfsetdash{{7.400000pt}{3.200000pt}}{0.000000pt}%
\pgfpathmoveto{\pgfqpoint{0.516000in}{0.959425in}}%
\pgfpathlineto{\pgfqpoint{2.107000in}{0.959425in}}%
\pgfusepath{stroke}%
\end{pgfscope}%
\end{pgfpicture}%
\makeatother%
\endgroup%

%% file: images/plots/exec_time_Chemical-I.pgf
\begingroup%
\makeatletter%
\begin{pgfpicture}%
\pgfpathrectangle{\pgfpointorigin}{\pgfqpoint{2.350000in}{1.500000in}}%
\pgfusepath{use as bounding box, clip}%
\begin{pgfscope}%
\pgfsetbuttcap%
\pgfsetmiterjoin%
\pgfsetlinewidth{0.000000pt}%
\definecolor{currentstroke}{rgb}{0.000000,0.000000,0.000000}%
\pgfsetstrokecolor{currentstroke}%
\pgfsetstrokeopacity{0.000000}%
\pgfsetdash{}{0pt}%
\pgfpathmoveto{\pgfqpoint{0.000000in}{0.000000in}}%
\pgfpathlineto{\pgfqpoint{2.350000in}{0.000000in}}%
\pgfpathlineto{\pgfqpoint{2.350000in}{1.500000in}}%
\pgfpathlineto{\pgfqpoint{0.000000in}{1.500000in}}%
\pgfpathclose%
\pgfusepath{}%
\end{pgfscope}%
\begin{pgfscope}%
\pgfsetbuttcap%
\pgfsetmiterjoin%
\pgfsetlinewidth{0.000000pt}%
\definecolor{currentstroke}{rgb}{0.000000,0.000000,0.000000}%
\pgfsetstrokecolor{currentstroke}%
\pgfsetstrokeopacity{0.000000}%
\pgfsetdash{}{0pt}%
\pgfpathmoveto{\pgfqpoint{0.681500in}{0.420000in}}%
\pgfpathlineto{\pgfqpoint{2.303000in}{0.420000in}}%
\pgfpathlineto{\pgfqpoint{2.303000in}{1.470000in}}%
\pgfpathlineto{\pgfqpoint{0.681500in}{1.470000in}}%
\pgfpathclose%
\pgfusepath{}%
\end{pgfscope}%
\begin{pgfscope}%
\pgfpathrectangle{\pgfqpoint{0.681500in}{0.420000in}}{\pgfqpoint{1.621500in}{1.050000in}} %
\pgfusepath{clip}%
\pgfsetroundcap%
\pgfsetroundjoin%
\pgfsetlinewidth{0.803000pt}%
\definecolor{currentstroke}{rgb}{0.800000,0.800000,0.800000}%
\pgfsetstrokecolor{currentstroke}%
\pgfsetdash{}{0pt}%
\pgfpathmoveto{\pgfqpoint{0.816625in}{0.420000in}}%
\pgfpathlineto{\pgfqpoint{0.816625in}{1.470000in}}%
\pgfusepath{stroke}%
\end{pgfscope}%
\begin{pgfscope}%
\definecolor{textcolor}{rgb}{0.150000,0.150000,0.150000}%
\pgfsetstrokecolor{textcolor}%
\pgfsetfillcolor{textcolor}%
\pgftext[x=0.816625in,y=0.304722in,,top]{\color{textcolor}\fontsize{8.800000}{10.560000}\selectfont 5}%
\end{pgfscope}%
\begin{pgfscope}%
\pgfpathrectangle{\pgfqpoint{0.681500in}{0.420000in}}{\pgfqpoint{1.621500in}{1.050000in}} %
\pgfusepath{clip}%
\pgfsetroundcap%
\pgfsetroundjoin%
\pgfsetlinewidth{0.803000pt}%
\definecolor{currentstroke}{rgb}{0.800000,0.800000,0.800000}%
\pgfsetstrokecolor{currentstroke}%
\pgfsetdash{}{0pt}%
\pgfpathmoveto{\pgfqpoint{1.086875in}{0.420000in}}%
\pgfpathlineto{\pgfqpoint{1.086875in}{1.470000in}}%
\pgfusepath{stroke}%
\end{pgfscope}%
\begin{pgfscope}%
\definecolor{textcolor}{rgb}{0.150000,0.150000,0.150000}%
\pgfsetstrokecolor{textcolor}%
\pgfsetfillcolor{textcolor}%
\pgftext[x=1.086875in,y=0.304722in,,top]{\color{textcolor}\fontsize{8.800000}{10.560000}\selectfont 10}%
\end{pgfscope}%
\begin{pgfscope}%
\pgfpathrectangle{\pgfqpoint{0.681500in}{0.420000in}}{\pgfqpoint{1.621500in}{1.050000in}} %
\pgfusepath{clip}%
\pgfsetroundcap%
\pgfsetroundjoin%
\pgfsetlinewidth{0.803000pt}%
\definecolor{currentstroke}{rgb}{0.800000,0.800000,0.800000}%
\pgfsetstrokecolor{currentstroke}%
\pgfsetdash{}{0pt}%
\pgfpathmoveto{\pgfqpoint{1.357125in}{0.420000in}}%
\pgfpathlineto{\pgfqpoint{1.357125in}{1.470000in}}%
\pgfusepath{stroke}%
\end{pgfscope}%
\begin{pgfscope}%
\definecolor{textcolor}{rgb}{0.150000,0.150000,0.150000}%
\pgfsetstrokecolor{textcolor}%
\pgfsetfillcolor{textcolor}%
\pgftext[x=1.357125in,y=0.304722in,,top]{\color{textcolor}\fontsize{8.800000}{10.560000}\selectfont 20}%
\end{pgfscope}%
\begin{pgfscope}%
\pgfpathrectangle{\pgfqpoint{0.681500in}{0.420000in}}{\pgfqpoint{1.621500in}{1.050000in}} %
\pgfusepath{clip}%
\pgfsetroundcap%
\pgfsetroundjoin%
\pgfsetlinewidth{0.803000pt}%
\definecolor{currentstroke}{rgb}{0.800000,0.800000,0.800000}%
\pgfsetstrokecolor{currentstroke}%
\pgfsetdash{}{0pt}%
\pgfpathmoveto{\pgfqpoint{1.627375in}{0.420000in}}%
\pgfpathlineto{\pgfqpoint{1.627375in}{1.470000in}}%
\pgfusepath{stroke}%
\end{pgfscope}%
\begin{pgfscope}%
\definecolor{textcolor}{rgb}{0.150000,0.150000,0.150000}%
\pgfsetstrokecolor{textcolor}%
\pgfsetfillcolor{textcolor}%
\pgftext[x=1.627375in,y=0.304722in,,top]{\color{textcolor}\fontsize{8.800000}{10.560000}\selectfont 30}%
\end{pgfscope}%
\begin{pgfscope}%
\pgfpathrectangle{\pgfqpoint{0.681500in}{0.420000in}}{\pgfqpoint{1.621500in}{1.050000in}} %
\pgfusepath{clip}%
\pgfsetroundcap%
\pgfsetroundjoin%
\pgfsetlinewidth{0.803000pt}%
\definecolor{currentstroke}{rgb}{0.800000,0.800000,0.800000}%
\pgfsetstrokecolor{currentstroke}%
\pgfsetdash{}{0pt}%
\pgfpathmoveto{\pgfqpoint{1.897625in}{0.420000in}}%
\pgfpathlineto{\pgfqpoint{1.897625in}{1.470000in}}%
\pgfusepath{stroke}%
\end{pgfscope}%
\begin{pgfscope}%
\definecolor{textcolor}{rgb}{0.150000,0.150000,0.150000}%
\pgfsetstrokecolor{textcolor}%
\pgfsetfillcolor{textcolor}%
\pgftext[x=1.897625in,y=0.304722in,,top]{\color{textcolor}\fontsize{8.800000}{10.560000}\selectfont 40}%
\end{pgfscope}%
\begin{pgfscope}%
\pgfpathrectangle{\pgfqpoint{0.681500in}{0.420000in}}{\pgfqpoint{1.621500in}{1.050000in}} %
\pgfusepath{clip}%
\pgfsetroundcap%
\pgfsetroundjoin%
\pgfsetlinewidth{0.803000pt}%
\definecolor{currentstroke}{rgb}{0.800000,0.800000,0.800000}%
\pgfsetstrokecolor{currentstroke}%
\pgfsetdash{}{0pt}%
\pgfpathmoveto{\pgfqpoint{2.167875in}{0.420000in}}%
\pgfpathlineto{\pgfqpoint{2.167875in}{1.470000in}}%
\pgfusepath{stroke}%
\end{pgfscope}%
\begin{pgfscope}%
\definecolor{textcolor}{rgb}{0.150000,0.150000,0.150000}%
\pgfsetstrokecolor{textcolor}%
\pgfsetfillcolor{textcolor}%
\pgftext[x=2.167875in,y=0.304722in,,top]{\color{textcolor}\fontsize{8.800000}{10.560000}\selectfont 50}%
\end{pgfscope}%
\begin{pgfscope}%
\definecolor{textcolor}{rgb}{0.150000,0.150000,0.150000}%
\pgfsetstrokecolor{textcolor}%
\pgfsetfillcolor{textcolor}%
\pgftext[x=1.492250in,y=0.138056in,,top]{\color{textcolor}\fontsize{9.600000}{11.520000}\selectfont Reduction in \%}%
\end{pgfscope}%
\begin{pgfscope}%
\pgfpathrectangle{\pgfqpoint{0.681500in}{0.420000in}}{\pgfqpoint{1.621500in}{1.050000in}} %
\pgfusepath{clip}%
\pgfsetroundcap%
\pgfsetroundjoin%
\pgfsetlinewidth{0.803000pt}%
\definecolor{currentstroke}{rgb}{0.800000,0.800000,0.800000}%
\pgfsetstrokecolor{currentstroke}%
\pgfsetdash{}{0pt}%
\pgfpathmoveto{\pgfqpoint{0.681500in}{0.420000in}}%
\pgfpathlineto{\pgfqpoint{2.303000in}{0.420000in}}%
\pgfusepath{stroke}%
\end{pgfscope}%
\begin{pgfscope}%
\definecolor{textcolor}{rgb}{0.150000,0.150000,0.150000}%
\pgfsetstrokecolor{textcolor}%
\pgfsetfillcolor{textcolor}%
\pgftext[x=0.192190in,y=0.376597in,left,base]{\color{textcolor}\fontsize{8.800000}{10.560000}\selectfont 00h 00m}%
\end{pgfscope}%
\begin{pgfscope}%
\pgfpathrectangle{\pgfqpoint{0.681500in}{0.420000in}}{\pgfqpoint{1.621500in}{1.050000in}} %
\pgfusepath{clip}%
\pgfsetroundcap%
\pgfsetroundjoin%
\pgfsetlinewidth{0.803000pt}%
\definecolor{currentstroke}{rgb}{0.800000,0.800000,0.800000}%
\pgfsetstrokecolor{currentstroke}%
\pgfsetdash{}{0pt}%
\pgfpathmoveto{\pgfqpoint{0.681500in}{0.804859in}}%
\pgfpathlineto{\pgfqpoint{2.303000in}{0.804859in}}%
\pgfusepath{stroke}%
\end{pgfscope}%
\begin{pgfscope}%
\definecolor{textcolor}{rgb}{0.150000,0.150000,0.150000}%
\pgfsetstrokecolor{textcolor}%
\pgfsetfillcolor{textcolor}%
\pgftext[x=0.192190in,y=0.761457in,left,base]{\color{textcolor}\fontsize{8.800000}{10.560000}\selectfont 01h 30m}%
\end{pgfscope}%
\begin{pgfscope}%
\pgfpathrectangle{\pgfqpoint{0.681500in}{0.420000in}}{\pgfqpoint{1.621500in}{1.050000in}} %
\pgfusepath{clip}%
\pgfsetroundcap%
\pgfsetroundjoin%
\pgfsetlinewidth{0.803000pt}%
\definecolor{currentstroke}{rgb}{0.800000,0.800000,0.800000}%
\pgfsetstrokecolor{currentstroke}%
\pgfsetdash{}{0pt}%
\pgfpathmoveto{\pgfqpoint{0.681500in}{1.189719in}}%
\pgfpathlineto{\pgfqpoint{2.303000in}{1.189719in}}%
\pgfusepath{stroke}%
\end{pgfscope}%
\begin{pgfscope}%
\definecolor{textcolor}{rgb}{0.150000,0.150000,0.150000}%
\pgfsetstrokecolor{textcolor}%
\pgfsetfillcolor{textcolor}%
\pgftext[x=0.192190in,y=1.146316in,left,base]{\color{textcolor}\fontsize{8.800000}{10.560000}\selectfont 03h 00m}%
\end{pgfscope}%
\begin{pgfscope}%
\definecolor{textcolor}{rgb}{0.150000,0.150000,0.150000}%
\pgfsetstrokecolor{textcolor}%
\pgfsetfillcolor{textcolor}%
\pgftext[x=0.136634in,y=0.945000in,,bottom,rotate=90.000000]{\color{textcolor}\fontsize{9.600000}{11.520000}\selectfont Execution Time}%
\end{pgfscope}%
\begin{pgfscope}%
\pgfpathrectangle{\pgfqpoint{0.681500in}{0.420000in}}{\pgfqpoint{1.621500in}{1.050000in}} %
\pgfusepath{clip}%
\pgfsetbuttcap%
\pgfsetroundjoin%
\definecolor{currentfill}{rgb}{0.737255,0.741176,0.133333}%
\pgfsetfillcolor{currentfill}%
\pgfsetlinewidth{1.138252pt}%
\definecolor{currentstroke}{rgb}{0.737255,0.741176,0.133333}%
\pgfsetstrokecolor{currentstroke}%
\pgfsetdash{}{0pt}%
\pgfpathmoveto{\pgfqpoint{0.816625in}{0.459364in}}%
\pgfpathcurveto{\pgfqpoint{0.823605in}{0.459364in}}{\pgfqpoint{0.830300in}{0.462137in}}{\pgfqpoint{0.835236in}{0.467073in}}%
\pgfpathcurveto{\pgfqpoint{0.840171in}{0.472008in}}{\pgfqpoint{0.842945in}{0.478703in}}{\pgfqpoint{0.842945in}{0.485683in}}%
\pgfpathcurveto{\pgfqpoint{0.842945in}{0.492663in}}{\pgfqpoint{0.840171in}{0.499359in}}{\pgfqpoint{0.835236in}{0.504294in}}%
\pgfpathcurveto{\pgfqpoint{0.830300in}{0.509230in}}{\pgfqpoint{0.823605in}{0.512003in}}{\pgfqpoint{0.816625in}{0.512003in}}%
\pgfpathcurveto{\pgfqpoint{0.809645in}{0.512003in}}{\pgfqpoint{0.802950in}{0.509230in}}{\pgfqpoint{0.798014in}{0.504294in}}%
\pgfpathcurveto{\pgfqpoint{0.793079in}{0.499359in}}{\pgfqpoint{0.790305in}{0.492663in}}{\pgfqpoint{0.790305in}{0.485683in}}%
\pgfpathcurveto{\pgfqpoint{0.790305in}{0.478703in}}{\pgfqpoint{0.793079in}{0.472008in}}{\pgfqpoint{0.798014in}{0.467073in}}%
\pgfpathcurveto{\pgfqpoint{0.802950in}{0.462137in}}{\pgfqpoint{0.809645in}{0.459364in}}{\pgfqpoint{0.816625in}{0.459364in}}%
\pgfpathclose%
\pgfusepath{stroke,fill}%
\end{pgfscope}%
\begin{pgfscope}%
\pgfpathrectangle{\pgfqpoint{0.681500in}{0.420000in}}{\pgfqpoint{1.621500in}{1.050000in}} %
\pgfusepath{clip}%
\pgfsetbuttcap%
\pgfsetroundjoin%
\definecolor{currentfill}{rgb}{0.737255,0.741176,0.133333}%
\pgfsetfillcolor{currentfill}%
\pgfsetlinewidth{1.138252pt}%
\definecolor{currentstroke}{rgb}{0.737255,0.741176,0.133333}%
\pgfsetstrokecolor{currentstroke}%
\pgfsetdash{}{0pt}%
\pgfpathmoveto{\pgfqpoint{1.086875in}{0.508440in}}%
\pgfpathcurveto{\pgfqpoint{1.093855in}{0.508440in}}{\pgfqpoint{1.100550in}{0.511213in}}{\pgfqpoint{1.105486in}{0.516149in}}%
\pgfpathcurveto{\pgfqpoint{1.110421in}{0.521085in}}{\pgfqpoint{1.113195in}{0.527780in}}{\pgfqpoint{1.113195in}{0.534760in}}%
\pgfpathcurveto{\pgfqpoint{1.113195in}{0.541740in}}{\pgfqpoint{1.110421in}{0.548435in}}{\pgfqpoint{1.105486in}{0.553371in}}%
\pgfpathcurveto{\pgfqpoint{1.100550in}{0.558306in}}{\pgfqpoint{1.093855in}{0.561079in}}{\pgfqpoint{1.086875in}{0.561079in}}%
\pgfpathcurveto{\pgfqpoint{1.079895in}{0.561079in}}{\pgfqpoint{1.073200in}{0.558306in}}{\pgfqpoint{1.068264in}{0.553371in}}%
\pgfpathcurveto{\pgfqpoint{1.063329in}{0.548435in}}{\pgfqpoint{1.060555in}{0.541740in}}{\pgfqpoint{1.060555in}{0.534760in}}%
\pgfpathcurveto{\pgfqpoint{1.060555in}{0.527780in}}{\pgfqpoint{1.063329in}{0.521085in}}{\pgfqpoint{1.068264in}{0.516149in}}%
\pgfpathcurveto{\pgfqpoint{1.073200in}{0.511213in}}{\pgfqpoint{1.079895in}{0.508440in}}{\pgfqpoint{1.086875in}{0.508440in}}%
\pgfpathclose%
\pgfusepath{stroke,fill}%
\end{pgfscope}%
\begin{pgfscope}%
\pgfpathrectangle{\pgfqpoint{0.681500in}{0.420000in}}{\pgfqpoint{1.621500in}{1.050000in}} %
\pgfusepath{clip}%
\pgfsetbuttcap%
\pgfsetroundjoin%
\definecolor{currentfill}{rgb}{0.737255,0.741176,0.133333}%
\pgfsetfillcolor{currentfill}%
\pgfsetlinewidth{1.138252pt}%
\definecolor{currentstroke}{rgb}{0.737255,0.741176,0.133333}%
\pgfsetstrokecolor{currentstroke}%
\pgfsetdash{}{0pt}%
\pgfpathmoveto{\pgfqpoint{1.357125in}{0.584703in}}%
\pgfpathcurveto{\pgfqpoint{1.364105in}{0.584703in}}{\pgfqpoint{1.370800in}{0.587476in}}{\pgfqpoint{1.375736in}{0.592412in}}%
\pgfpathcurveto{\pgfqpoint{1.380671in}{0.597348in}}{\pgfqpoint{1.383445in}{0.604043in}}{\pgfqpoint{1.383445in}{0.611023in}}%
\pgfpathcurveto{\pgfqpoint{1.383445in}{0.618003in}}{\pgfqpoint{1.380671in}{0.624698in}}{\pgfqpoint{1.375736in}{0.629634in}}%
\pgfpathcurveto{\pgfqpoint{1.370800in}{0.634569in}}{\pgfqpoint{1.364105in}{0.637342in}}{\pgfqpoint{1.357125in}{0.637342in}}%
\pgfpathcurveto{\pgfqpoint{1.350145in}{0.637342in}}{\pgfqpoint{1.343450in}{0.634569in}}{\pgfqpoint{1.338514in}{0.629634in}}%
\pgfpathcurveto{\pgfqpoint{1.333579in}{0.624698in}}{\pgfqpoint{1.330805in}{0.618003in}}{\pgfqpoint{1.330805in}{0.611023in}}%
\pgfpathcurveto{\pgfqpoint{1.330805in}{0.604043in}}{\pgfqpoint{1.333579in}{0.597348in}}{\pgfqpoint{1.338514in}{0.592412in}}%
\pgfpathcurveto{\pgfqpoint{1.343450in}{0.587476in}}{\pgfqpoint{1.350145in}{0.584703in}}{\pgfqpoint{1.357125in}{0.584703in}}%
\pgfpathclose%
\pgfusepath{stroke,fill}%
\end{pgfscope}%
\begin{pgfscope}%
\pgfpathrectangle{\pgfqpoint{0.681500in}{0.420000in}}{\pgfqpoint{1.621500in}{1.050000in}} %
\pgfusepath{clip}%
\pgfsetbuttcap%
\pgfsetroundjoin%
\definecolor{currentfill}{rgb}{0.737255,0.741176,0.133333}%
\pgfsetfillcolor{currentfill}%
\pgfsetlinewidth{1.138252pt}%
\definecolor{currentstroke}{rgb}{0.737255,0.741176,0.133333}%
\pgfsetstrokecolor{currentstroke}%
\pgfsetdash{}{0pt}%
\pgfpathmoveto{\pgfqpoint{1.627375in}{0.687638in}}%
\pgfpathcurveto{\pgfqpoint{1.634355in}{0.687638in}}{\pgfqpoint{1.641050in}{0.690412in}}{\pgfqpoint{1.645986in}{0.695347in}}%
\pgfpathcurveto{\pgfqpoint{1.650921in}{0.700283in}}{\pgfqpoint{1.653695in}{0.706978in}}{\pgfqpoint{1.653695in}{0.713958in}}%
\pgfpathcurveto{\pgfqpoint{1.653695in}{0.720938in}}{\pgfqpoint{1.650921in}{0.727633in}}{\pgfqpoint{1.645986in}{0.732569in}}%
\pgfpathcurveto{\pgfqpoint{1.641050in}{0.737504in}}{\pgfqpoint{1.634355in}{0.740278in}}{\pgfqpoint{1.627375in}{0.740278in}}%
\pgfpathcurveto{\pgfqpoint{1.620395in}{0.740278in}}{\pgfqpoint{1.613700in}{0.737504in}}{\pgfqpoint{1.608764in}{0.732569in}}%
\pgfpathcurveto{\pgfqpoint{1.603829in}{0.727633in}}{\pgfqpoint{1.601055in}{0.720938in}}{\pgfqpoint{1.601055in}{0.713958in}}%
\pgfpathcurveto{\pgfqpoint{1.601055in}{0.706978in}}{\pgfqpoint{1.603829in}{0.700283in}}{\pgfqpoint{1.608764in}{0.695347in}}%
\pgfpathcurveto{\pgfqpoint{1.613700in}{0.690412in}}{\pgfqpoint{1.620395in}{0.687638in}}{\pgfqpoint{1.627375in}{0.687638in}}%
\pgfpathclose%
\pgfusepath{stroke,fill}%
\end{pgfscope}%
\begin{pgfscope}%
\pgfpathrectangle{\pgfqpoint{0.681500in}{0.420000in}}{\pgfqpoint{1.621500in}{1.050000in}} %
\pgfusepath{clip}%
\pgfsetbuttcap%
\pgfsetroundjoin%
\definecolor{currentfill}{rgb}{0.737255,0.741176,0.133333}%
\pgfsetfillcolor{currentfill}%
\pgfsetlinewidth{1.138252pt}%
\definecolor{currentstroke}{rgb}{0.737255,0.741176,0.133333}%
\pgfsetstrokecolor{currentstroke}%
\pgfsetdash{}{0pt}%
\pgfpathmoveto{\pgfqpoint{1.897625in}{0.789254in}}%
\pgfpathcurveto{\pgfqpoint{1.904605in}{0.789254in}}{\pgfqpoint{1.911300in}{0.792027in}}{\pgfqpoint{1.916236in}{0.796963in}}%
\pgfpathcurveto{\pgfqpoint{1.921171in}{0.801899in}}{\pgfqpoint{1.923945in}{0.808594in}}{\pgfqpoint{1.923945in}{0.815574in}}%
\pgfpathcurveto{\pgfqpoint{1.923945in}{0.822554in}}{\pgfqpoint{1.921171in}{0.829249in}}{\pgfqpoint{1.916236in}{0.834185in}}%
\pgfpathcurveto{\pgfqpoint{1.911300in}{0.839120in}}{\pgfqpoint{1.904605in}{0.841893in}}{\pgfqpoint{1.897625in}{0.841893in}}%
\pgfpathcurveto{\pgfqpoint{1.890645in}{0.841893in}}{\pgfqpoint{1.883950in}{0.839120in}}{\pgfqpoint{1.879014in}{0.834185in}}%
\pgfpathcurveto{\pgfqpoint{1.874079in}{0.829249in}}{\pgfqpoint{1.871305in}{0.822554in}}{\pgfqpoint{1.871305in}{0.815574in}}%
\pgfpathcurveto{\pgfqpoint{1.871305in}{0.808594in}}{\pgfqpoint{1.874079in}{0.801899in}}{\pgfqpoint{1.879014in}{0.796963in}}%
\pgfpathcurveto{\pgfqpoint{1.883950in}{0.792027in}}{\pgfqpoint{1.890645in}{0.789254in}}{\pgfqpoint{1.897625in}{0.789254in}}%
\pgfpathclose%
\pgfusepath{stroke,fill}%
\end{pgfscope}%
\begin{pgfscope}%
\pgfpathrectangle{\pgfqpoint{0.681500in}{0.420000in}}{\pgfqpoint{1.621500in}{1.050000in}} %
\pgfusepath{clip}%
\pgfsetbuttcap%
\pgfsetroundjoin%
\definecolor{currentfill}{rgb}{0.737255,0.741176,0.133333}%
\pgfsetfillcolor{currentfill}%
\pgfsetlinewidth{1.138252pt}%
\definecolor{currentstroke}{rgb}{0.737255,0.741176,0.133333}%
\pgfsetstrokecolor{currentstroke}%
\pgfsetdash{}{0pt}%
\pgfpathmoveto{\pgfqpoint{2.167875in}{0.866532in}}%
\pgfpathcurveto{\pgfqpoint{2.174855in}{0.866532in}}{\pgfqpoint{2.181550in}{0.869305in}}{\pgfqpoint{2.186486in}{0.874241in}}%
\pgfpathcurveto{\pgfqpoint{2.191421in}{0.879176in}}{\pgfqpoint{2.194195in}{0.885871in}}{\pgfqpoint{2.194195in}{0.892852in}}%
\pgfpathcurveto{\pgfqpoint{2.194195in}{0.899832in}}{\pgfqpoint{2.191421in}{0.906527in}}{\pgfqpoint{2.186486in}{0.911462in}}%
\pgfpathcurveto{\pgfqpoint{2.181550in}{0.916398in}}{\pgfqpoint{2.174855in}{0.919171in}}{\pgfqpoint{2.167875in}{0.919171in}}%
\pgfpathcurveto{\pgfqpoint{2.160895in}{0.919171in}}{\pgfqpoint{2.154200in}{0.916398in}}{\pgfqpoint{2.149264in}{0.911462in}}%
\pgfpathcurveto{\pgfqpoint{2.144329in}{0.906527in}}{\pgfqpoint{2.141555in}{0.899832in}}{\pgfqpoint{2.141555in}{0.892852in}}%
\pgfpathcurveto{\pgfqpoint{2.141555in}{0.885871in}}{\pgfqpoint{2.144329in}{0.879176in}}{\pgfqpoint{2.149264in}{0.874241in}}%
\pgfpathcurveto{\pgfqpoint{2.154200in}{0.869305in}}{\pgfqpoint{2.160895in}{0.866532in}}{\pgfqpoint{2.167875in}{0.866532in}}%
\pgfpathclose%
\pgfusepath{stroke,fill}%
\end{pgfscope}%
\begin{pgfscope}%
\pgfpathrectangle{\pgfqpoint{0.681500in}{0.420000in}}{\pgfqpoint{1.621500in}{1.050000in}} %
\pgfusepath{clip}%
\pgfsetroundcap%
\pgfsetroundjoin%
\pgfsetlinewidth{1.517670pt}%
\definecolor{currentstroke}{rgb}{0.737255,0.741176,0.133333}%
\pgfsetstrokecolor{currentstroke}%
\pgfsetdash{}{0pt}%
\pgfpathmoveto{\pgfqpoint{0.816625in}{0.485683in}}%
\pgfpathlineto{\pgfqpoint{1.086875in}{0.534760in}}%
\pgfpathlineto{\pgfqpoint{1.357125in}{0.611023in}}%
\pgfpathlineto{\pgfqpoint{1.627375in}{0.713958in}}%
\pgfpathlineto{\pgfqpoint{1.897625in}{0.815574in}}%
\pgfpathlineto{\pgfqpoint{2.167875in}{0.892852in}}%
\pgfusepath{stroke}%
\end{pgfscope}%
\begin{pgfscope}%
\pgfpathrectangle{\pgfqpoint{0.681500in}{0.420000in}}{\pgfqpoint{1.621500in}{1.050000in}} %
\pgfusepath{clip}%
\pgfsetroundcap%
\pgfsetroundjoin%
\pgfsetlinewidth{2.168100pt}%
\definecolor{currentstroke}{rgb}{0.737255,0.741176,0.133333}%
\pgfsetstrokecolor{currentstroke}%
\pgfsetdash{}{0pt}%
\pgfpathmoveto{\pgfqpoint{0.816625in}{0.481041in}}%
\pgfpathlineto{\pgfqpoint{0.816625in}{0.489621in}}%
\pgfusepath{stroke}%
\end{pgfscope}%
\begin{pgfscope}%
\pgfpathrectangle{\pgfqpoint{0.681500in}{0.420000in}}{\pgfqpoint{1.621500in}{1.050000in}} %
\pgfusepath{clip}%
\pgfsetroundcap%
\pgfsetroundjoin%
\pgfsetlinewidth{2.168100pt}%
\definecolor{currentstroke}{rgb}{0.737255,0.741176,0.133333}%
\pgfsetstrokecolor{currentstroke}%
\pgfsetdash{}{0pt}%
\pgfpathmoveto{\pgfqpoint{1.086875in}{0.525461in}}%
\pgfpathlineto{\pgfqpoint{1.086875in}{0.541228in}}%
\pgfusepath{stroke}%
\end{pgfscope}%
\begin{pgfscope}%
\pgfpathrectangle{\pgfqpoint{0.681500in}{0.420000in}}{\pgfqpoint{1.621500in}{1.050000in}} %
\pgfusepath{clip}%
\pgfsetroundcap%
\pgfsetroundjoin%
\pgfsetlinewidth{2.168100pt}%
\definecolor{currentstroke}{rgb}{0.737255,0.741176,0.133333}%
\pgfsetstrokecolor{currentstroke}%
\pgfsetdash{}{0pt}%
\pgfpathmoveto{\pgfqpoint{1.357125in}{0.599435in}}%
\pgfpathlineto{\pgfqpoint{1.357125in}{0.620119in}}%
\pgfusepath{stroke}%
\end{pgfscope}%
\begin{pgfscope}%
\pgfpathrectangle{\pgfqpoint{0.681500in}{0.420000in}}{\pgfqpoint{1.621500in}{1.050000in}} %
\pgfusepath{clip}%
\pgfsetroundcap%
\pgfsetroundjoin%
\pgfsetlinewidth{2.168100pt}%
\definecolor{currentstroke}{rgb}{0.737255,0.741176,0.133333}%
\pgfsetstrokecolor{currentstroke}%
\pgfsetdash{}{0pt}%
\pgfpathmoveto{\pgfqpoint{1.627375in}{0.683945in}}%
\pgfpathlineto{\pgfqpoint{1.627375in}{0.731661in}}%
\pgfusepath{stroke}%
\end{pgfscope}%
\begin{pgfscope}%
\pgfpathrectangle{\pgfqpoint{0.681500in}{0.420000in}}{\pgfqpoint{1.621500in}{1.050000in}} %
\pgfusepath{clip}%
\pgfsetroundcap%
\pgfsetroundjoin%
\pgfsetlinewidth{2.168100pt}%
\definecolor{currentstroke}{rgb}{0.737255,0.741176,0.133333}%
\pgfsetstrokecolor{currentstroke}%
\pgfsetdash{}{0pt}%
\pgfpathmoveto{\pgfqpoint{1.897625in}{0.779721in}}%
\pgfpathlineto{\pgfqpoint{1.897625in}{0.824162in}}%
\pgfusepath{stroke}%
\end{pgfscope}%
\begin{pgfscope}%
\pgfpathrectangle{\pgfqpoint{0.681500in}{0.420000in}}{\pgfqpoint{1.621500in}{1.050000in}} %
\pgfusepath{clip}%
\pgfsetroundcap%
\pgfsetroundjoin%
\pgfsetlinewidth{2.168100pt}%
\definecolor{currentstroke}{rgb}{0.737255,0.741176,0.133333}%
\pgfsetstrokecolor{currentstroke}%
\pgfsetdash{}{0pt}%
\pgfpathmoveto{\pgfqpoint{2.167875in}{0.878804in}}%
\pgfpathlineto{\pgfqpoint{2.167875in}{0.897711in}}%
\pgfusepath{stroke}%
\end{pgfscope}%
\begin{pgfscope}%
\pgfpathrectangle{\pgfqpoint{0.681500in}{0.420000in}}{\pgfqpoint{1.621500in}{1.050000in}} %
\pgfusepath{clip}%
\pgfsetbuttcap%
\pgfsetroundjoin%
\definecolor{currentfill}{rgb}{0.090196,0.745098,0.811765}%
\pgfsetfillcolor{currentfill}%
\pgfsetlinewidth{1.138252pt}%
\definecolor{currentstroke}{rgb}{0.090196,0.745098,0.811765}%
\pgfsetstrokecolor{currentstroke}%
\pgfsetdash{}{0pt}%
\pgfpathmoveto{\pgfqpoint{0.816625in}{0.457622in}}%
\pgfpathcurveto{\pgfqpoint{0.823605in}{0.457622in}}{\pgfqpoint{0.830300in}{0.460395in}}{\pgfqpoint{0.835236in}{0.465331in}}%
\pgfpathcurveto{\pgfqpoint{0.840171in}{0.470267in}}{\pgfqpoint{0.842945in}{0.476962in}}{\pgfqpoint{0.842945in}{0.483942in}}%
\pgfpathcurveto{\pgfqpoint{0.842945in}{0.490922in}}{\pgfqpoint{0.840171in}{0.497617in}}{\pgfqpoint{0.835236in}{0.502552in}}%
\pgfpathcurveto{\pgfqpoint{0.830300in}{0.507488in}}{\pgfqpoint{0.823605in}{0.510261in}}{\pgfqpoint{0.816625in}{0.510261in}}%
\pgfpathcurveto{\pgfqpoint{0.809645in}{0.510261in}}{\pgfqpoint{0.802950in}{0.507488in}}{\pgfqpoint{0.798014in}{0.502552in}}%
\pgfpathcurveto{\pgfqpoint{0.793079in}{0.497617in}}{\pgfqpoint{0.790305in}{0.490922in}}{\pgfqpoint{0.790305in}{0.483942in}}%
\pgfpathcurveto{\pgfqpoint{0.790305in}{0.476962in}}{\pgfqpoint{0.793079in}{0.470267in}}{\pgfqpoint{0.798014in}{0.465331in}}%
\pgfpathcurveto{\pgfqpoint{0.802950in}{0.460395in}}{\pgfqpoint{0.809645in}{0.457622in}}{\pgfqpoint{0.816625in}{0.457622in}}%
\pgfpathclose%
\pgfusepath{stroke,fill}%
\end{pgfscope}%
\begin{pgfscope}%
\pgfpathrectangle{\pgfqpoint{0.681500in}{0.420000in}}{\pgfqpoint{1.621500in}{1.050000in}} %
\pgfusepath{clip}%
\pgfsetbuttcap%
\pgfsetroundjoin%
\definecolor{currentfill}{rgb}{0.090196,0.745098,0.811765}%
\pgfsetfillcolor{currentfill}%
\pgfsetlinewidth{1.138252pt}%
\definecolor{currentstroke}{rgb}{0.090196,0.745098,0.811765}%
\pgfsetstrokecolor{currentstroke}%
\pgfsetdash{}{0pt}%
\pgfpathmoveto{\pgfqpoint{1.086875in}{0.509255in}}%
\pgfpathcurveto{\pgfqpoint{1.093855in}{0.509255in}}{\pgfqpoint{1.100550in}{0.512028in}}{\pgfqpoint{1.105486in}{0.516964in}}%
\pgfpathcurveto{\pgfqpoint{1.110421in}{0.521899in}}{\pgfqpoint{1.113195in}{0.528594in}}{\pgfqpoint{1.113195in}{0.535574in}}%
\pgfpathcurveto{\pgfqpoint{1.113195in}{0.542554in}}{\pgfqpoint{1.110421in}{0.549249in}}{\pgfqpoint{1.105486in}{0.554185in}}%
\pgfpathcurveto{\pgfqpoint{1.100550in}{0.559121in}}{\pgfqpoint{1.093855in}{0.561894in}}{\pgfqpoint{1.086875in}{0.561894in}}%
\pgfpathcurveto{\pgfqpoint{1.079895in}{0.561894in}}{\pgfqpoint{1.073200in}{0.559121in}}{\pgfqpoint{1.068264in}{0.554185in}}%
\pgfpathcurveto{\pgfqpoint{1.063329in}{0.549249in}}{\pgfqpoint{1.060555in}{0.542554in}}{\pgfqpoint{1.060555in}{0.535574in}}%
\pgfpathcurveto{\pgfqpoint{1.060555in}{0.528594in}}{\pgfqpoint{1.063329in}{0.521899in}}{\pgfqpoint{1.068264in}{0.516964in}}%
\pgfpathcurveto{\pgfqpoint{1.073200in}{0.512028in}}{\pgfqpoint{1.079895in}{0.509255in}}{\pgfqpoint{1.086875in}{0.509255in}}%
\pgfpathclose%
\pgfusepath{stroke,fill}%
\end{pgfscope}%
\begin{pgfscope}%
\pgfpathrectangle{\pgfqpoint{0.681500in}{0.420000in}}{\pgfqpoint{1.621500in}{1.050000in}} %
\pgfusepath{clip}%
\pgfsetbuttcap%
\pgfsetroundjoin%
\definecolor{currentfill}{rgb}{0.090196,0.745098,0.811765}%
\pgfsetfillcolor{currentfill}%
\pgfsetlinewidth{1.138252pt}%
\definecolor{currentstroke}{rgb}{0.090196,0.745098,0.811765}%
\pgfsetstrokecolor{currentstroke}%
\pgfsetdash{}{0pt}%
\pgfpathmoveto{\pgfqpoint{1.357125in}{0.589019in}}%
\pgfpathcurveto{\pgfqpoint{1.364105in}{0.589019in}}{\pgfqpoint{1.370800in}{0.591792in}}{\pgfqpoint{1.375736in}{0.596727in}}%
\pgfpathcurveto{\pgfqpoint{1.380671in}{0.601663in}}{\pgfqpoint{1.383445in}{0.608358in}}{\pgfqpoint{1.383445in}{0.615338in}}%
\pgfpathcurveto{\pgfqpoint{1.383445in}{0.622318in}}{\pgfqpoint{1.380671in}{0.629013in}}{\pgfqpoint{1.375736in}{0.633949in}}%
\pgfpathcurveto{\pgfqpoint{1.370800in}{0.638885in}}{\pgfqpoint{1.364105in}{0.641658in}}{\pgfqpoint{1.357125in}{0.641658in}}%
\pgfpathcurveto{\pgfqpoint{1.350145in}{0.641658in}}{\pgfqpoint{1.343450in}{0.638885in}}{\pgfqpoint{1.338514in}{0.633949in}}%
\pgfpathcurveto{\pgfqpoint{1.333579in}{0.629013in}}{\pgfqpoint{1.330805in}{0.622318in}}{\pgfqpoint{1.330805in}{0.615338in}}%
\pgfpathcurveto{\pgfqpoint{1.330805in}{0.608358in}}{\pgfqpoint{1.333579in}{0.601663in}}{\pgfqpoint{1.338514in}{0.596727in}}%
\pgfpathcurveto{\pgfqpoint{1.343450in}{0.591792in}}{\pgfqpoint{1.350145in}{0.589019in}}{\pgfqpoint{1.357125in}{0.589019in}}%
\pgfpathclose%
\pgfusepath{stroke,fill}%
\end{pgfscope}%
\begin{pgfscope}%
\pgfpathrectangle{\pgfqpoint{0.681500in}{0.420000in}}{\pgfqpoint{1.621500in}{1.050000in}} %
\pgfusepath{clip}%
\pgfsetbuttcap%
\pgfsetroundjoin%
\definecolor{currentfill}{rgb}{0.090196,0.745098,0.811765}%
\pgfsetfillcolor{currentfill}%
\pgfsetlinewidth{1.138252pt}%
\definecolor{currentstroke}{rgb}{0.090196,0.745098,0.811765}%
\pgfsetstrokecolor{currentstroke}%
\pgfsetdash{}{0pt}%
\pgfpathmoveto{\pgfqpoint{1.627375in}{0.682117in}}%
\pgfpathcurveto{\pgfqpoint{1.634355in}{0.682117in}}{\pgfqpoint{1.641050in}{0.684890in}}{\pgfqpoint{1.645986in}{0.689826in}}%
\pgfpathcurveto{\pgfqpoint{1.650921in}{0.694761in}}{\pgfqpoint{1.653695in}{0.701456in}}{\pgfqpoint{1.653695in}{0.708436in}}%
\pgfpathcurveto{\pgfqpoint{1.653695in}{0.715416in}}{\pgfqpoint{1.650921in}{0.722112in}}{\pgfqpoint{1.645986in}{0.727047in}}%
\pgfpathcurveto{\pgfqpoint{1.641050in}{0.731983in}}{\pgfqpoint{1.634355in}{0.734756in}}{\pgfqpoint{1.627375in}{0.734756in}}%
\pgfpathcurveto{\pgfqpoint{1.620395in}{0.734756in}}{\pgfqpoint{1.613700in}{0.731983in}}{\pgfqpoint{1.608764in}{0.727047in}}%
\pgfpathcurveto{\pgfqpoint{1.603829in}{0.722112in}}{\pgfqpoint{1.601055in}{0.715416in}}{\pgfqpoint{1.601055in}{0.708436in}}%
\pgfpathcurveto{\pgfqpoint{1.601055in}{0.701456in}}{\pgfqpoint{1.603829in}{0.694761in}}{\pgfqpoint{1.608764in}{0.689826in}}%
\pgfpathcurveto{\pgfqpoint{1.613700in}{0.684890in}}{\pgfqpoint{1.620395in}{0.682117in}}{\pgfqpoint{1.627375in}{0.682117in}}%
\pgfpathclose%
\pgfusepath{stroke,fill}%
\end{pgfscope}%
\begin{pgfscope}%
\pgfpathrectangle{\pgfqpoint{0.681500in}{0.420000in}}{\pgfqpoint{1.621500in}{1.050000in}} %
\pgfusepath{clip}%
\pgfsetbuttcap%
\pgfsetroundjoin%
\definecolor{currentfill}{rgb}{0.090196,0.745098,0.811765}%
\pgfsetfillcolor{currentfill}%
\pgfsetlinewidth{1.138252pt}%
\definecolor{currentstroke}{rgb}{0.090196,0.745098,0.811765}%
\pgfsetstrokecolor{currentstroke}%
\pgfsetdash{}{0pt}%
\pgfpathmoveto{\pgfqpoint{1.897625in}{0.780605in}}%
\pgfpathcurveto{\pgfqpoint{1.904605in}{0.780605in}}{\pgfqpoint{1.911300in}{0.783378in}}{\pgfqpoint{1.916236in}{0.788314in}}%
\pgfpathcurveto{\pgfqpoint{1.921171in}{0.793250in}}{\pgfqpoint{1.923945in}{0.799945in}}{\pgfqpoint{1.923945in}{0.806925in}}%
\pgfpathcurveto{\pgfqpoint{1.923945in}{0.813905in}}{\pgfqpoint{1.921171in}{0.820600in}}{\pgfqpoint{1.916236in}{0.825536in}}%
\pgfpathcurveto{\pgfqpoint{1.911300in}{0.830471in}}{\pgfqpoint{1.904605in}{0.833244in}}{\pgfqpoint{1.897625in}{0.833244in}}%
\pgfpathcurveto{\pgfqpoint{1.890645in}{0.833244in}}{\pgfqpoint{1.883950in}{0.830471in}}{\pgfqpoint{1.879014in}{0.825536in}}%
\pgfpathcurveto{\pgfqpoint{1.874079in}{0.820600in}}{\pgfqpoint{1.871305in}{0.813905in}}{\pgfqpoint{1.871305in}{0.806925in}}%
\pgfpathcurveto{\pgfqpoint{1.871305in}{0.799945in}}{\pgfqpoint{1.874079in}{0.793250in}}{\pgfqpoint{1.879014in}{0.788314in}}%
\pgfpathcurveto{\pgfqpoint{1.883950in}{0.783378in}}{\pgfqpoint{1.890645in}{0.780605in}}{\pgfqpoint{1.897625in}{0.780605in}}%
\pgfpathclose%
\pgfusepath{stroke,fill}%
\end{pgfscope}%
\begin{pgfscope}%
\pgfpathrectangle{\pgfqpoint{0.681500in}{0.420000in}}{\pgfqpoint{1.621500in}{1.050000in}} %
\pgfusepath{clip}%
\pgfsetbuttcap%
\pgfsetroundjoin%
\definecolor{currentfill}{rgb}{0.090196,0.745098,0.811765}%
\pgfsetfillcolor{currentfill}%
\pgfsetlinewidth{1.138252pt}%
\definecolor{currentstroke}{rgb}{0.090196,0.745098,0.811765}%
\pgfsetstrokecolor{currentstroke}%
\pgfsetdash{}{0pt}%
\pgfpathmoveto{\pgfqpoint{2.167875in}{0.863329in}}%
\pgfpathcurveto{\pgfqpoint{2.174855in}{0.863329in}}{\pgfqpoint{2.181550in}{0.866103in}}{\pgfqpoint{2.186486in}{0.871038in}}%
\pgfpathcurveto{\pgfqpoint{2.191421in}{0.875974in}}{\pgfqpoint{2.194195in}{0.882669in}}{\pgfqpoint{2.194195in}{0.889649in}}%
\pgfpathcurveto{\pgfqpoint{2.194195in}{0.896629in}}{\pgfqpoint{2.191421in}{0.903324in}}{\pgfqpoint{2.186486in}{0.908260in}}%
\pgfpathcurveto{\pgfqpoint{2.181550in}{0.913195in}}{\pgfqpoint{2.174855in}{0.915969in}}{\pgfqpoint{2.167875in}{0.915969in}}%
\pgfpathcurveto{\pgfqpoint{2.160895in}{0.915969in}}{\pgfqpoint{2.154200in}{0.913195in}}{\pgfqpoint{2.149264in}{0.908260in}}%
\pgfpathcurveto{\pgfqpoint{2.144329in}{0.903324in}}{\pgfqpoint{2.141555in}{0.896629in}}{\pgfqpoint{2.141555in}{0.889649in}}%
\pgfpathcurveto{\pgfqpoint{2.141555in}{0.882669in}}{\pgfqpoint{2.144329in}{0.875974in}}{\pgfqpoint{2.149264in}{0.871038in}}%
\pgfpathcurveto{\pgfqpoint{2.154200in}{0.866103in}}{\pgfqpoint{2.160895in}{0.863329in}}{\pgfqpoint{2.167875in}{0.863329in}}%
\pgfpathclose%
\pgfusepath{stroke,fill}%
\end{pgfscope}%
\begin{pgfscope}%
\pgfpathrectangle{\pgfqpoint{0.681500in}{0.420000in}}{\pgfqpoint{1.621500in}{1.050000in}} %
\pgfusepath{clip}%
\pgfsetroundcap%
\pgfsetroundjoin%
\pgfsetlinewidth{1.517670pt}%
\definecolor{currentstroke}{rgb}{0.090196,0.745098,0.811765}%
\pgfsetstrokecolor{currentstroke}%
\pgfsetdash{}{0pt}%
\pgfpathmoveto{\pgfqpoint{0.816625in}{0.483942in}}%
\pgfpathlineto{\pgfqpoint{1.086875in}{0.535574in}}%
\pgfpathlineto{\pgfqpoint{1.357125in}{0.615338in}}%
\pgfpathlineto{\pgfqpoint{1.627375in}{0.708436in}}%
\pgfpathlineto{\pgfqpoint{1.897625in}{0.806925in}}%
\pgfpathlineto{\pgfqpoint{2.167875in}{0.889649in}}%
\pgfusepath{stroke}%
\end{pgfscope}%
\begin{pgfscope}%
\pgfpathrectangle{\pgfqpoint{0.681500in}{0.420000in}}{\pgfqpoint{1.621500in}{1.050000in}} %
\pgfusepath{clip}%
\pgfsetroundcap%
\pgfsetroundjoin%
\pgfsetlinewidth{2.168100pt}%
\definecolor{currentstroke}{rgb}{0.090196,0.745098,0.811765}%
\pgfsetstrokecolor{currentstroke}%
\pgfsetdash{}{0pt}%
\pgfpathmoveto{\pgfqpoint{0.816625in}{0.481250in}}%
\pgfpathlineto{\pgfqpoint{0.816625in}{0.487244in}}%
\pgfusepath{stroke}%
\end{pgfscope}%
\begin{pgfscope}%
\pgfpathrectangle{\pgfqpoint{0.681500in}{0.420000in}}{\pgfqpoint{1.621500in}{1.050000in}} %
\pgfusepath{clip}%
\pgfsetroundcap%
\pgfsetroundjoin%
\pgfsetlinewidth{2.168100pt}%
\definecolor{currentstroke}{rgb}{0.090196,0.745098,0.811765}%
\pgfsetstrokecolor{currentstroke}%
\pgfsetdash{}{0pt}%
\pgfpathmoveto{\pgfqpoint{1.086875in}{0.531720in}}%
\pgfpathlineto{\pgfqpoint{1.086875in}{0.540633in}}%
\pgfusepath{stroke}%
\end{pgfscope}%
\begin{pgfscope}%
\pgfpathrectangle{\pgfqpoint{0.681500in}{0.420000in}}{\pgfqpoint{1.621500in}{1.050000in}} %
\pgfusepath{clip}%
\pgfsetroundcap%
\pgfsetroundjoin%
\pgfsetlinewidth{2.168100pt}%
\definecolor{currentstroke}{rgb}{0.090196,0.745098,0.811765}%
\pgfsetstrokecolor{currentstroke}%
\pgfsetdash{}{0pt}%
\pgfpathmoveto{\pgfqpoint{1.357125in}{0.603834in}}%
\pgfpathlineto{\pgfqpoint{1.357125in}{0.625230in}}%
\pgfusepath{stroke}%
\end{pgfscope}%
\begin{pgfscope}%
\pgfpathrectangle{\pgfqpoint{0.681500in}{0.420000in}}{\pgfqpoint{1.621500in}{1.050000in}} %
\pgfusepath{clip}%
\pgfsetroundcap%
\pgfsetroundjoin%
\pgfsetlinewidth{2.168100pt}%
\definecolor{currentstroke}{rgb}{0.090196,0.745098,0.811765}%
\pgfsetstrokecolor{currentstroke}%
\pgfsetdash{}{0pt}%
\pgfpathmoveto{\pgfqpoint{1.627375in}{0.701663in}}%
\pgfpathlineto{\pgfqpoint{1.627375in}{0.724961in}}%
\pgfusepath{stroke}%
\end{pgfscope}%
\begin{pgfscope}%
\pgfpathrectangle{\pgfqpoint{0.681500in}{0.420000in}}{\pgfqpoint{1.621500in}{1.050000in}} %
\pgfusepath{clip}%
\pgfsetroundcap%
\pgfsetroundjoin%
\pgfsetlinewidth{2.168100pt}%
\definecolor{currentstroke}{rgb}{0.090196,0.745098,0.811765}%
\pgfsetstrokecolor{currentstroke}%
\pgfsetdash{}{0pt}%
\pgfpathmoveto{\pgfqpoint{1.897625in}{0.781326in}}%
\pgfpathlineto{\pgfqpoint{1.897625in}{0.828595in}}%
\pgfusepath{stroke}%
\end{pgfscope}%
\begin{pgfscope}%
\pgfpathrectangle{\pgfqpoint{0.681500in}{0.420000in}}{\pgfqpoint{1.621500in}{1.050000in}} %
\pgfusepath{clip}%
\pgfsetroundcap%
\pgfsetroundjoin%
\pgfsetlinewidth{2.168100pt}%
\definecolor{currentstroke}{rgb}{0.090196,0.745098,0.811765}%
\pgfsetstrokecolor{currentstroke}%
\pgfsetdash{}{0pt}%
\pgfpathmoveto{\pgfqpoint{2.167875in}{0.854588in}}%
\pgfpathlineto{\pgfqpoint{2.167875in}{0.908565in}}%
\pgfusepath{stroke}%
\end{pgfscope}%
\begin{pgfscope}%
\pgfsetrectcap%
\pgfsetmiterjoin%
\pgfsetlinewidth{1.003750pt}%
\definecolor{currentstroke}{rgb}{0.800000,0.800000,0.800000}%
\pgfsetstrokecolor{currentstroke}%
\pgfsetdash{}{0pt}%
\pgfpathmoveto{\pgfqpoint{0.681500in}{0.420000in}}%
\pgfpathlineto{\pgfqpoint{0.681500in}{1.470000in}}%
\pgfusepath{stroke}%
\end{pgfscope}%
\begin{pgfscope}%
\pgfsetrectcap%
\pgfsetmiterjoin%
\pgfsetlinewidth{1.003750pt}%
\definecolor{currentstroke}{rgb}{0.800000,0.800000,0.800000}%
\pgfsetstrokecolor{currentstroke}%
\pgfsetdash{}{0pt}%
\pgfpathmoveto{\pgfqpoint{2.303000in}{0.420000in}}%
\pgfpathlineto{\pgfqpoint{2.303000in}{1.470000in}}%
\pgfusepath{stroke}%
\end{pgfscope}%
\begin{pgfscope}%
\pgfsetrectcap%
\pgfsetmiterjoin%
\pgfsetlinewidth{1.003750pt}%
\definecolor{currentstroke}{rgb}{0.800000,0.800000,0.800000}%
\pgfsetstrokecolor{currentstroke}%
\pgfsetdash{}{0pt}%
\pgfpathmoveto{\pgfqpoint{0.681500in}{0.420000in}}%
\pgfpathlineto{\pgfqpoint{2.303000in}{0.420000in}}%
\pgfusepath{stroke}%
\end{pgfscope}%
\begin{pgfscope}%
\pgfsetrectcap%
\pgfsetmiterjoin%
\pgfsetlinewidth{1.003750pt}%
\definecolor{currentstroke}{rgb}{0.800000,0.800000,0.800000}%
\pgfsetstrokecolor{currentstroke}%
\pgfsetdash{}{0pt}%
\pgfpathmoveto{\pgfqpoint{0.681500in}{1.470000in}}%
\pgfpathlineto{\pgfqpoint{2.303000in}{1.470000in}}%
\pgfusepath{stroke}%
\end{pgfscope}%
\begin{pgfscope}%
\pgfpathrectangle{\pgfqpoint{0.681500in}{0.420000in}}{\pgfqpoint{1.621500in}{1.050000in}} %
\pgfusepath{clip}%
\pgfsetbuttcap%
\pgfsetroundjoin%
\definecolor{currentfill}{rgb}{0.498039,0.498039,0.498039}%
\pgfsetfillcolor{currentfill}%
\pgfsetlinewidth{1.138252pt}%
\definecolor{currentstroke}{rgb}{0.498039,0.498039,0.498039}%
\pgfsetstrokecolor{currentstroke}%
\pgfsetdash{}{0pt}%
\pgfpathmoveto{\pgfqpoint{0.816625in}{0.448476in}}%
\pgfpathcurveto{\pgfqpoint{0.823605in}{0.448476in}}{\pgfqpoint{0.830300in}{0.451250in}}{\pgfqpoint{0.835236in}{0.456185in}}%
\pgfpathcurveto{\pgfqpoint{0.840171in}{0.461121in}}{\pgfqpoint{0.842945in}{0.467816in}}{\pgfqpoint{0.842945in}{0.474796in}}%
\pgfpathcurveto{\pgfqpoint{0.842945in}{0.481776in}}{\pgfqpoint{0.840171in}{0.488471in}}{\pgfqpoint{0.835236in}{0.493407in}}%
\pgfpathcurveto{\pgfqpoint{0.830300in}{0.498342in}}{\pgfqpoint{0.823605in}{0.501116in}}{\pgfqpoint{0.816625in}{0.501116in}}%
\pgfpathcurveto{\pgfqpoint{0.809645in}{0.501116in}}{\pgfqpoint{0.802950in}{0.498342in}}{\pgfqpoint{0.798014in}{0.493407in}}%
\pgfpathcurveto{\pgfqpoint{0.793079in}{0.488471in}}{\pgfqpoint{0.790305in}{0.481776in}}{\pgfqpoint{0.790305in}{0.474796in}}%
\pgfpathcurveto{\pgfqpoint{0.790305in}{0.467816in}}{\pgfqpoint{0.793079in}{0.461121in}}{\pgfqpoint{0.798014in}{0.456185in}}%
\pgfpathcurveto{\pgfqpoint{0.802950in}{0.451250in}}{\pgfqpoint{0.809645in}{0.448476in}}{\pgfqpoint{0.816625in}{0.448476in}}%
\pgfpathclose%
\pgfusepath{stroke,fill}%
\end{pgfscope}%
\begin{pgfscope}%
\pgfpathrectangle{\pgfqpoint{0.681500in}{0.420000in}}{\pgfqpoint{1.621500in}{1.050000in}} %
\pgfusepath{clip}%
\pgfsetbuttcap%
\pgfsetroundjoin%
\definecolor{currentfill}{rgb}{0.498039,0.498039,0.498039}%
\pgfsetfillcolor{currentfill}%
\pgfsetlinewidth{1.138252pt}%
\definecolor{currentstroke}{rgb}{0.498039,0.498039,0.498039}%
\pgfsetstrokecolor{currentstroke}%
\pgfsetdash{}{0pt}%
\pgfpathmoveto{\pgfqpoint{1.086875in}{0.493723in}}%
\pgfpathcurveto{\pgfqpoint{1.093855in}{0.493723in}}{\pgfqpoint{1.100550in}{0.496496in}}{\pgfqpoint{1.105486in}{0.501432in}}%
\pgfpathcurveto{\pgfqpoint{1.110421in}{0.506367in}}{\pgfqpoint{1.113195in}{0.513063in}}{\pgfqpoint{1.113195in}{0.520043in}}%
\pgfpathcurveto{\pgfqpoint{1.113195in}{0.527023in}}{\pgfqpoint{1.110421in}{0.533718in}}{\pgfqpoint{1.105486in}{0.538653in}}%
\pgfpathcurveto{\pgfqpoint{1.100550in}{0.543589in}}{\pgfqpoint{1.093855in}{0.546362in}}{\pgfqpoint{1.086875in}{0.546362in}}%
\pgfpathcurveto{\pgfqpoint{1.079895in}{0.546362in}}{\pgfqpoint{1.073200in}{0.543589in}}{\pgfqpoint{1.068264in}{0.538653in}}%
\pgfpathcurveto{\pgfqpoint{1.063329in}{0.533718in}}{\pgfqpoint{1.060555in}{0.527023in}}{\pgfqpoint{1.060555in}{0.520043in}}%
\pgfpathcurveto{\pgfqpoint{1.060555in}{0.513063in}}{\pgfqpoint{1.063329in}{0.506367in}}{\pgfqpoint{1.068264in}{0.501432in}}%
\pgfpathcurveto{\pgfqpoint{1.073200in}{0.496496in}}{\pgfqpoint{1.079895in}{0.493723in}}{\pgfqpoint{1.086875in}{0.493723in}}%
\pgfpathclose%
\pgfusepath{stroke,fill}%
\end{pgfscope}%
\begin{pgfscope}%
\pgfpathrectangle{\pgfqpoint{0.681500in}{0.420000in}}{\pgfqpoint{1.621500in}{1.050000in}} %
\pgfusepath{clip}%
\pgfsetbuttcap%
\pgfsetroundjoin%
\definecolor{currentfill}{rgb}{0.498039,0.498039,0.498039}%
\pgfsetfillcolor{currentfill}%
\pgfsetlinewidth{1.138252pt}%
\definecolor{currentstroke}{rgb}{0.498039,0.498039,0.498039}%
\pgfsetstrokecolor{currentstroke}%
\pgfsetdash{}{0pt}%
\pgfpathmoveto{\pgfqpoint{1.357125in}{0.546178in}}%
\pgfpathcurveto{\pgfqpoint{1.364105in}{0.546178in}}{\pgfqpoint{1.370800in}{0.548951in}}{\pgfqpoint{1.375736in}{0.553887in}}%
\pgfpathcurveto{\pgfqpoint{1.380671in}{0.558823in}}{\pgfqpoint{1.383445in}{0.565518in}}{\pgfqpoint{1.383445in}{0.572498in}}%
\pgfpathcurveto{\pgfqpoint{1.383445in}{0.579478in}}{\pgfqpoint{1.380671in}{0.586173in}}{\pgfqpoint{1.375736in}{0.591108in}}%
\pgfpathcurveto{\pgfqpoint{1.370800in}{0.596044in}}{\pgfqpoint{1.364105in}{0.598817in}}{\pgfqpoint{1.357125in}{0.598817in}}%
\pgfpathcurveto{\pgfqpoint{1.350145in}{0.598817in}}{\pgfqpoint{1.343450in}{0.596044in}}{\pgfqpoint{1.338514in}{0.591108in}}%
\pgfpathcurveto{\pgfqpoint{1.333579in}{0.586173in}}{\pgfqpoint{1.330805in}{0.579478in}}{\pgfqpoint{1.330805in}{0.572498in}}%
\pgfpathcurveto{\pgfqpoint{1.330805in}{0.565518in}}{\pgfqpoint{1.333579in}{0.558823in}}{\pgfqpoint{1.338514in}{0.553887in}}%
\pgfpathcurveto{\pgfqpoint{1.343450in}{0.548951in}}{\pgfqpoint{1.350145in}{0.546178in}}{\pgfqpoint{1.357125in}{0.546178in}}%
\pgfpathclose%
\pgfusepath{stroke,fill}%
\end{pgfscope}%
\begin{pgfscope}%
\pgfpathrectangle{\pgfqpoint{0.681500in}{0.420000in}}{\pgfqpoint{1.621500in}{1.050000in}} %
\pgfusepath{clip}%
\pgfsetbuttcap%
\pgfsetroundjoin%
\definecolor{currentfill}{rgb}{0.498039,0.498039,0.498039}%
\pgfsetfillcolor{currentfill}%
\pgfsetlinewidth{1.138252pt}%
\definecolor{currentstroke}{rgb}{0.498039,0.498039,0.498039}%
\pgfsetstrokecolor{currentstroke}%
\pgfsetdash{}{0pt}%
\pgfpathmoveto{\pgfqpoint{1.627375in}{0.585392in}}%
\pgfpathcurveto{\pgfqpoint{1.634355in}{0.585392in}}{\pgfqpoint{1.641050in}{0.588165in}}{\pgfqpoint{1.645986in}{0.593101in}}%
\pgfpathcurveto{\pgfqpoint{1.650921in}{0.598036in}}{\pgfqpoint{1.653695in}{0.604732in}}{\pgfqpoint{1.653695in}{0.611712in}}%
\pgfpathcurveto{\pgfqpoint{1.653695in}{0.618692in}}{\pgfqpoint{1.650921in}{0.625387in}}{\pgfqpoint{1.645986in}{0.630322in}}%
\pgfpathcurveto{\pgfqpoint{1.641050in}{0.635258in}}{\pgfqpoint{1.634355in}{0.638031in}}{\pgfqpoint{1.627375in}{0.638031in}}%
\pgfpathcurveto{\pgfqpoint{1.620395in}{0.638031in}}{\pgfqpoint{1.613700in}{0.635258in}}{\pgfqpoint{1.608764in}{0.630322in}}%
\pgfpathcurveto{\pgfqpoint{1.603829in}{0.625387in}}{\pgfqpoint{1.601055in}{0.618692in}}{\pgfqpoint{1.601055in}{0.611712in}}%
\pgfpathcurveto{\pgfqpoint{1.601055in}{0.604732in}}{\pgfqpoint{1.603829in}{0.598036in}}{\pgfqpoint{1.608764in}{0.593101in}}%
\pgfpathcurveto{\pgfqpoint{1.613700in}{0.588165in}}{\pgfqpoint{1.620395in}{0.585392in}}{\pgfqpoint{1.627375in}{0.585392in}}%
\pgfpathclose%
\pgfusepath{stroke,fill}%
\end{pgfscope}%
\begin{pgfscope}%
\pgfpathrectangle{\pgfqpoint{0.681500in}{0.420000in}}{\pgfqpoint{1.621500in}{1.050000in}} %
\pgfusepath{clip}%
\pgfsetbuttcap%
\pgfsetroundjoin%
\definecolor{currentfill}{rgb}{0.498039,0.498039,0.498039}%
\pgfsetfillcolor{currentfill}%
\pgfsetlinewidth{1.138252pt}%
\definecolor{currentstroke}{rgb}{0.498039,0.498039,0.498039}%
\pgfsetstrokecolor{currentstroke}%
\pgfsetdash{}{0pt}%
\pgfpathmoveto{\pgfqpoint{1.897625in}{0.650571in}}%
\pgfpathcurveto{\pgfqpoint{1.904605in}{0.650571in}}{\pgfqpoint{1.911300in}{0.653344in}}{\pgfqpoint{1.916236in}{0.658280in}}%
\pgfpathcurveto{\pgfqpoint{1.921171in}{0.663215in}}{\pgfqpoint{1.923945in}{0.669910in}}{\pgfqpoint{1.923945in}{0.676890in}}%
\pgfpathcurveto{\pgfqpoint{1.923945in}{0.683870in}}{\pgfqpoint{1.921171in}{0.690565in}}{\pgfqpoint{1.916236in}{0.695501in}}%
\pgfpathcurveto{\pgfqpoint{1.911300in}{0.700437in}}{\pgfqpoint{1.904605in}{0.703210in}}{\pgfqpoint{1.897625in}{0.703210in}}%
\pgfpathcurveto{\pgfqpoint{1.890645in}{0.703210in}}{\pgfqpoint{1.883950in}{0.700437in}}{\pgfqpoint{1.879014in}{0.695501in}}%
\pgfpathcurveto{\pgfqpoint{1.874079in}{0.690565in}}{\pgfqpoint{1.871305in}{0.683870in}}{\pgfqpoint{1.871305in}{0.676890in}}%
\pgfpathcurveto{\pgfqpoint{1.871305in}{0.669910in}}{\pgfqpoint{1.874079in}{0.663215in}}{\pgfqpoint{1.879014in}{0.658280in}}%
\pgfpathcurveto{\pgfqpoint{1.883950in}{0.653344in}}{\pgfqpoint{1.890645in}{0.650571in}}{\pgfqpoint{1.897625in}{0.650571in}}%
\pgfpathclose%
\pgfusepath{stroke,fill}%
\end{pgfscope}%
\begin{pgfscope}%
\pgfpathrectangle{\pgfqpoint{0.681500in}{0.420000in}}{\pgfqpoint{1.621500in}{1.050000in}} %
\pgfusepath{clip}%
\pgfsetbuttcap%
\pgfsetroundjoin%
\definecolor{currentfill}{rgb}{0.498039,0.498039,0.498039}%
\pgfsetfillcolor{currentfill}%
\pgfsetlinewidth{1.138252pt}%
\definecolor{currentstroke}{rgb}{0.498039,0.498039,0.498039}%
\pgfsetstrokecolor{currentstroke}%
\pgfsetdash{}{0pt}%
\pgfpathmoveto{\pgfqpoint{2.167875in}{0.665287in}}%
\pgfpathcurveto{\pgfqpoint{2.174855in}{0.665287in}}{\pgfqpoint{2.181550in}{0.668060in}}{\pgfqpoint{2.186486in}{0.672996in}}%
\pgfpathcurveto{\pgfqpoint{2.191421in}{0.677931in}}{\pgfqpoint{2.194195in}{0.684626in}}{\pgfqpoint{2.194195in}{0.691607in}}%
\pgfpathcurveto{\pgfqpoint{2.194195in}{0.698587in}}{\pgfqpoint{2.191421in}{0.705282in}}{\pgfqpoint{2.186486in}{0.710217in}}%
\pgfpathcurveto{\pgfqpoint{2.181550in}{0.715153in}}{\pgfqpoint{2.174855in}{0.717926in}}{\pgfqpoint{2.167875in}{0.717926in}}%
\pgfpathcurveto{\pgfqpoint{2.160895in}{0.717926in}}{\pgfqpoint{2.154200in}{0.715153in}}{\pgfqpoint{2.149264in}{0.710217in}}%
\pgfpathcurveto{\pgfqpoint{2.144329in}{0.705282in}}{\pgfqpoint{2.141555in}{0.698587in}}{\pgfqpoint{2.141555in}{0.691607in}}%
\pgfpathcurveto{\pgfqpoint{2.141555in}{0.684626in}}{\pgfqpoint{2.144329in}{0.677931in}}{\pgfqpoint{2.149264in}{0.672996in}}%
\pgfpathcurveto{\pgfqpoint{2.154200in}{0.668060in}}{\pgfqpoint{2.160895in}{0.665287in}}{\pgfqpoint{2.167875in}{0.665287in}}%
\pgfpathclose%
\pgfusepath{stroke,fill}%
\end{pgfscope}%
\begin{pgfscope}%
\pgfpathrectangle{\pgfqpoint{0.681500in}{0.420000in}}{\pgfqpoint{1.621500in}{1.050000in}} %
\pgfusepath{clip}%
\pgfsetroundcap%
\pgfsetroundjoin%
\pgfsetlinewidth{1.517670pt}%
\definecolor{currentstroke}{rgb}{0.498039,0.498039,0.498039}%
\pgfsetstrokecolor{currentstroke}%
\pgfsetdash{}{0pt}%
\pgfpathmoveto{\pgfqpoint{0.816625in}{0.474796in}}%
\pgfpathlineto{\pgfqpoint{1.086875in}{0.520043in}}%
\pgfpathlineto{\pgfqpoint{1.357125in}{0.572498in}}%
\pgfpathlineto{\pgfqpoint{1.627375in}{0.611712in}}%
\pgfpathlineto{\pgfqpoint{1.897625in}{0.676890in}}%
\pgfpathlineto{\pgfqpoint{2.167875in}{0.691607in}}%
\pgfusepath{stroke}%
\end{pgfscope}%
\begin{pgfscope}%
\pgfpathrectangle{\pgfqpoint{0.681500in}{0.420000in}}{\pgfqpoint{1.621500in}{1.050000in}} %
\pgfusepath{clip}%
\pgfsetroundcap%
\pgfsetroundjoin%
\pgfsetlinewidth{2.168100pt}%
\definecolor{currentstroke}{rgb}{0.498039,0.498039,0.498039}%
\pgfsetstrokecolor{currentstroke}%
\pgfsetdash{}{0pt}%
\pgfpathmoveto{\pgfqpoint{0.816625in}{0.471362in}}%
\pgfpathlineto{\pgfqpoint{0.816625in}{0.478814in}}%
\pgfusepath{stroke}%
\end{pgfscope}%
\begin{pgfscope}%
\pgfpathrectangle{\pgfqpoint{0.681500in}{0.420000in}}{\pgfqpoint{1.621500in}{1.050000in}} %
\pgfusepath{clip}%
\pgfsetroundcap%
\pgfsetroundjoin%
\pgfsetlinewidth{2.168100pt}%
\definecolor{currentstroke}{rgb}{0.498039,0.498039,0.498039}%
\pgfsetstrokecolor{currentstroke}%
\pgfsetdash{}{0pt}%
\pgfpathmoveto{\pgfqpoint{1.086875in}{0.513940in}}%
\pgfpathlineto{\pgfqpoint{1.086875in}{0.520604in}}%
\pgfusepath{stroke}%
\end{pgfscope}%
\begin{pgfscope}%
\pgfpathrectangle{\pgfqpoint{0.681500in}{0.420000in}}{\pgfqpoint{1.621500in}{1.050000in}} %
\pgfusepath{clip}%
\pgfsetroundcap%
\pgfsetroundjoin%
\pgfsetlinewidth{2.168100pt}%
\definecolor{currentstroke}{rgb}{0.498039,0.498039,0.498039}%
\pgfsetstrokecolor{currentstroke}%
\pgfsetdash{}{0pt}%
\pgfpathmoveto{\pgfqpoint{1.357125in}{0.566912in}}%
\pgfpathlineto{\pgfqpoint{1.357125in}{0.578000in}}%
\pgfusepath{stroke}%
\end{pgfscope}%
\begin{pgfscope}%
\pgfpathrectangle{\pgfqpoint{0.681500in}{0.420000in}}{\pgfqpoint{1.621500in}{1.050000in}} %
\pgfusepath{clip}%
\pgfsetroundcap%
\pgfsetroundjoin%
\pgfsetlinewidth{2.168100pt}%
\definecolor{currentstroke}{rgb}{0.498039,0.498039,0.498039}%
\pgfsetstrokecolor{currentstroke}%
\pgfsetdash{}{0pt}%
\pgfpathmoveto{\pgfqpoint{1.627375in}{0.603977in}}%
\pgfpathlineto{\pgfqpoint{1.627375in}{0.630592in}}%
\pgfusepath{stroke}%
\end{pgfscope}%
\begin{pgfscope}%
\pgfpathrectangle{\pgfqpoint{0.681500in}{0.420000in}}{\pgfqpoint{1.621500in}{1.050000in}} %
\pgfusepath{clip}%
\pgfsetroundcap%
\pgfsetroundjoin%
\pgfsetlinewidth{2.168100pt}%
\definecolor{currentstroke}{rgb}{0.498039,0.498039,0.498039}%
\pgfsetstrokecolor{currentstroke}%
\pgfsetdash{}{0pt}%
\pgfpathmoveto{\pgfqpoint{1.897625in}{0.664677in}}%
\pgfpathlineto{\pgfqpoint{1.897625in}{0.682733in}}%
\pgfusepath{stroke}%
\end{pgfscope}%
\begin{pgfscope}%
\pgfpathrectangle{\pgfqpoint{0.681500in}{0.420000in}}{\pgfqpoint{1.621500in}{1.050000in}} %
\pgfusepath{clip}%
\pgfsetroundcap%
\pgfsetroundjoin%
\pgfsetlinewidth{2.168100pt}%
\definecolor{currentstroke}{rgb}{0.498039,0.498039,0.498039}%
\pgfsetstrokecolor{currentstroke}%
\pgfsetdash{}{0pt}%
\pgfpathmoveto{\pgfqpoint{2.167875in}{0.684189in}}%
\pgfpathlineto{\pgfqpoint{2.167875in}{0.708548in}}%
\pgfusepath{stroke}%
\end{pgfscope}%
\begin{pgfscope}%
\pgfpathrectangle{\pgfqpoint{0.681500in}{0.420000in}}{\pgfqpoint{1.621500in}{1.050000in}} %
\pgfusepath{clip}%
\pgfsetbuttcap%
\pgfsetroundjoin%
\pgfsetlinewidth{2.007500pt}%
\definecolor{currentstroke}{rgb}{0.890196,0.466667,0.760784}%
\pgfsetstrokecolor{currentstroke}%
\pgfsetdash{{7.400000pt}{3.200000pt}}{0.000000pt}%
\pgfpathmoveto{\pgfqpoint{0.681500in}{1.291468in}}%
\pgfpathlineto{\pgfqpoint{2.303000in}{1.291468in}}%
\pgfusepath{stroke}%
\end{pgfscope}%
\end{pgfpicture}%
\makeatother%
\endgroup%

%% file: images/plots/exec_time_Sarcos.pgf
\begingroup%
\makeatletter%
\begin{pgfpicture}%
\pgfpathrectangle{\pgfpointorigin}{\pgfqpoint{2.150000in}{1.500000in}}%
\pgfusepath{use as bounding box, clip}%
\begin{pgfscope}%
\pgfsetbuttcap%
\pgfsetmiterjoin%
\pgfsetlinewidth{0.000000pt}%
\definecolor{currentstroke}{rgb}{0.000000,0.000000,0.000000}%
\pgfsetstrokecolor{currentstroke}%
\pgfsetstrokeopacity{0.000000}%
\pgfsetdash{}{0pt}%
\pgfpathmoveto{\pgfqpoint{0.000000in}{0.000000in}}%
\pgfpathlineto{\pgfqpoint{2.150000in}{0.000000in}}%
\pgfpathlineto{\pgfqpoint{2.150000in}{1.500000in}}%
\pgfpathlineto{\pgfqpoint{0.000000in}{1.500000in}}%
\pgfpathclose%
\pgfusepath{}%
\end{pgfscope}%
\begin{pgfscope}%
\pgfsetbuttcap%
\pgfsetmiterjoin%
\pgfsetlinewidth{0.000000pt}%
\definecolor{currentstroke}{rgb}{0.000000,0.000000,0.000000}%
\pgfsetstrokecolor{currentstroke}%
\pgfsetstrokeopacity{0.000000}%
\pgfsetdash{}{0pt}%
\pgfpathmoveto{\pgfqpoint{0.516000in}{0.420000in}}%
\pgfpathlineto{\pgfqpoint{2.107000in}{0.420000in}}%
\pgfpathlineto{\pgfqpoint{2.107000in}{1.470000in}}%
\pgfpathlineto{\pgfqpoint{0.516000in}{1.470000in}}%
\pgfpathclose%
\pgfusepath{}%
\end{pgfscope}%
\begin{pgfscope}%
\pgfpathrectangle{\pgfqpoint{0.516000in}{0.420000in}}{\pgfqpoint{1.591000in}{1.050000in}} %
\pgfusepath{clip}%
\pgfsetroundcap%
\pgfsetroundjoin%
\pgfsetlinewidth{0.803000pt}%
\definecolor{currentstroke}{rgb}{0.800000,0.800000,0.800000}%
\pgfsetstrokecolor{currentstroke}%
\pgfsetdash{}{0pt}%
\pgfpathmoveto{\pgfqpoint{0.648583in}{0.420000in}}%
\pgfpathlineto{\pgfqpoint{0.648583in}{1.470000in}}%
\pgfusepath{stroke}%
\end{pgfscope}%
\begin{pgfscope}%
\definecolor{textcolor}{rgb}{0.150000,0.150000,0.150000}%
\pgfsetstrokecolor{textcolor}%
\pgfsetfillcolor{textcolor}%
\pgftext[x=0.648583in,y=0.304722in,,top]{\color{textcolor}\fontsize{8.800000}{10.560000}\selectfont 5}%
\end{pgfscope}%
\begin{pgfscope}%
\pgfpathrectangle{\pgfqpoint{0.516000in}{0.420000in}}{\pgfqpoint{1.591000in}{1.050000in}} %
\pgfusepath{clip}%
\pgfsetroundcap%
\pgfsetroundjoin%
\pgfsetlinewidth{0.803000pt}%
\definecolor{currentstroke}{rgb}{0.800000,0.800000,0.800000}%
\pgfsetstrokecolor{currentstroke}%
\pgfsetdash{}{0pt}%
\pgfpathmoveto{\pgfqpoint{0.913750in}{0.420000in}}%
\pgfpathlineto{\pgfqpoint{0.913750in}{1.470000in}}%
\pgfusepath{stroke}%
\end{pgfscope}%
\begin{pgfscope}%
\definecolor{textcolor}{rgb}{0.150000,0.150000,0.150000}%
\pgfsetstrokecolor{textcolor}%
\pgfsetfillcolor{textcolor}%
\pgftext[x=0.913750in,y=0.304722in,,top]{\color{textcolor}\fontsize{8.800000}{10.560000}\selectfont 10}%
\end{pgfscope}%
\begin{pgfscope}%
\pgfpathrectangle{\pgfqpoint{0.516000in}{0.420000in}}{\pgfqpoint{1.591000in}{1.050000in}} %
\pgfusepath{clip}%
\pgfsetroundcap%
\pgfsetroundjoin%
\pgfsetlinewidth{0.803000pt}%
\definecolor{currentstroke}{rgb}{0.800000,0.800000,0.800000}%
\pgfsetstrokecolor{currentstroke}%
\pgfsetdash{}{0pt}%
\pgfpathmoveto{\pgfqpoint{1.178917in}{0.420000in}}%
\pgfpathlineto{\pgfqpoint{1.178917in}{1.470000in}}%
\pgfusepath{stroke}%
\end{pgfscope}%
\begin{pgfscope}%
\definecolor{textcolor}{rgb}{0.150000,0.150000,0.150000}%
\pgfsetstrokecolor{textcolor}%
\pgfsetfillcolor{textcolor}%
\pgftext[x=1.178917in,y=0.304722in,,top]{\color{textcolor}\fontsize{8.800000}{10.560000}\selectfont 20}%
\end{pgfscope}%
\begin{pgfscope}%
\pgfpathrectangle{\pgfqpoint{0.516000in}{0.420000in}}{\pgfqpoint{1.591000in}{1.050000in}} %
\pgfusepath{clip}%
\pgfsetroundcap%
\pgfsetroundjoin%
\pgfsetlinewidth{0.803000pt}%
\definecolor{currentstroke}{rgb}{0.800000,0.800000,0.800000}%
\pgfsetstrokecolor{currentstroke}%
\pgfsetdash{}{0pt}%
\pgfpathmoveto{\pgfqpoint{1.444083in}{0.420000in}}%
\pgfpathlineto{\pgfqpoint{1.444083in}{1.470000in}}%
\pgfusepath{stroke}%
\end{pgfscope}%
\begin{pgfscope}%
\definecolor{textcolor}{rgb}{0.150000,0.150000,0.150000}%
\pgfsetstrokecolor{textcolor}%
\pgfsetfillcolor{textcolor}%
\pgftext[x=1.444083in,y=0.304722in,,top]{\color{textcolor}\fontsize{8.800000}{10.560000}\selectfont 30}%
\end{pgfscope}%
\begin{pgfscope}%
\pgfpathrectangle{\pgfqpoint{0.516000in}{0.420000in}}{\pgfqpoint{1.591000in}{1.050000in}} %
\pgfusepath{clip}%
\pgfsetroundcap%
\pgfsetroundjoin%
\pgfsetlinewidth{0.803000pt}%
\definecolor{currentstroke}{rgb}{0.800000,0.800000,0.800000}%
\pgfsetstrokecolor{currentstroke}%
\pgfsetdash{}{0pt}%
\pgfpathmoveto{\pgfqpoint{1.709250in}{0.420000in}}%
\pgfpathlineto{\pgfqpoint{1.709250in}{1.470000in}}%
\pgfusepath{stroke}%
\end{pgfscope}%
\begin{pgfscope}%
\definecolor{textcolor}{rgb}{0.150000,0.150000,0.150000}%
\pgfsetstrokecolor{textcolor}%
\pgfsetfillcolor{textcolor}%
\pgftext[x=1.709250in,y=0.304722in,,top]{\color{textcolor}\fontsize{8.800000}{10.560000}\selectfont 40}%
\end{pgfscope}%
\begin{pgfscope}%
\pgfpathrectangle{\pgfqpoint{0.516000in}{0.420000in}}{\pgfqpoint{1.591000in}{1.050000in}} %
\pgfusepath{clip}%
\pgfsetroundcap%
\pgfsetroundjoin%
\pgfsetlinewidth{0.803000pt}%
\definecolor{currentstroke}{rgb}{0.800000,0.800000,0.800000}%
\pgfsetstrokecolor{currentstroke}%
\pgfsetdash{}{0pt}%
\pgfpathmoveto{\pgfqpoint{1.974417in}{0.420000in}}%
\pgfpathlineto{\pgfqpoint{1.974417in}{1.470000in}}%
\pgfusepath{stroke}%
\end{pgfscope}%
\begin{pgfscope}%
\definecolor{textcolor}{rgb}{0.150000,0.150000,0.150000}%
\pgfsetstrokecolor{textcolor}%
\pgfsetfillcolor{textcolor}%
\pgftext[x=1.974417in,y=0.304722in,,top]{\color{textcolor}\fontsize{8.800000}{10.560000}\selectfont 50}%
\end{pgfscope}%
\begin{pgfscope}%
\definecolor{textcolor}{rgb}{0.150000,0.150000,0.150000}%
\pgfsetstrokecolor{textcolor}%
\pgfsetfillcolor{textcolor}%
\pgftext[x=1.311500in,y=0.138056in,,top]{\color{textcolor}\fontsize{9.600000}{11.520000}\selectfont Reduction in \%}%
\end{pgfscope}%
\begin{pgfscope}%
\pgfpathrectangle{\pgfqpoint{0.516000in}{0.420000in}}{\pgfqpoint{1.591000in}{1.050000in}} %
\pgfusepath{clip}%
\pgfsetroundcap%
\pgfsetroundjoin%
\pgfsetlinewidth{0.803000pt}%
\definecolor{currentstroke}{rgb}{0.800000,0.800000,0.800000}%
\pgfsetstrokecolor{currentstroke}%
\pgfsetdash{}{0pt}%
\pgfpathmoveto{\pgfqpoint{0.516000in}{0.420000in}}%
\pgfpathlineto{\pgfqpoint{2.107000in}{0.420000in}}%
\pgfusepath{stroke}%
\end{pgfscope}%
\begin{pgfscope}%
\definecolor{textcolor}{rgb}{0.150000,0.150000,0.150000}%
\pgfsetstrokecolor{textcolor}%
\pgfsetfillcolor{textcolor}%
\pgftext[x=0.026690in,y=0.376597in,left,base]{\color{textcolor}\fontsize{8.800000}{10.560000}\selectfont 00h 00m}%
\end{pgfscope}%
\begin{pgfscope}%
\pgfpathrectangle{\pgfqpoint{0.516000in}{0.420000in}}{\pgfqpoint{1.591000in}{1.050000in}} %
\pgfusepath{clip}%
\pgfsetroundcap%
\pgfsetroundjoin%
\pgfsetlinewidth{0.803000pt}%
\definecolor{currentstroke}{rgb}{0.800000,0.800000,0.800000}%
\pgfsetstrokecolor{currentstroke}%
\pgfsetdash{}{0pt}%
\pgfpathmoveto{\pgfqpoint{0.516000in}{0.722825in}}%
\pgfpathlineto{\pgfqpoint{2.107000in}{0.722825in}}%
\pgfusepath{stroke}%
\end{pgfscope}%
\begin{pgfscope}%
\definecolor{textcolor}{rgb}{0.150000,0.150000,0.150000}%
\pgfsetstrokecolor{textcolor}%
\pgfsetfillcolor{textcolor}%
\pgftext[x=0.126612in,y=0.679423in,left,base]{\color{textcolor}\fontsize{8.800000}{10.560000}\selectfont 2d 21h}%
\end{pgfscope}%
\begin{pgfscope}%
\pgfpathrectangle{\pgfqpoint{0.516000in}{0.420000in}}{\pgfqpoint{1.591000in}{1.050000in}} %
\pgfusepath{clip}%
\pgfsetroundcap%
\pgfsetroundjoin%
\pgfsetlinewidth{0.803000pt}%
\definecolor{currentstroke}{rgb}{0.800000,0.800000,0.800000}%
\pgfsetstrokecolor{currentstroke}%
\pgfsetdash{}{0pt}%
\pgfpathmoveto{\pgfqpoint{0.516000in}{1.025651in}}%
\pgfpathlineto{\pgfqpoint{2.107000in}{1.025651in}}%
\pgfusepath{stroke}%
\end{pgfscope}%
\begin{pgfscope}%
\definecolor{textcolor}{rgb}{0.150000,0.150000,0.150000}%
\pgfsetstrokecolor{textcolor}%
\pgfsetfillcolor{textcolor}%
\pgftext[x=0.126612in,y=0.982248in,left,base]{\color{textcolor}\fontsize{8.800000}{10.560000}\selectfont 5d 19h}%
\end{pgfscope}%
\begin{pgfscope}%
\pgfpathrectangle{\pgfqpoint{0.516000in}{0.420000in}}{\pgfqpoint{1.591000in}{1.050000in}} %
\pgfusepath{clip}%
\pgfsetroundcap%
\pgfsetroundjoin%
\pgfsetlinewidth{0.803000pt}%
\definecolor{currentstroke}{rgb}{0.800000,0.800000,0.800000}%
\pgfsetstrokecolor{currentstroke}%
\pgfsetdash{}{0pt}%
\pgfpathmoveto{\pgfqpoint{0.516000in}{1.328476in}}%
\pgfpathlineto{\pgfqpoint{2.107000in}{1.328476in}}%
\pgfusepath{stroke}%
\end{pgfscope}%
\begin{pgfscope}%
\definecolor{textcolor}{rgb}{0.150000,0.150000,0.150000}%
\pgfsetstrokecolor{textcolor}%
\pgfsetfillcolor{textcolor}%
\pgftext[x=0.126612in,y=1.285073in,left,base]{\color{textcolor}\fontsize{8.800000}{10.560000}\selectfont 8d 16h}%
\end{pgfscope}%
\begin{pgfscope}%
\pgfpathrectangle{\pgfqpoint{0.516000in}{0.420000in}}{\pgfqpoint{1.591000in}{1.050000in}} %
\pgfusepath{clip}%
\pgfsetbuttcap%
\pgfsetroundjoin%
\definecolor{currentfill}{rgb}{0.737255,0.741176,0.133333}%
\pgfsetfillcolor{currentfill}%
\pgfsetlinewidth{1.138252pt}%
\definecolor{currentstroke}{rgb}{0.737255,0.741176,0.133333}%
\pgfsetstrokecolor{currentstroke}%
\pgfsetdash{}{0pt}%
\pgfpathmoveto{\pgfqpoint{0.648583in}{0.441195in}}%
\pgfpathcurveto{\pgfqpoint{0.655563in}{0.441195in}}{\pgfqpoint{0.662258in}{0.443968in}}{\pgfqpoint{0.667194in}{0.448903in}}%
\pgfpathcurveto{\pgfqpoint{0.672130in}{0.453839in}}{\pgfqpoint{0.674903in}{0.460534in}}{\pgfqpoint{0.674903in}{0.467514in}}%
\pgfpathcurveto{\pgfqpoint{0.674903in}{0.474494in}}{\pgfqpoint{0.672130in}{0.481189in}}{\pgfqpoint{0.667194in}{0.486125in}}%
\pgfpathcurveto{\pgfqpoint{0.662258in}{0.491061in}}{\pgfqpoint{0.655563in}{0.493834in}}{\pgfqpoint{0.648583in}{0.493834in}}%
\pgfpathcurveto{\pgfqpoint{0.641603in}{0.493834in}}{\pgfqpoint{0.634908in}{0.491061in}}{\pgfqpoint{0.629973in}{0.486125in}}%
\pgfpathcurveto{\pgfqpoint{0.625037in}{0.481189in}}{\pgfqpoint{0.622264in}{0.474494in}}{\pgfqpoint{0.622264in}{0.467514in}}%
\pgfpathcurveto{\pgfqpoint{0.622264in}{0.460534in}}{\pgfqpoint{0.625037in}{0.453839in}}{\pgfqpoint{0.629973in}{0.448903in}}%
\pgfpathcurveto{\pgfqpoint{0.634908in}{0.443968in}}{\pgfqpoint{0.641603in}{0.441195in}}{\pgfqpoint{0.648583in}{0.441195in}}%
\pgfpathclose%
\pgfusepath{stroke,fill}%
\end{pgfscope}%
\begin{pgfscope}%
\pgfpathrectangle{\pgfqpoint{0.516000in}{0.420000in}}{\pgfqpoint{1.591000in}{1.050000in}} %
\pgfusepath{clip}%
\pgfsetbuttcap%
\pgfsetroundjoin%
\definecolor{currentfill}{rgb}{0.737255,0.741176,0.133333}%
\pgfsetfillcolor{currentfill}%
\pgfsetlinewidth{1.138252pt}%
\definecolor{currentstroke}{rgb}{0.737255,0.741176,0.133333}%
\pgfsetstrokecolor{currentstroke}%
\pgfsetdash{}{0pt}%
\pgfpathmoveto{\pgfqpoint{0.913750in}{0.486420in}}%
\pgfpathcurveto{\pgfqpoint{0.920730in}{0.486420in}}{\pgfqpoint{0.927425in}{0.489193in}}{\pgfqpoint{0.932361in}{0.494128in}}%
\pgfpathcurveto{\pgfqpoint{0.937296in}{0.499064in}}{\pgfqpoint{0.940070in}{0.505759in}}{\pgfqpoint{0.940070in}{0.512739in}}%
\pgfpathcurveto{\pgfqpoint{0.940070in}{0.519719in}}{\pgfqpoint{0.937296in}{0.526414in}}{\pgfqpoint{0.932361in}{0.531350in}}%
\pgfpathcurveto{\pgfqpoint{0.927425in}{0.536286in}}{\pgfqpoint{0.920730in}{0.539059in}}{\pgfqpoint{0.913750in}{0.539059in}}%
\pgfpathcurveto{\pgfqpoint{0.906770in}{0.539059in}}{\pgfqpoint{0.900075in}{0.536286in}}{\pgfqpoint{0.895139in}{0.531350in}}%
\pgfpathcurveto{\pgfqpoint{0.890204in}{0.526414in}}{\pgfqpoint{0.887430in}{0.519719in}}{\pgfqpoint{0.887430in}{0.512739in}}%
\pgfpathcurveto{\pgfqpoint{0.887430in}{0.505759in}}{\pgfqpoint{0.890204in}{0.499064in}}{\pgfqpoint{0.895139in}{0.494128in}}%
\pgfpathcurveto{\pgfqpoint{0.900075in}{0.489193in}}{\pgfqpoint{0.906770in}{0.486420in}}{\pgfqpoint{0.913750in}{0.486420in}}%
\pgfpathclose%
\pgfusepath{stroke,fill}%
\end{pgfscope}%
\begin{pgfscope}%
\pgfpathrectangle{\pgfqpoint{0.516000in}{0.420000in}}{\pgfqpoint{1.591000in}{1.050000in}} %
\pgfusepath{clip}%
\pgfsetbuttcap%
\pgfsetroundjoin%
\definecolor{currentfill}{rgb}{0.737255,0.741176,0.133333}%
\pgfsetfillcolor{currentfill}%
\pgfsetlinewidth{1.138252pt}%
\definecolor{currentstroke}{rgb}{0.737255,0.741176,0.133333}%
\pgfsetstrokecolor{currentstroke}%
\pgfsetdash{}{0pt}%
\pgfpathmoveto{\pgfqpoint{1.178917in}{0.582963in}}%
\pgfpathcurveto{\pgfqpoint{1.185897in}{0.582963in}}{\pgfqpoint{1.192592in}{0.585736in}}{\pgfqpoint{1.197527in}{0.590672in}}%
\pgfpathcurveto{\pgfqpoint{1.202463in}{0.595607in}}{\pgfqpoint{1.205236in}{0.602303in}}{\pgfqpoint{1.205236in}{0.609283in}}%
\pgfpathcurveto{\pgfqpoint{1.205236in}{0.616263in}}{\pgfqpoint{1.202463in}{0.622958in}}{\pgfqpoint{1.197527in}{0.627893in}}%
\pgfpathcurveto{\pgfqpoint{1.192592in}{0.632829in}}{\pgfqpoint{1.185897in}{0.635602in}}{\pgfqpoint{1.178917in}{0.635602in}}%
\pgfpathcurveto{\pgfqpoint{1.171937in}{0.635602in}}{\pgfqpoint{1.165242in}{0.632829in}}{\pgfqpoint{1.160306in}{0.627893in}}%
\pgfpathcurveto{\pgfqpoint{1.155370in}{0.622958in}}{\pgfqpoint{1.152597in}{0.616263in}}{\pgfqpoint{1.152597in}{0.609283in}}%
\pgfpathcurveto{\pgfqpoint{1.152597in}{0.602303in}}{\pgfqpoint{1.155370in}{0.595607in}}{\pgfqpoint{1.160306in}{0.590672in}}%
\pgfpathcurveto{\pgfqpoint{1.165242in}{0.585736in}}{\pgfqpoint{1.171937in}{0.582963in}}{\pgfqpoint{1.178917in}{0.582963in}}%
\pgfpathclose%
\pgfusepath{stroke,fill}%
\end{pgfscope}%
\begin{pgfscope}%
\pgfpathrectangle{\pgfqpoint{0.516000in}{0.420000in}}{\pgfqpoint{1.591000in}{1.050000in}} %
\pgfusepath{clip}%
\pgfsetbuttcap%
\pgfsetroundjoin%
\definecolor{currentfill}{rgb}{0.737255,0.741176,0.133333}%
\pgfsetfillcolor{currentfill}%
\pgfsetlinewidth{1.138252pt}%
\definecolor{currentstroke}{rgb}{0.737255,0.741176,0.133333}%
\pgfsetstrokecolor{currentstroke}%
\pgfsetdash{}{0pt}%
\pgfpathmoveto{\pgfqpoint{1.444083in}{0.655481in}}%
\pgfpathcurveto{\pgfqpoint{1.451063in}{0.655481in}}{\pgfqpoint{1.457758in}{0.658254in}}{\pgfqpoint{1.462694in}{0.663190in}}%
\pgfpathcurveto{\pgfqpoint{1.467630in}{0.668126in}}{\pgfqpoint{1.470403in}{0.674821in}}{\pgfqpoint{1.470403in}{0.681801in}}%
\pgfpathcurveto{\pgfqpoint{1.470403in}{0.688781in}}{\pgfqpoint{1.467630in}{0.695476in}}{\pgfqpoint{1.462694in}{0.700411in}}%
\pgfpathcurveto{\pgfqpoint{1.457758in}{0.705347in}}{\pgfqpoint{1.451063in}{0.708120in}}{\pgfqpoint{1.444083in}{0.708120in}}%
\pgfpathcurveto{\pgfqpoint{1.437103in}{0.708120in}}{\pgfqpoint{1.430408in}{0.705347in}}{\pgfqpoint{1.425473in}{0.700411in}}%
\pgfpathcurveto{\pgfqpoint{1.420537in}{0.695476in}}{\pgfqpoint{1.417764in}{0.688781in}}{\pgfqpoint{1.417764in}{0.681801in}}%
\pgfpathcurveto{\pgfqpoint{1.417764in}{0.674821in}}{\pgfqpoint{1.420537in}{0.668126in}}{\pgfqpoint{1.425473in}{0.663190in}}%
\pgfpathcurveto{\pgfqpoint{1.430408in}{0.658254in}}{\pgfqpoint{1.437103in}{0.655481in}}{\pgfqpoint{1.444083in}{0.655481in}}%
\pgfpathclose%
\pgfusepath{stroke,fill}%
\end{pgfscope}%
\begin{pgfscope}%
\pgfpathrectangle{\pgfqpoint{0.516000in}{0.420000in}}{\pgfqpoint{1.591000in}{1.050000in}} %
\pgfusepath{clip}%
\pgfsetbuttcap%
\pgfsetroundjoin%
\definecolor{currentfill}{rgb}{0.737255,0.741176,0.133333}%
\pgfsetfillcolor{currentfill}%
\pgfsetlinewidth{1.138252pt}%
\definecolor{currentstroke}{rgb}{0.737255,0.741176,0.133333}%
\pgfsetstrokecolor{currentstroke}%
\pgfsetdash{}{0pt}%
\pgfpathmoveto{\pgfqpoint{1.709250in}{0.751674in}}%
\pgfpathcurveto{\pgfqpoint{1.716230in}{0.751674in}}{\pgfqpoint{1.722925in}{0.754447in}}{\pgfqpoint{1.727861in}{0.759383in}}%
\pgfpathcurveto{\pgfqpoint{1.732796in}{0.764318in}}{\pgfqpoint{1.735570in}{0.771014in}}{\pgfqpoint{1.735570in}{0.777994in}}%
\pgfpathcurveto{\pgfqpoint{1.735570in}{0.784974in}}{\pgfqpoint{1.732796in}{0.791669in}}{\pgfqpoint{1.727861in}{0.796604in}}%
\pgfpathcurveto{\pgfqpoint{1.722925in}{0.801540in}}{\pgfqpoint{1.716230in}{0.804313in}}{\pgfqpoint{1.709250in}{0.804313in}}%
\pgfpathcurveto{\pgfqpoint{1.702270in}{0.804313in}}{\pgfqpoint{1.695575in}{0.801540in}}{\pgfqpoint{1.690639in}{0.796604in}}%
\pgfpathcurveto{\pgfqpoint{1.685704in}{0.791669in}}{\pgfqpoint{1.682930in}{0.784974in}}{\pgfqpoint{1.682930in}{0.777994in}}%
\pgfpathcurveto{\pgfqpoint{1.682930in}{0.771014in}}{\pgfqpoint{1.685704in}{0.764318in}}{\pgfqpoint{1.690639in}{0.759383in}}%
\pgfpathcurveto{\pgfqpoint{1.695575in}{0.754447in}}{\pgfqpoint{1.702270in}{0.751674in}}{\pgfqpoint{1.709250in}{0.751674in}}%
\pgfpathclose%
\pgfusepath{stroke,fill}%
\end{pgfscope}%
\begin{pgfscope}%
\pgfpathrectangle{\pgfqpoint{0.516000in}{0.420000in}}{\pgfqpoint{1.591000in}{1.050000in}} %
\pgfusepath{clip}%
\pgfsetbuttcap%
\pgfsetroundjoin%
\definecolor{currentfill}{rgb}{0.737255,0.741176,0.133333}%
\pgfsetfillcolor{currentfill}%
\pgfsetlinewidth{1.138252pt}%
\definecolor{currentstroke}{rgb}{0.737255,0.741176,0.133333}%
\pgfsetstrokecolor{currentstroke}%
\pgfsetdash{}{0pt}%
\pgfpathmoveto{\pgfqpoint{1.974417in}{0.848305in}}%
\pgfpathcurveto{\pgfqpoint{1.981397in}{0.848305in}}{\pgfqpoint{1.988092in}{0.851078in}}{\pgfqpoint{1.993027in}{0.856014in}}%
\pgfpathcurveto{\pgfqpoint{1.997963in}{0.860950in}}{\pgfqpoint{2.000736in}{0.867645in}}{\pgfqpoint{2.000736in}{0.874625in}}%
\pgfpathcurveto{\pgfqpoint{2.000736in}{0.881605in}}{\pgfqpoint{1.997963in}{0.888300in}}{\pgfqpoint{1.993027in}{0.893236in}}%
\pgfpathcurveto{\pgfqpoint{1.988092in}{0.898171in}}{\pgfqpoint{1.981397in}{0.900944in}}{\pgfqpoint{1.974417in}{0.900944in}}%
\pgfpathcurveto{\pgfqpoint{1.967437in}{0.900944in}}{\pgfqpoint{1.960742in}{0.898171in}}{\pgfqpoint{1.955806in}{0.893236in}}%
\pgfpathcurveto{\pgfqpoint{1.950870in}{0.888300in}}{\pgfqpoint{1.948097in}{0.881605in}}{\pgfqpoint{1.948097in}{0.874625in}}%
\pgfpathcurveto{\pgfqpoint{1.948097in}{0.867645in}}{\pgfqpoint{1.950870in}{0.860950in}}{\pgfqpoint{1.955806in}{0.856014in}}%
\pgfpathcurveto{\pgfqpoint{1.960742in}{0.851078in}}{\pgfqpoint{1.967437in}{0.848305in}}{\pgfqpoint{1.974417in}{0.848305in}}%
\pgfpathclose%
\pgfusepath{stroke,fill}%
\end{pgfscope}%
\begin{pgfscope}%
\pgfpathrectangle{\pgfqpoint{0.516000in}{0.420000in}}{\pgfqpoint{1.591000in}{1.050000in}} %
\pgfusepath{clip}%
\pgfsetroundcap%
\pgfsetroundjoin%
\pgfsetlinewidth{1.517670pt}%
\definecolor{currentstroke}{rgb}{0.737255,0.741176,0.133333}%
\pgfsetstrokecolor{currentstroke}%
\pgfsetdash{}{0pt}%
\pgfpathmoveto{\pgfqpoint{0.648583in}{0.467514in}}%
\pgfpathlineto{\pgfqpoint{0.913750in}{0.512739in}}%
\pgfpathlineto{\pgfqpoint{1.178917in}{0.609283in}}%
\pgfpathlineto{\pgfqpoint{1.444083in}{0.681801in}}%
\pgfpathlineto{\pgfqpoint{1.709250in}{0.777994in}}%
\pgfpathlineto{\pgfqpoint{1.974417in}{0.874625in}}%
\pgfusepath{stroke}%
\end{pgfscope}%
\begin{pgfscope}%
\pgfpathrectangle{\pgfqpoint{0.516000in}{0.420000in}}{\pgfqpoint{1.591000in}{1.050000in}} %
\pgfusepath{clip}%
\pgfsetroundcap%
\pgfsetroundjoin%
\pgfsetlinewidth{2.168100pt}%
\definecolor{currentstroke}{rgb}{0.737255,0.741176,0.133333}%
\pgfsetstrokecolor{currentstroke}%
\pgfsetdash{}{0pt}%
\pgfpathmoveto{\pgfqpoint{0.648583in}{0.465194in}}%
\pgfpathlineto{\pgfqpoint{0.648583in}{0.468848in}}%
\pgfusepath{stroke}%
\end{pgfscope}%
\begin{pgfscope}%
\pgfpathrectangle{\pgfqpoint{0.516000in}{0.420000in}}{\pgfqpoint{1.591000in}{1.050000in}} %
\pgfusepath{clip}%
\pgfsetroundcap%
\pgfsetroundjoin%
\pgfsetlinewidth{2.168100pt}%
\definecolor{currentstroke}{rgb}{0.737255,0.741176,0.133333}%
\pgfsetstrokecolor{currentstroke}%
\pgfsetdash{}{0pt}%
\pgfpathmoveto{\pgfqpoint{0.913750in}{0.506319in}}%
\pgfpathlineto{\pgfqpoint{0.913750in}{0.519069in}}%
\pgfusepath{stroke}%
\end{pgfscope}%
\begin{pgfscope}%
\pgfpathrectangle{\pgfqpoint{0.516000in}{0.420000in}}{\pgfqpoint{1.591000in}{1.050000in}} %
\pgfusepath{clip}%
\pgfsetroundcap%
\pgfsetroundjoin%
\pgfsetlinewidth{2.168100pt}%
\definecolor{currentstroke}{rgb}{0.737255,0.741176,0.133333}%
\pgfsetstrokecolor{currentstroke}%
\pgfsetdash{}{0pt}%
\pgfpathmoveto{\pgfqpoint{1.178917in}{0.593950in}}%
\pgfpathlineto{\pgfqpoint{1.178917in}{0.615033in}}%
\pgfusepath{stroke}%
\end{pgfscope}%
\begin{pgfscope}%
\pgfpathrectangle{\pgfqpoint{0.516000in}{0.420000in}}{\pgfqpoint{1.591000in}{1.050000in}} %
\pgfusepath{clip}%
\pgfsetroundcap%
\pgfsetroundjoin%
\pgfsetlinewidth{2.168100pt}%
\definecolor{currentstroke}{rgb}{0.737255,0.741176,0.133333}%
\pgfsetstrokecolor{currentstroke}%
\pgfsetdash{}{0pt}%
\pgfpathmoveto{\pgfqpoint{1.444083in}{0.672380in}}%
\pgfpathlineto{\pgfqpoint{1.444083in}{0.710505in}}%
\pgfusepath{stroke}%
\end{pgfscope}%
\begin{pgfscope}%
\pgfpathrectangle{\pgfqpoint{0.516000in}{0.420000in}}{\pgfqpoint{1.591000in}{1.050000in}} %
\pgfusepath{clip}%
\pgfsetroundcap%
\pgfsetroundjoin%
\pgfsetlinewidth{2.168100pt}%
\definecolor{currentstroke}{rgb}{0.737255,0.741176,0.133333}%
\pgfsetstrokecolor{currentstroke}%
\pgfsetdash{}{0pt}%
\pgfpathmoveto{\pgfqpoint{1.709250in}{0.769459in}}%
\pgfpathlineto{\pgfqpoint{1.709250in}{0.790886in}}%
\pgfusepath{stroke}%
\end{pgfscope}%
\begin{pgfscope}%
\pgfpathrectangle{\pgfqpoint{0.516000in}{0.420000in}}{\pgfqpoint{1.591000in}{1.050000in}} %
\pgfusepath{clip}%
\pgfsetroundcap%
\pgfsetroundjoin%
\pgfsetlinewidth{2.168100pt}%
\definecolor{currentstroke}{rgb}{0.737255,0.741176,0.133333}%
\pgfsetstrokecolor{currentstroke}%
\pgfsetdash{}{0pt}%
\pgfpathmoveto{\pgfqpoint{1.974417in}{0.842782in}}%
\pgfpathlineto{\pgfqpoint{1.974417in}{0.884938in}}%
\pgfusepath{stroke}%
\end{pgfscope}%
\begin{pgfscope}%
\pgfpathrectangle{\pgfqpoint{0.516000in}{0.420000in}}{\pgfqpoint{1.591000in}{1.050000in}} %
\pgfusepath{clip}%
\pgfsetbuttcap%
\pgfsetroundjoin%
\definecolor{currentfill}{rgb}{0.090196,0.745098,0.811765}%
\pgfsetfillcolor{currentfill}%
\pgfsetlinewidth{1.138252pt}%
\definecolor{currentstroke}{rgb}{0.090196,0.745098,0.811765}%
\pgfsetstrokecolor{currentstroke}%
\pgfsetdash{}{0pt}%
\pgfpathmoveto{\pgfqpoint{0.648583in}{0.443179in}}%
\pgfpathcurveto{\pgfqpoint{0.655563in}{0.443179in}}{\pgfqpoint{0.662258in}{0.445952in}}{\pgfqpoint{0.667194in}{0.450888in}}%
\pgfpathcurveto{\pgfqpoint{0.672130in}{0.455823in}}{\pgfqpoint{0.674903in}{0.462518in}}{\pgfqpoint{0.674903in}{0.469498in}}%
\pgfpathcurveto{\pgfqpoint{0.674903in}{0.476478in}}{\pgfqpoint{0.672130in}{0.483174in}}{\pgfqpoint{0.667194in}{0.488109in}}%
\pgfpathcurveto{\pgfqpoint{0.662258in}{0.493045in}}{\pgfqpoint{0.655563in}{0.495818in}}{\pgfqpoint{0.648583in}{0.495818in}}%
\pgfpathcurveto{\pgfqpoint{0.641603in}{0.495818in}}{\pgfqpoint{0.634908in}{0.493045in}}{\pgfqpoint{0.629973in}{0.488109in}}%
\pgfpathcurveto{\pgfqpoint{0.625037in}{0.483174in}}{\pgfqpoint{0.622264in}{0.476478in}}{\pgfqpoint{0.622264in}{0.469498in}}%
\pgfpathcurveto{\pgfqpoint{0.622264in}{0.462518in}}{\pgfqpoint{0.625037in}{0.455823in}}{\pgfqpoint{0.629973in}{0.450888in}}%
\pgfpathcurveto{\pgfqpoint{0.634908in}{0.445952in}}{\pgfqpoint{0.641603in}{0.443179in}}{\pgfqpoint{0.648583in}{0.443179in}}%
\pgfpathclose%
\pgfusepath{stroke,fill}%
\end{pgfscope}%
\begin{pgfscope}%
\pgfpathrectangle{\pgfqpoint{0.516000in}{0.420000in}}{\pgfqpoint{1.591000in}{1.050000in}} %
\pgfusepath{clip}%
\pgfsetbuttcap%
\pgfsetroundjoin%
\definecolor{currentfill}{rgb}{0.090196,0.745098,0.811765}%
\pgfsetfillcolor{currentfill}%
\pgfsetlinewidth{1.138252pt}%
\definecolor{currentstroke}{rgb}{0.090196,0.745098,0.811765}%
\pgfsetstrokecolor{currentstroke}%
\pgfsetdash{}{0pt}%
\pgfpathmoveto{\pgfqpoint{0.913750in}{0.488328in}}%
\pgfpathcurveto{\pgfqpoint{0.920730in}{0.488328in}}{\pgfqpoint{0.927425in}{0.491102in}}{\pgfqpoint{0.932361in}{0.496037in}}%
\pgfpathcurveto{\pgfqpoint{0.937296in}{0.500973in}}{\pgfqpoint{0.940070in}{0.507668in}}{\pgfqpoint{0.940070in}{0.514648in}}%
\pgfpathcurveto{\pgfqpoint{0.940070in}{0.521628in}}{\pgfqpoint{0.937296in}{0.528323in}}{\pgfqpoint{0.932361in}{0.533259in}}%
\pgfpathcurveto{\pgfqpoint{0.927425in}{0.538194in}}{\pgfqpoint{0.920730in}{0.540968in}}{\pgfqpoint{0.913750in}{0.540968in}}%
\pgfpathcurveto{\pgfqpoint{0.906770in}{0.540968in}}{\pgfqpoint{0.900075in}{0.538194in}}{\pgfqpoint{0.895139in}{0.533259in}}%
\pgfpathcurveto{\pgfqpoint{0.890204in}{0.528323in}}{\pgfqpoint{0.887430in}{0.521628in}}{\pgfqpoint{0.887430in}{0.514648in}}%
\pgfpathcurveto{\pgfqpoint{0.887430in}{0.507668in}}{\pgfqpoint{0.890204in}{0.500973in}}{\pgfqpoint{0.895139in}{0.496037in}}%
\pgfpathcurveto{\pgfqpoint{0.900075in}{0.491102in}}{\pgfqpoint{0.906770in}{0.488328in}}{\pgfqpoint{0.913750in}{0.488328in}}%
\pgfpathclose%
\pgfusepath{stroke,fill}%
\end{pgfscope}%
\begin{pgfscope}%
\pgfpathrectangle{\pgfqpoint{0.516000in}{0.420000in}}{\pgfqpoint{1.591000in}{1.050000in}} %
\pgfusepath{clip}%
\pgfsetbuttcap%
\pgfsetroundjoin%
\definecolor{currentfill}{rgb}{0.090196,0.745098,0.811765}%
\pgfsetfillcolor{currentfill}%
\pgfsetlinewidth{1.138252pt}%
\definecolor{currentstroke}{rgb}{0.090196,0.745098,0.811765}%
\pgfsetstrokecolor{currentstroke}%
\pgfsetdash{}{0pt}%
\pgfpathmoveto{\pgfqpoint{1.178917in}{0.578702in}}%
\pgfpathcurveto{\pgfqpoint{1.185897in}{0.578702in}}{\pgfqpoint{1.192592in}{0.581476in}}{\pgfqpoint{1.197527in}{0.586411in}}%
\pgfpathcurveto{\pgfqpoint{1.202463in}{0.591347in}}{\pgfqpoint{1.205236in}{0.598042in}}{\pgfqpoint{1.205236in}{0.605022in}}%
\pgfpathcurveto{\pgfqpoint{1.205236in}{0.612002in}}{\pgfqpoint{1.202463in}{0.618697in}}{\pgfqpoint{1.197527in}{0.623633in}}%
\pgfpathcurveto{\pgfqpoint{1.192592in}{0.628568in}}{\pgfqpoint{1.185897in}{0.631342in}}{\pgfqpoint{1.178917in}{0.631342in}}%
\pgfpathcurveto{\pgfqpoint{1.171937in}{0.631342in}}{\pgfqpoint{1.165242in}{0.628568in}}{\pgfqpoint{1.160306in}{0.623633in}}%
\pgfpathcurveto{\pgfqpoint{1.155370in}{0.618697in}}{\pgfqpoint{1.152597in}{0.612002in}}{\pgfqpoint{1.152597in}{0.605022in}}%
\pgfpathcurveto{\pgfqpoint{1.152597in}{0.598042in}}{\pgfqpoint{1.155370in}{0.591347in}}{\pgfqpoint{1.160306in}{0.586411in}}%
\pgfpathcurveto{\pgfqpoint{1.165242in}{0.581476in}}{\pgfqpoint{1.171937in}{0.578702in}}{\pgfqpoint{1.178917in}{0.578702in}}%
\pgfpathclose%
\pgfusepath{stroke,fill}%
\end{pgfscope}%
\begin{pgfscope}%
\pgfpathrectangle{\pgfqpoint{0.516000in}{0.420000in}}{\pgfqpoint{1.591000in}{1.050000in}} %
\pgfusepath{clip}%
\pgfsetbuttcap%
\pgfsetroundjoin%
\definecolor{currentfill}{rgb}{0.090196,0.745098,0.811765}%
\pgfsetfillcolor{currentfill}%
\pgfsetlinewidth{1.138252pt}%
\definecolor{currentstroke}{rgb}{0.090196,0.745098,0.811765}%
\pgfsetstrokecolor{currentstroke}%
\pgfsetdash{}{0pt}%
\pgfpathmoveto{\pgfqpoint{1.444083in}{0.653209in}}%
\pgfpathcurveto{\pgfqpoint{1.451063in}{0.653209in}}{\pgfqpoint{1.457758in}{0.655982in}}{\pgfqpoint{1.462694in}{0.660918in}}%
\pgfpathcurveto{\pgfqpoint{1.467630in}{0.665853in}}{\pgfqpoint{1.470403in}{0.672549in}}{\pgfqpoint{1.470403in}{0.679529in}}%
\pgfpathcurveto{\pgfqpoint{1.470403in}{0.686509in}}{\pgfqpoint{1.467630in}{0.693204in}}{\pgfqpoint{1.462694in}{0.698139in}}%
\pgfpathcurveto{\pgfqpoint{1.457758in}{0.703075in}}{\pgfqpoint{1.451063in}{0.705848in}}{\pgfqpoint{1.444083in}{0.705848in}}%
\pgfpathcurveto{\pgfqpoint{1.437103in}{0.705848in}}{\pgfqpoint{1.430408in}{0.703075in}}{\pgfqpoint{1.425473in}{0.698139in}}%
\pgfpathcurveto{\pgfqpoint{1.420537in}{0.693204in}}{\pgfqpoint{1.417764in}{0.686509in}}{\pgfqpoint{1.417764in}{0.679529in}}%
\pgfpathcurveto{\pgfqpoint{1.417764in}{0.672549in}}{\pgfqpoint{1.420537in}{0.665853in}}{\pgfqpoint{1.425473in}{0.660918in}}%
\pgfpathcurveto{\pgfqpoint{1.430408in}{0.655982in}}{\pgfqpoint{1.437103in}{0.653209in}}{\pgfqpoint{1.444083in}{0.653209in}}%
\pgfpathclose%
\pgfusepath{stroke,fill}%
\end{pgfscope}%
\begin{pgfscope}%
\pgfpathrectangle{\pgfqpoint{0.516000in}{0.420000in}}{\pgfqpoint{1.591000in}{1.050000in}} %
\pgfusepath{clip}%
\pgfsetbuttcap%
\pgfsetroundjoin%
\definecolor{currentfill}{rgb}{0.090196,0.745098,0.811765}%
\pgfsetfillcolor{currentfill}%
\pgfsetlinewidth{1.138252pt}%
\definecolor{currentstroke}{rgb}{0.090196,0.745098,0.811765}%
\pgfsetstrokecolor{currentstroke}%
\pgfsetdash{}{0pt}%
\pgfpathmoveto{\pgfqpoint{1.709250in}{0.754933in}}%
\pgfpathcurveto{\pgfqpoint{1.716230in}{0.754933in}}{\pgfqpoint{1.722925in}{0.757706in}}{\pgfqpoint{1.727861in}{0.762641in}}%
\pgfpathcurveto{\pgfqpoint{1.732796in}{0.767577in}}{\pgfqpoint{1.735570in}{0.774272in}}{\pgfqpoint{1.735570in}{0.781252in}}%
\pgfpathcurveto{\pgfqpoint{1.735570in}{0.788232in}}{\pgfqpoint{1.732796in}{0.794927in}}{\pgfqpoint{1.727861in}{0.799863in}}%
\pgfpathcurveto{\pgfqpoint{1.722925in}{0.804799in}}{\pgfqpoint{1.716230in}{0.807572in}}{\pgfqpoint{1.709250in}{0.807572in}}%
\pgfpathcurveto{\pgfqpoint{1.702270in}{0.807572in}}{\pgfqpoint{1.695575in}{0.804799in}}{\pgfqpoint{1.690639in}{0.799863in}}%
\pgfpathcurveto{\pgfqpoint{1.685704in}{0.794927in}}{\pgfqpoint{1.682930in}{0.788232in}}{\pgfqpoint{1.682930in}{0.781252in}}%
\pgfpathcurveto{\pgfqpoint{1.682930in}{0.774272in}}{\pgfqpoint{1.685704in}{0.767577in}}{\pgfqpoint{1.690639in}{0.762641in}}%
\pgfpathcurveto{\pgfqpoint{1.695575in}{0.757706in}}{\pgfqpoint{1.702270in}{0.754933in}}{\pgfqpoint{1.709250in}{0.754933in}}%
\pgfpathclose%
\pgfusepath{stroke,fill}%
\end{pgfscope}%
\begin{pgfscope}%
\pgfpathrectangle{\pgfqpoint{0.516000in}{0.420000in}}{\pgfqpoint{1.591000in}{1.050000in}} %
\pgfusepath{clip}%
\pgfsetbuttcap%
\pgfsetroundjoin%
\definecolor{currentfill}{rgb}{0.090196,0.745098,0.811765}%
\pgfsetfillcolor{currentfill}%
\pgfsetlinewidth{1.138252pt}%
\definecolor{currentstroke}{rgb}{0.090196,0.745098,0.811765}%
\pgfsetstrokecolor{currentstroke}%
\pgfsetdash{}{0pt}%
\pgfpathmoveto{\pgfqpoint{1.974417in}{0.844602in}}%
\pgfpathcurveto{\pgfqpoint{1.981397in}{0.844602in}}{\pgfqpoint{1.988092in}{0.847375in}}{\pgfqpoint{1.993027in}{0.852311in}}%
\pgfpathcurveto{\pgfqpoint{1.997963in}{0.857246in}}{\pgfqpoint{2.000736in}{0.863941in}}{\pgfqpoint{2.000736in}{0.870922in}}%
\pgfpathcurveto{\pgfqpoint{2.000736in}{0.877902in}}{\pgfqpoint{1.997963in}{0.884597in}}{\pgfqpoint{1.993027in}{0.889532in}}%
\pgfpathcurveto{\pgfqpoint{1.988092in}{0.894468in}}{\pgfqpoint{1.981397in}{0.897241in}}{\pgfqpoint{1.974417in}{0.897241in}}%
\pgfpathcurveto{\pgfqpoint{1.967437in}{0.897241in}}{\pgfqpoint{1.960742in}{0.894468in}}{\pgfqpoint{1.955806in}{0.889532in}}%
\pgfpathcurveto{\pgfqpoint{1.950870in}{0.884597in}}{\pgfqpoint{1.948097in}{0.877902in}}{\pgfqpoint{1.948097in}{0.870922in}}%
\pgfpathcurveto{\pgfqpoint{1.948097in}{0.863941in}}{\pgfqpoint{1.950870in}{0.857246in}}{\pgfqpoint{1.955806in}{0.852311in}}%
\pgfpathcurveto{\pgfqpoint{1.960742in}{0.847375in}}{\pgfqpoint{1.967437in}{0.844602in}}{\pgfqpoint{1.974417in}{0.844602in}}%
\pgfpathclose%
\pgfusepath{stroke,fill}%
\end{pgfscope}%
\begin{pgfscope}%
\pgfpathrectangle{\pgfqpoint{0.516000in}{0.420000in}}{\pgfqpoint{1.591000in}{1.050000in}} %
\pgfusepath{clip}%
\pgfsetroundcap%
\pgfsetroundjoin%
\pgfsetlinewidth{1.517670pt}%
\definecolor{currentstroke}{rgb}{0.090196,0.745098,0.811765}%
\pgfsetstrokecolor{currentstroke}%
\pgfsetdash{}{0pt}%
\pgfpathmoveto{\pgfqpoint{0.648583in}{0.469498in}}%
\pgfpathlineto{\pgfqpoint{0.913750in}{0.514648in}}%
\pgfpathlineto{\pgfqpoint{1.178917in}{0.605022in}}%
\pgfpathlineto{\pgfqpoint{1.444083in}{0.679529in}}%
\pgfpathlineto{\pgfqpoint{1.709250in}{0.781252in}}%
\pgfpathlineto{\pgfqpoint{1.974417in}{0.870922in}}%
\pgfusepath{stroke}%
\end{pgfscope}%
\begin{pgfscope}%
\pgfpathrectangle{\pgfqpoint{0.516000in}{0.420000in}}{\pgfqpoint{1.591000in}{1.050000in}} %
\pgfusepath{clip}%
\pgfsetroundcap%
\pgfsetroundjoin%
\pgfsetlinewidth{2.168100pt}%
\definecolor{currentstroke}{rgb}{0.090196,0.745098,0.811765}%
\pgfsetstrokecolor{currentstroke}%
\pgfsetdash{}{0pt}%
\pgfpathmoveto{\pgfqpoint{0.648583in}{0.464729in}}%
\pgfpathlineto{\pgfqpoint{0.648583in}{0.472509in}}%
\pgfusepath{stroke}%
\end{pgfscope}%
\begin{pgfscope}%
\pgfpathrectangle{\pgfqpoint{0.516000in}{0.420000in}}{\pgfqpoint{1.591000in}{1.050000in}} %
\pgfusepath{clip}%
\pgfsetroundcap%
\pgfsetroundjoin%
\pgfsetlinewidth{2.168100pt}%
\definecolor{currentstroke}{rgb}{0.090196,0.745098,0.811765}%
\pgfsetstrokecolor{currentstroke}%
\pgfsetdash{}{0pt}%
\pgfpathmoveto{\pgfqpoint{0.913750in}{0.509586in}}%
\pgfpathlineto{\pgfqpoint{0.913750in}{0.519430in}}%
\pgfusepath{stroke}%
\end{pgfscope}%
\begin{pgfscope}%
\pgfpathrectangle{\pgfqpoint{0.516000in}{0.420000in}}{\pgfqpoint{1.591000in}{1.050000in}} %
\pgfusepath{clip}%
\pgfsetroundcap%
\pgfsetroundjoin%
\pgfsetlinewidth{2.168100pt}%
\definecolor{currentstroke}{rgb}{0.090196,0.745098,0.811765}%
\pgfsetstrokecolor{currentstroke}%
\pgfsetdash{}{0pt}%
\pgfpathmoveto{\pgfqpoint{1.178917in}{0.592851in}}%
\pgfpathlineto{\pgfqpoint{1.178917in}{0.613816in}}%
\pgfusepath{stroke}%
\end{pgfscope}%
\begin{pgfscope}%
\pgfpathrectangle{\pgfqpoint{0.516000in}{0.420000in}}{\pgfqpoint{1.591000in}{1.050000in}} %
\pgfusepath{clip}%
\pgfsetroundcap%
\pgfsetroundjoin%
\pgfsetlinewidth{2.168100pt}%
\definecolor{currentstroke}{rgb}{0.090196,0.745098,0.811765}%
\pgfsetstrokecolor{currentstroke}%
\pgfsetdash{}{0pt}%
\pgfpathmoveto{\pgfqpoint{1.444083in}{0.668160in}}%
\pgfpathlineto{\pgfqpoint{1.444083in}{0.696567in}}%
\pgfusepath{stroke}%
\end{pgfscope}%
\begin{pgfscope}%
\pgfpathrectangle{\pgfqpoint{0.516000in}{0.420000in}}{\pgfqpoint{1.591000in}{1.050000in}} %
\pgfusepath{clip}%
\pgfsetroundcap%
\pgfsetroundjoin%
\pgfsetlinewidth{2.168100pt}%
\definecolor{currentstroke}{rgb}{0.090196,0.745098,0.811765}%
\pgfsetstrokecolor{currentstroke}%
\pgfsetdash{}{0pt}%
\pgfpathmoveto{\pgfqpoint{1.709250in}{0.766893in}}%
\pgfpathlineto{\pgfqpoint{1.709250in}{0.790067in}}%
\pgfusepath{stroke}%
\end{pgfscope}%
\begin{pgfscope}%
\pgfpathrectangle{\pgfqpoint{0.516000in}{0.420000in}}{\pgfqpoint{1.591000in}{1.050000in}} %
\pgfusepath{clip}%
\pgfsetroundcap%
\pgfsetroundjoin%
\pgfsetlinewidth{2.168100pt}%
\definecolor{currentstroke}{rgb}{0.090196,0.745098,0.811765}%
\pgfsetstrokecolor{currentstroke}%
\pgfsetdash{}{0pt}%
\pgfpathmoveto{\pgfqpoint{1.974417in}{0.862641in}}%
\pgfpathlineto{\pgfqpoint{1.974417in}{0.879185in}}%
\pgfusepath{stroke}%
\end{pgfscope}%
\begin{pgfscope}%
\pgfsetrectcap%
\pgfsetmiterjoin%
\pgfsetlinewidth{1.003750pt}%
\definecolor{currentstroke}{rgb}{0.800000,0.800000,0.800000}%
\pgfsetstrokecolor{currentstroke}%
\pgfsetdash{}{0pt}%
\pgfpathmoveto{\pgfqpoint{0.516000in}{0.420000in}}%
\pgfpathlineto{\pgfqpoint{0.516000in}{1.470000in}}%
\pgfusepath{stroke}%
\end{pgfscope}%
\begin{pgfscope}%
\pgfsetrectcap%
\pgfsetmiterjoin%
\pgfsetlinewidth{1.003750pt}%
\definecolor{currentstroke}{rgb}{0.800000,0.800000,0.800000}%
\pgfsetstrokecolor{currentstroke}%
\pgfsetdash{}{0pt}%
\pgfpathmoveto{\pgfqpoint{2.107000in}{0.420000in}}%
\pgfpathlineto{\pgfqpoint{2.107000in}{1.470000in}}%
\pgfusepath{stroke}%
\end{pgfscope}%
\begin{pgfscope}%
\pgfsetrectcap%
\pgfsetmiterjoin%
\pgfsetlinewidth{1.003750pt}%
\definecolor{currentstroke}{rgb}{0.800000,0.800000,0.800000}%
\pgfsetstrokecolor{currentstroke}%
\pgfsetdash{}{0pt}%
\pgfpathmoveto{\pgfqpoint{0.516000in}{0.420000in}}%
\pgfpathlineto{\pgfqpoint{2.107000in}{0.420000in}}%
\pgfusepath{stroke}%
\end{pgfscope}%
\begin{pgfscope}%
\pgfsetrectcap%
\pgfsetmiterjoin%
\pgfsetlinewidth{1.003750pt}%
\definecolor{currentstroke}{rgb}{0.800000,0.800000,0.800000}%
\pgfsetstrokecolor{currentstroke}%
\pgfsetdash{}{0pt}%
\pgfpathmoveto{\pgfqpoint{0.516000in}{1.470000in}}%
\pgfpathlineto{\pgfqpoint{2.107000in}{1.470000in}}%
\pgfusepath{stroke}%
\end{pgfscope}%
\begin{pgfscope}%
\pgfpathrectangle{\pgfqpoint{0.516000in}{0.420000in}}{\pgfqpoint{1.591000in}{1.050000in}} %
\pgfusepath{clip}%
\pgfsetbuttcap%
\pgfsetroundjoin%
\definecolor{currentfill}{rgb}{0.498039,0.498039,0.498039}%
\pgfsetfillcolor{currentfill}%
\pgfsetlinewidth{1.138252pt}%
\definecolor{currentstroke}{rgb}{0.498039,0.498039,0.498039}%
\pgfsetstrokecolor{currentstroke}%
\pgfsetdash{}{0pt}%
\pgfpathmoveto{\pgfqpoint{0.648583in}{0.429008in}}%
\pgfpathcurveto{\pgfqpoint{0.655563in}{0.429008in}}{\pgfqpoint{0.662258in}{0.431781in}}{\pgfqpoint{0.667194in}{0.436717in}}%
\pgfpathcurveto{\pgfqpoint{0.672130in}{0.441652in}}{\pgfqpoint{0.674903in}{0.448347in}}{\pgfqpoint{0.674903in}{0.455327in}}%
\pgfpathcurveto{\pgfqpoint{0.674903in}{0.462307in}}{\pgfqpoint{0.672130in}{0.469002in}}{\pgfqpoint{0.667194in}{0.473938in}}%
\pgfpathcurveto{\pgfqpoint{0.662258in}{0.478874in}}{\pgfqpoint{0.655563in}{0.481647in}}{\pgfqpoint{0.648583in}{0.481647in}}%
\pgfpathcurveto{\pgfqpoint{0.641603in}{0.481647in}}{\pgfqpoint{0.634908in}{0.478874in}}{\pgfqpoint{0.629973in}{0.473938in}}%
\pgfpathcurveto{\pgfqpoint{0.625037in}{0.469002in}}{\pgfqpoint{0.622264in}{0.462307in}}{\pgfqpoint{0.622264in}{0.455327in}}%
\pgfpathcurveto{\pgfqpoint{0.622264in}{0.448347in}}{\pgfqpoint{0.625037in}{0.441652in}}{\pgfqpoint{0.629973in}{0.436717in}}%
\pgfpathcurveto{\pgfqpoint{0.634908in}{0.431781in}}{\pgfqpoint{0.641603in}{0.429008in}}{\pgfqpoint{0.648583in}{0.429008in}}%
\pgfpathclose%
\pgfusepath{stroke,fill}%
\end{pgfscope}%
\begin{pgfscope}%
\pgfpathrectangle{\pgfqpoint{0.516000in}{0.420000in}}{\pgfqpoint{1.591000in}{1.050000in}} %
\pgfusepath{clip}%
\pgfsetbuttcap%
\pgfsetroundjoin%
\definecolor{currentfill}{rgb}{0.498039,0.498039,0.498039}%
\pgfsetfillcolor{currentfill}%
\pgfsetlinewidth{1.138252pt}%
\definecolor{currentstroke}{rgb}{0.498039,0.498039,0.498039}%
\pgfsetstrokecolor{currentstroke}%
\pgfsetdash{}{0pt}%
\pgfpathmoveto{\pgfqpoint{0.913750in}{0.445743in}}%
\pgfpathcurveto{\pgfqpoint{0.920730in}{0.445743in}}{\pgfqpoint{0.927425in}{0.448517in}}{\pgfqpoint{0.932361in}{0.453452in}}%
\pgfpathcurveto{\pgfqpoint{0.937296in}{0.458388in}}{\pgfqpoint{0.940070in}{0.465083in}}{\pgfqpoint{0.940070in}{0.472063in}}%
\pgfpathcurveto{\pgfqpoint{0.940070in}{0.479043in}}{\pgfqpoint{0.937296in}{0.485738in}}{\pgfqpoint{0.932361in}{0.490674in}}%
\pgfpathcurveto{\pgfqpoint{0.927425in}{0.495609in}}{\pgfqpoint{0.920730in}{0.498383in}}{\pgfqpoint{0.913750in}{0.498383in}}%
\pgfpathcurveto{\pgfqpoint{0.906770in}{0.498383in}}{\pgfqpoint{0.900075in}{0.495609in}}{\pgfqpoint{0.895139in}{0.490674in}}%
\pgfpathcurveto{\pgfqpoint{0.890204in}{0.485738in}}{\pgfqpoint{0.887430in}{0.479043in}}{\pgfqpoint{0.887430in}{0.472063in}}%
\pgfpathcurveto{\pgfqpoint{0.887430in}{0.465083in}}{\pgfqpoint{0.890204in}{0.458388in}}{\pgfqpoint{0.895139in}{0.453452in}}%
\pgfpathcurveto{\pgfqpoint{0.900075in}{0.448517in}}{\pgfqpoint{0.906770in}{0.445743in}}{\pgfqpoint{0.913750in}{0.445743in}}%
\pgfpathclose%
\pgfusepath{stroke,fill}%
\end{pgfscope}%
\begin{pgfscope}%
\pgfpathrectangle{\pgfqpoint{0.516000in}{0.420000in}}{\pgfqpoint{1.591000in}{1.050000in}} %
\pgfusepath{clip}%
\pgfsetbuttcap%
\pgfsetroundjoin%
\definecolor{currentfill}{rgb}{0.498039,0.498039,0.498039}%
\pgfsetfillcolor{currentfill}%
\pgfsetlinewidth{1.138252pt}%
\definecolor{currentstroke}{rgb}{0.498039,0.498039,0.498039}%
\pgfsetstrokecolor{currentstroke}%
\pgfsetdash{}{0pt}%
\pgfpathmoveto{\pgfqpoint{1.178917in}{0.472301in}}%
\pgfpathcurveto{\pgfqpoint{1.185897in}{0.472301in}}{\pgfqpoint{1.192592in}{0.475074in}}{\pgfqpoint{1.197527in}{0.480010in}}%
\pgfpathcurveto{\pgfqpoint{1.202463in}{0.484945in}}{\pgfqpoint{1.205236in}{0.491640in}}{\pgfqpoint{1.205236in}{0.498620in}}%
\pgfpathcurveto{\pgfqpoint{1.205236in}{0.505600in}}{\pgfqpoint{1.202463in}{0.512296in}}{\pgfqpoint{1.197527in}{0.517231in}}%
\pgfpathcurveto{\pgfqpoint{1.192592in}{0.522167in}}{\pgfqpoint{1.185897in}{0.524940in}}{\pgfqpoint{1.178917in}{0.524940in}}%
\pgfpathcurveto{\pgfqpoint{1.171937in}{0.524940in}}{\pgfqpoint{1.165242in}{0.522167in}}{\pgfqpoint{1.160306in}{0.517231in}}%
\pgfpathcurveto{\pgfqpoint{1.155370in}{0.512296in}}{\pgfqpoint{1.152597in}{0.505600in}}{\pgfqpoint{1.152597in}{0.498620in}}%
\pgfpathcurveto{\pgfqpoint{1.152597in}{0.491640in}}{\pgfqpoint{1.155370in}{0.484945in}}{\pgfqpoint{1.160306in}{0.480010in}}%
\pgfpathcurveto{\pgfqpoint{1.165242in}{0.475074in}}{\pgfqpoint{1.171937in}{0.472301in}}{\pgfqpoint{1.178917in}{0.472301in}}%
\pgfpathclose%
\pgfusepath{stroke,fill}%
\end{pgfscope}%
\begin{pgfscope}%
\pgfpathrectangle{\pgfqpoint{0.516000in}{0.420000in}}{\pgfqpoint{1.591000in}{1.050000in}} %
\pgfusepath{clip}%
\pgfsetbuttcap%
\pgfsetroundjoin%
\definecolor{currentfill}{rgb}{0.498039,0.498039,0.498039}%
\pgfsetfillcolor{currentfill}%
\pgfsetlinewidth{1.138252pt}%
\definecolor{currentstroke}{rgb}{0.498039,0.498039,0.498039}%
\pgfsetstrokecolor{currentstroke}%
\pgfsetdash{}{0pt}%
\pgfpathmoveto{\pgfqpoint{1.444083in}{0.498228in}}%
\pgfpathcurveto{\pgfqpoint{1.451063in}{0.498228in}}{\pgfqpoint{1.457758in}{0.501001in}}{\pgfqpoint{1.462694in}{0.505937in}}%
\pgfpathcurveto{\pgfqpoint{1.467630in}{0.510872in}}{\pgfqpoint{1.470403in}{0.517567in}}{\pgfqpoint{1.470403in}{0.524547in}}%
\pgfpathcurveto{\pgfqpoint{1.470403in}{0.531527in}}{\pgfqpoint{1.467630in}{0.538223in}}{\pgfqpoint{1.462694in}{0.543158in}}%
\pgfpathcurveto{\pgfqpoint{1.457758in}{0.548094in}}{\pgfqpoint{1.451063in}{0.550867in}}{\pgfqpoint{1.444083in}{0.550867in}}%
\pgfpathcurveto{\pgfqpoint{1.437103in}{0.550867in}}{\pgfqpoint{1.430408in}{0.548094in}}{\pgfqpoint{1.425473in}{0.543158in}}%
\pgfpathcurveto{\pgfqpoint{1.420537in}{0.538223in}}{\pgfqpoint{1.417764in}{0.531527in}}{\pgfqpoint{1.417764in}{0.524547in}}%
\pgfpathcurveto{\pgfqpoint{1.417764in}{0.517567in}}{\pgfqpoint{1.420537in}{0.510872in}}{\pgfqpoint{1.425473in}{0.505937in}}%
\pgfpathcurveto{\pgfqpoint{1.430408in}{0.501001in}}{\pgfqpoint{1.437103in}{0.498228in}}{\pgfqpoint{1.444083in}{0.498228in}}%
\pgfpathclose%
\pgfusepath{stroke,fill}%
\end{pgfscope}%
\begin{pgfscope}%
\pgfpathrectangle{\pgfqpoint{0.516000in}{0.420000in}}{\pgfqpoint{1.591000in}{1.050000in}} %
\pgfusepath{clip}%
\pgfsetbuttcap%
\pgfsetroundjoin%
\definecolor{currentfill}{rgb}{0.498039,0.498039,0.498039}%
\pgfsetfillcolor{currentfill}%
\pgfsetlinewidth{1.138252pt}%
\definecolor{currentstroke}{rgb}{0.498039,0.498039,0.498039}%
\pgfsetstrokecolor{currentstroke}%
\pgfsetdash{}{0pt}%
\pgfpathmoveto{\pgfqpoint{1.709250in}{0.513604in}}%
\pgfpathcurveto{\pgfqpoint{1.716230in}{0.513604in}}{\pgfqpoint{1.722925in}{0.516378in}}{\pgfqpoint{1.727861in}{0.521313in}}%
\pgfpathcurveto{\pgfqpoint{1.732796in}{0.526249in}}{\pgfqpoint{1.735570in}{0.532944in}}{\pgfqpoint{1.735570in}{0.539924in}}%
\pgfpathcurveto{\pgfqpoint{1.735570in}{0.546904in}}{\pgfqpoint{1.732796in}{0.553599in}}{\pgfqpoint{1.727861in}{0.558535in}}%
\pgfpathcurveto{\pgfqpoint{1.722925in}{0.563470in}}{\pgfqpoint{1.716230in}{0.566244in}}{\pgfqpoint{1.709250in}{0.566244in}}%
\pgfpathcurveto{\pgfqpoint{1.702270in}{0.566244in}}{\pgfqpoint{1.695575in}{0.563470in}}{\pgfqpoint{1.690639in}{0.558535in}}%
\pgfpathcurveto{\pgfqpoint{1.685704in}{0.553599in}}{\pgfqpoint{1.682930in}{0.546904in}}{\pgfqpoint{1.682930in}{0.539924in}}%
\pgfpathcurveto{\pgfqpoint{1.682930in}{0.532944in}}{\pgfqpoint{1.685704in}{0.526249in}}{\pgfqpoint{1.690639in}{0.521313in}}%
\pgfpathcurveto{\pgfqpoint{1.695575in}{0.516378in}}{\pgfqpoint{1.702270in}{0.513604in}}{\pgfqpoint{1.709250in}{0.513604in}}%
\pgfpathclose%
\pgfusepath{stroke,fill}%
\end{pgfscope}%
\begin{pgfscope}%
\pgfpathrectangle{\pgfqpoint{0.516000in}{0.420000in}}{\pgfqpoint{1.591000in}{1.050000in}} %
\pgfusepath{clip}%
\pgfsetbuttcap%
\pgfsetroundjoin%
\definecolor{currentfill}{rgb}{0.498039,0.498039,0.498039}%
\pgfsetfillcolor{currentfill}%
\pgfsetlinewidth{1.138252pt}%
\definecolor{currentstroke}{rgb}{0.498039,0.498039,0.498039}%
\pgfsetstrokecolor{currentstroke}%
\pgfsetdash{}{0pt}%
\pgfpathmoveto{\pgfqpoint{1.974417in}{0.524656in}}%
\pgfpathcurveto{\pgfqpoint{1.981397in}{0.524656in}}{\pgfqpoint{1.988092in}{0.527430in}}{\pgfqpoint{1.993027in}{0.532365in}}%
\pgfpathcurveto{\pgfqpoint{1.997963in}{0.537301in}}{\pgfqpoint{2.000736in}{0.543996in}}{\pgfqpoint{2.000736in}{0.550976in}}%
\pgfpathcurveto{\pgfqpoint{2.000736in}{0.557956in}}{\pgfqpoint{1.997963in}{0.564651in}}{\pgfqpoint{1.993027in}{0.569587in}}%
\pgfpathcurveto{\pgfqpoint{1.988092in}{0.574522in}}{\pgfqpoint{1.981397in}{0.577296in}}{\pgfqpoint{1.974417in}{0.577296in}}%
\pgfpathcurveto{\pgfqpoint{1.967437in}{0.577296in}}{\pgfqpoint{1.960742in}{0.574522in}}{\pgfqpoint{1.955806in}{0.569587in}}%
\pgfpathcurveto{\pgfqpoint{1.950870in}{0.564651in}}{\pgfqpoint{1.948097in}{0.557956in}}{\pgfqpoint{1.948097in}{0.550976in}}%
\pgfpathcurveto{\pgfqpoint{1.948097in}{0.543996in}}{\pgfqpoint{1.950870in}{0.537301in}}{\pgfqpoint{1.955806in}{0.532365in}}%
\pgfpathcurveto{\pgfqpoint{1.960742in}{0.527430in}}{\pgfqpoint{1.967437in}{0.524656in}}{\pgfqpoint{1.974417in}{0.524656in}}%
\pgfpathclose%
\pgfusepath{stroke,fill}%
\end{pgfscope}%
\begin{pgfscope}%
\pgfpathrectangle{\pgfqpoint{0.516000in}{0.420000in}}{\pgfqpoint{1.591000in}{1.050000in}} %
\pgfusepath{clip}%
\pgfsetroundcap%
\pgfsetroundjoin%
\pgfsetlinewidth{1.517670pt}%
\definecolor{currentstroke}{rgb}{0.498039,0.498039,0.498039}%
\pgfsetstrokecolor{currentstroke}%
\pgfsetdash{}{0pt}%
\pgfpathmoveto{\pgfqpoint{0.648583in}{0.455327in}}%
\pgfpathlineto{\pgfqpoint{0.913750in}{0.472063in}}%
\pgfpathlineto{\pgfqpoint{1.178917in}{0.498620in}}%
\pgfpathlineto{\pgfqpoint{1.444083in}{0.524547in}}%
\pgfpathlineto{\pgfqpoint{1.709250in}{0.539924in}}%
\pgfpathlineto{\pgfqpoint{1.974417in}{0.550976in}}%
\pgfusepath{stroke}%
\end{pgfscope}%
\begin{pgfscope}%
\pgfpathrectangle{\pgfqpoint{0.516000in}{0.420000in}}{\pgfqpoint{1.591000in}{1.050000in}} %
\pgfusepath{clip}%
\pgfsetroundcap%
\pgfsetroundjoin%
\pgfsetlinewidth{2.168100pt}%
\definecolor{currentstroke}{rgb}{0.498039,0.498039,0.498039}%
\pgfsetstrokecolor{currentstroke}%
\pgfsetdash{}{0pt}%
\pgfpathmoveto{\pgfqpoint{0.648583in}{0.453911in}}%
\pgfpathlineto{\pgfqpoint{0.648583in}{0.456422in}}%
\pgfusepath{stroke}%
\end{pgfscope}%
\begin{pgfscope}%
\pgfpathrectangle{\pgfqpoint{0.516000in}{0.420000in}}{\pgfqpoint{1.591000in}{1.050000in}} %
\pgfusepath{clip}%
\pgfsetroundcap%
\pgfsetroundjoin%
\pgfsetlinewidth{2.168100pt}%
\definecolor{currentstroke}{rgb}{0.498039,0.498039,0.498039}%
\pgfsetstrokecolor{currentstroke}%
\pgfsetdash{}{0pt}%
\pgfpathmoveto{\pgfqpoint{0.913750in}{0.471492in}}%
\pgfpathlineto{\pgfqpoint{0.913750in}{0.475257in}}%
\pgfusepath{stroke}%
\end{pgfscope}%
\begin{pgfscope}%
\pgfpathrectangle{\pgfqpoint{0.516000in}{0.420000in}}{\pgfqpoint{1.591000in}{1.050000in}} %
\pgfusepath{clip}%
\pgfsetroundcap%
\pgfsetroundjoin%
\pgfsetlinewidth{2.168100pt}%
\definecolor{currentstroke}{rgb}{0.498039,0.498039,0.498039}%
\pgfsetstrokecolor{currentstroke}%
\pgfsetdash{}{0pt}%
\pgfpathmoveto{\pgfqpoint{1.178917in}{0.494039in}}%
\pgfpathlineto{\pgfqpoint{1.178917in}{0.502080in}}%
\pgfusepath{stroke}%
\end{pgfscope}%
\begin{pgfscope}%
\pgfpathrectangle{\pgfqpoint{0.516000in}{0.420000in}}{\pgfqpoint{1.591000in}{1.050000in}} %
\pgfusepath{clip}%
\pgfsetroundcap%
\pgfsetroundjoin%
\pgfsetlinewidth{2.168100pt}%
\definecolor{currentstroke}{rgb}{0.498039,0.498039,0.498039}%
\pgfsetstrokecolor{currentstroke}%
\pgfsetdash{}{0pt}%
\pgfpathmoveto{\pgfqpoint{1.444083in}{0.518538in}}%
\pgfpathlineto{\pgfqpoint{1.444083in}{0.529538in}}%
\pgfusepath{stroke}%
\end{pgfscope}%
\begin{pgfscope}%
\pgfpathrectangle{\pgfqpoint{0.516000in}{0.420000in}}{\pgfqpoint{1.591000in}{1.050000in}} %
\pgfusepath{clip}%
\pgfsetroundcap%
\pgfsetroundjoin%
\pgfsetlinewidth{2.168100pt}%
\definecolor{currentstroke}{rgb}{0.498039,0.498039,0.498039}%
\pgfsetstrokecolor{currentstroke}%
\pgfsetdash{}{0pt}%
\pgfpathmoveto{\pgfqpoint{1.709250in}{0.531304in}}%
\pgfpathlineto{\pgfqpoint{1.709250in}{0.541245in}}%
\pgfusepath{stroke}%
\end{pgfscope}%
\begin{pgfscope}%
\pgfpathrectangle{\pgfqpoint{0.516000in}{0.420000in}}{\pgfqpoint{1.591000in}{1.050000in}} %
\pgfusepath{clip}%
\pgfsetroundcap%
\pgfsetroundjoin%
\pgfsetlinewidth{2.168100pt}%
\definecolor{currentstroke}{rgb}{0.498039,0.498039,0.498039}%
\pgfsetstrokecolor{currentstroke}%
\pgfsetdash{}{0pt}%
\pgfpathmoveto{\pgfqpoint{1.974417in}{0.546191in}}%
\pgfpathlineto{\pgfqpoint{1.974417in}{0.553954in}}%
\pgfusepath{stroke}%
\end{pgfscope}%
\begin{pgfscope}%
\pgfpathrectangle{\pgfqpoint{0.516000in}{0.420000in}}{\pgfqpoint{1.591000in}{1.050000in}} %
\pgfusepath{clip}%
\pgfsetbuttcap%
\pgfsetroundjoin%
\pgfsetlinewidth{2.007500pt}%
\definecolor{currentstroke}{rgb}{0.890196,0.466667,0.760784}%
\pgfsetstrokecolor{currentstroke}%
\pgfsetdash{{7.400000pt}{3.200000pt}}{0.000000pt}%
\pgfpathmoveto{\pgfqpoint{0.516000in}{1.290527in}}%
\pgfpathlineto{\pgfqpoint{2.107000in}{1.290527in}}%
\pgfusepath{stroke}%
\end{pgfscope}%
\end{pgfpicture}%
\makeatother%
\endgroup%

%% file: images/tables/OSGA_results.tex
\begin{tabular}{llcccccccc}
\hline
        & &  Chemical-I &      Sarcos &       Tower &     puma8NH \\
\hline
Original & {} &  .863 (.02) &  .949 (.00) &  .937 (.01) &  .684 (.00) \\
Sampling & 1 &  .002 (.00) &  .928 (.01) &  .390 (.31) &  .369 (.06) \\
        & 5 &  .431 (.36) &  .946 (.00) &  .907 (.02) &  .572 (.04) \\
        & 10 &  .396 (.36) &  .947 (.00) &  .919 (.02) &  .632 (.02) \\
        & 20 &  .728 (.21) &  .948 (.00) &  .925 (.01) &  .657 (.00) \\
        & 30 &  .787 (.08) &  .948 (.00) &  .933 (.00) &  .668 (.01) \\
        & 40 &  .836 (.01) &  .947 (.00) &  .934 (.01) &  .675 (.01) \\
        & 50 &  .834 (.02) &  .949 (.00) &  .936 (.00) &  .677 (.01) \\
K-Means & 1 &  .018 (.04) &  .943 (.01) &  .729 (.14) &  .480 (.11) \\
        & 5 &  .523 (.15) &  .948 (.00) &  .918 (.01) &  .596 (.07) \\
        & 10 &  .704 (.13) &  .948 (.00) &  .926 (.00) &  .596 (.09) \\
        & 20 &  .806 (.06) &  .948 (.00) &  .933 (.01) &  .656 (.05) \\
        & 30 &  .828 (.03) &  .948 (.00) &  .932 (.00) &  .665 (.01) \\
        & 40 &  .853 (.04) &  .948 (.00) &  .932 (.00) &  .672 (.01) \\
        & 50 &  .848 (.02) &  .947 (.00) &  .931 (.00) &  .680 (.00) \\
Binning & 1 &  .009 (.01) &  .504 (.45) &  .378 (.37) &  .021 (.03) \\
        & 5 &  .253 (.30) &  .866 (.01) &  .313 (.35) &  .189 (.31) \\
        & 10 &  .345 (.21) &  .893 (.01) &  .700 (.27) &  .532 (.05) \\
        & 20 &  .578 (.13) &  .892 (.01) &  .806 (.05) &  .530 (.07) \\
        & 30 &  .673 (.05) &  .899 (.01) &  .853 (.03) &  .576 (.05) \\
        & 40 &  .699 (.05) &  .898 (.01) &  .872 (.05) &  .630 (.03) \\
        & 50 &  .718 (.08) &  .897 (.01) &  .883 (.02) &  .648 (.01) \\
\hline
\end{tabular}

%% file: images/plots/legend_datasets_Sarcos.pgf
\begingroup%
\makeatletter%
\begin{pgfpicture}%
\pgfpathrectangle{\pgfpointorigin}{\pgfqpoint{4.500000in}{0.500000in}}%
\pgfusepath{use as bounding box, clip}%
\begin{pgfscope}%
\pgfsetbuttcap%
\pgfsetmiterjoin%
\pgfsetlinewidth{0.000000pt}%
\definecolor{currentstroke}{rgb}{0.000000,0.000000,0.000000}%
\pgfsetstrokecolor{currentstroke}%
\pgfsetstrokeopacity{0.000000}%
\pgfsetdash{}{0pt}%
\pgfpathmoveto{\pgfqpoint{0.000000in}{0.000000in}}%
\pgfpathlineto{\pgfqpoint{4.500000in}{0.000000in}}%
\pgfpathlineto{\pgfqpoint{4.500000in}{0.500000in}}%
\pgfpathlineto{\pgfqpoint{0.000000in}{0.500000in}}%
\pgfpathclose%
\pgfusepath{}%
\end{pgfscope}%
\begin{pgfscope}%
\pgfsetbuttcap%
\pgfsetroundjoin%
\definecolor{currentfill}{rgb}{0.121569,0.466667,0.705882}%
\pgfsetfillcolor{currentfill}%
\pgfsetlinewidth{1.138252pt}%
\definecolor{currentstroke}{rgb}{0.121569,0.466667,0.705882}%
\pgfsetstrokecolor{currentstroke}%
\pgfsetdash{}{0pt}%
\pgfpathmoveto{\pgfqpoint{0.804130in}{0.238056in}}%
\pgfpathlineto{\pgfqpoint{0.826463in}{0.275278in}}%
\pgfpathlineto{\pgfqpoint{0.804130in}{0.312499in}}%
\pgfpathlineto{\pgfqpoint{0.781797in}{0.275278in}}%
\pgfpathclose%
\pgfusepath{stroke,fill}%
\end{pgfscope}%
\begin{pgfscope}%
\definecolor{textcolor}{rgb}{0.150000,0.150000,0.150000}%
\pgfsetstrokecolor{textcolor}%
\pgfsetfillcolor{textcolor}%
\pgftext[x=0.926352in,y=0.243194in,left,base]{\color{textcolor}\fontsize{8.800000}{10.560000}\selectfont RF Test}%
\end{pgfscope}%
\begin{pgfscope}%
\pgfsetbuttcap%
\pgfsetroundjoin%
\definecolor{currentfill}{rgb}{0.631373,0.788235,0.956863}%
\pgfsetfillcolor{currentfill}%
\pgfsetlinewidth{0.813038pt}%
\definecolor{currentstroke}{rgb}{0.631373,0.788235,0.956863}%
\pgfsetstrokecolor{currentstroke}%
\pgfsetdash{}{0pt}%
\pgfpathmoveto{\pgfqpoint{0.804130in}{0.076469in}}%
\pgfpathlineto{\pgfqpoint{0.820082in}{0.103056in}}%
\pgfpathlineto{\pgfqpoint{0.804130in}{0.129642in}}%
\pgfpathlineto{\pgfqpoint{0.788178in}{0.103056in}}%
\pgfpathclose%
\pgfusepath{stroke,fill}%
\end{pgfscope}%
\begin{pgfscope}%
\definecolor{textcolor}{rgb}{0.150000,0.150000,0.150000}%
\pgfsetstrokecolor{textcolor}%
\pgfsetfillcolor{textcolor}%
\pgftext[x=0.926352in,y=0.070972in,left,base]{\color{textcolor}\fontsize{8.800000}{10.560000}\selectfont Training}%
\end{pgfscope}%
\begin{pgfscope}%
\pgfsetbuttcap%
\pgfsetroundjoin%
\definecolor{currentfill}{rgb}{1.000000,0.498039,0.054902}%
\pgfsetfillcolor{currentfill}%
\pgfsetlinewidth{1.138252pt}%
\definecolor{currentstroke}{rgb}{1.000000,0.498039,0.054902}%
\pgfsetstrokecolor{currentstroke}%
\pgfsetdash{}{0pt}%
\pgfpathmoveto{\pgfqpoint{2.073494in}{0.248958in}}%
\pgfpathcurveto{\pgfqpoint{2.080474in}{0.248958in}}{\pgfqpoint{2.087169in}{0.251731in}}{\pgfqpoint{2.092105in}{0.256667in}}%
\pgfpathcurveto{\pgfqpoint{2.097041in}{0.261603in}}{\pgfqpoint{2.099814in}{0.268298in}}{\pgfqpoint{2.099814in}{0.275278in}}%
\pgfpathcurveto{\pgfqpoint{2.099814in}{0.282258in}}{\pgfqpoint{2.097041in}{0.288953in}}{\pgfqpoint{2.092105in}{0.293889in}}%
\pgfpathcurveto{\pgfqpoint{2.087169in}{0.298824in}}{\pgfqpoint{2.080474in}{0.301597in}}{\pgfqpoint{2.073494in}{0.301597in}}%
\pgfpathcurveto{\pgfqpoint{2.066514in}{0.301597in}}{\pgfqpoint{2.059819in}{0.298824in}}{\pgfqpoint{2.054883in}{0.293889in}}%
\pgfpathcurveto{\pgfqpoint{2.049948in}{0.288953in}}{\pgfqpoint{2.047175in}{0.282258in}}{\pgfqpoint{2.047175in}{0.275278in}}%
\pgfpathcurveto{\pgfqpoint{2.047175in}{0.268298in}}{\pgfqpoint{2.049948in}{0.261603in}}{\pgfqpoint{2.054883in}{0.256667in}}%
\pgfpathcurveto{\pgfqpoint{2.059819in}{0.251731in}}{\pgfqpoint{2.066514in}{0.248958in}}{\pgfqpoint{2.073494in}{0.248958in}}%
\pgfpathclose%
\pgfusepath{stroke,fill}%
\end{pgfscope}%
\begin{pgfscope}%
\definecolor{textcolor}{rgb}{0.150000,0.150000,0.150000}%
\pgfsetstrokecolor{textcolor}%
\pgfsetfillcolor{textcolor}%
\pgftext[x=2.195716in,y=0.243194in,left,base]{\color{textcolor}\fontsize{8.800000}{10.560000}\selectfont GP Test}%
\end{pgfscope}%
\begin{pgfscope}%
\pgfsetbuttcap%
\pgfsetroundjoin%
\definecolor{currentfill}{rgb}{1.000000,0.705882,0.509804}%
\pgfsetfillcolor{currentfill}%
\pgfsetlinewidth{0.813038pt}%
\definecolor{currentstroke}{rgb}{1.000000,0.705882,0.509804}%
\pgfsetstrokecolor{currentstroke}%
\pgfsetdash{}{0pt}%
\pgfpathmoveto{\pgfqpoint{2.073494in}{0.084256in}}%
\pgfpathcurveto{\pgfqpoint{2.078480in}{0.084256in}}{\pgfqpoint{2.083262in}{0.086237in}}{\pgfqpoint{2.086788in}{0.089762in}}%
\pgfpathcurveto{\pgfqpoint{2.090313in}{0.093288in}}{\pgfqpoint{2.092294in}{0.098070in}}{\pgfqpoint{2.092294in}{0.103056in}}%
\pgfpathcurveto{\pgfqpoint{2.092294in}{0.108041in}}{\pgfqpoint{2.090313in}{0.112824in}}{\pgfqpoint{2.086788in}{0.116349in}}%
\pgfpathcurveto{\pgfqpoint{2.083262in}{0.119874in}}{\pgfqpoint{2.078480in}{0.121855in}}{\pgfqpoint{2.073494in}{0.121855in}}%
\pgfpathcurveto{\pgfqpoint{2.068508in}{0.121855in}}{\pgfqpoint{2.063726in}{0.119874in}}{\pgfqpoint{2.060201in}{0.116349in}}%
\pgfpathcurveto{\pgfqpoint{2.056675in}{0.112824in}}{\pgfqpoint{2.054694in}{0.108041in}}{\pgfqpoint{2.054694in}{0.103056in}}%
\pgfpathcurveto{\pgfqpoint{2.054694in}{0.098070in}}{\pgfqpoint{2.056675in}{0.093288in}}{\pgfqpoint{2.060201in}{0.089762in}}%
\pgfpathcurveto{\pgfqpoint{2.063726in}{0.086237in}}{\pgfqpoint{2.068508in}{0.084256in}}{\pgfqpoint{2.073494in}{0.084256in}}%
\pgfpathclose%
\pgfusepath{stroke,fill}%
\end{pgfscope}%
\begin{pgfscope}%
\definecolor{textcolor}{rgb}{0.150000,0.150000,0.150000}%
\pgfsetstrokecolor{textcolor}%
\pgfsetfillcolor{textcolor}%
\pgftext[x=2.195716in,y=0.070972in,left,base]{\color{textcolor}\fontsize{8.800000}{10.560000}\selectfont Training}%
\end{pgfscope}%
\begin{pgfscope}%
\pgfsetbuttcap%
\pgfsetroundjoin%
\definecolor{currentfill}{rgb}{0.172549,0.627451,0.172549}%
\pgfsetfillcolor{currentfill}%
\pgfsetlinewidth{1.138252pt}%
\definecolor{currentstroke}{rgb}{0.172549,0.627451,0.172549}%
\pgfsetstrokecolor{currentstroke}%
\pgfsetdash{}{0pt}%
\pgfpathmoveto{\pgfqpoint{3.316539in}{0.248958in}}%
\pgfpathlineto{\pgfqpoint{3.369178in}{0.248958in}}%
\pgfpathlineto{\pgfqpoint{3.369178in}{0.301597in}}%
\pgfpathlineto{\pgfqpoint{3.316539in}{0.301597in}}%
\pgfpathclose%
\pgfusepath{stroke,fill}%
\end{pgfscope}%
\begin{pgfscope}%
\definecolor{textcolor}{rgb}{0.150000,0.150000,0.150000}%
\pgfsetstrokecolor{textcolor}%
\pgfsetfillcolor{textcolor}%
\pgftext[x=3.465080in,y=0.243194in,left,base]{\color{textcolor}\fontsize{8.800000}{10.560000}\selectfont LR Test}%
\end{pgfscope}%
\begin{pgfscope}%
\pgfsetbuttcap%
\pgfsetroundjoin%
\definecolor{currentfill}{rgb}{0.552941,0.898039,0.631373}%
\pgfsetfillcolor{currentfill}%
\pgfsetlinewidth{0.813038pt}%
\definecolor{currentstroke}{rgb}{0.552941,0.898039,0.631373}%
\pgfsetstrokecolor{currentstroke}%
\pgfsetdash{}{0pt}%
\pgfpathmoveto{\pgfqpoint{3.324058in}{0.084256in}}%
\pgfpathlineto{\pgfqpoint{3.361658in}{0.084256in}}%
\pgfpathlineto{\pgfqpoint{3.361658in}{0.121855in}}%
\pgfpathlineto{\pgfqpoint{3.324058in}{0.121855in}}%
\pgfpathclose%
\pgfusepath{stroke,fill}%
\end{pgfscope}%
\begin{pgfscope}%
\definecolor{textcolor}{rgb}{0.150000,0.150000,0.150000}%
\pgfsetstrokecolor{textcolor}%
\pgfsetfillcolor{textcolor}%
\pgftext[x=3.465080in,y=0.070972in,left,base]{\color{textcolor}\fontsize{8.800000}{10.560000}\selectfont Training}%
\end{pgfscope}%
\end{pgfpicture}%
\makeatother%
\endgroup%